\documentclass[journal]{IEEEtran}
\usepackage[pdftex]{graphicx}
\usepackage{comment}
\usepackage{siunitx}
\usepackage{enumitem}
\usepackage[usenames, dvipsnames]{color}
\usepackage{graphicx}
\usepackage{floatrow}
\usepackage{mathtools,amssymb} 
\usepackage{cleveref}
\usepackage{tabularx}
\newcolumntype{C}[1]{>{\centering\arraybackslash}p{#1}}

\usepackage{multirow}
\usepackage{tabularx}
\usepackage{makecell}
\usepackage{siunitx}
\usepackage{adjustbox}
\usepackage{chngcntr}
\usepackage[normalem]{ulem}
\usepackage{algorithm}
\usepackage{algorithmicx}
\usepackage{algpseudocode}
\usepackage{xcolor}
\usepackage{lipsum}
\newcolumntype{L}[1]{>{\raggedright\arraybackslash}p{#1}}
\newcolumntype{R}[1]{>{\raggedleft\arraybackslash}p{#1}}

\usepackage[caption=false, font=footnotesize]{subfig}
\usepackage{wrapfig}

\newif\ifdraft
\draftfalse
\drafttrue
\ifdraft
  \newcommand{\ian}[1]{{\textcolor{red}{ Ian: #1 }}}
  \newcommand{\before}[1]{{\textcolor{magenta}{ BeforeModified: #1 }}}
  \newcommand{\tak}[1]{{\textcolor{blue}{ Tak: #1 }}}
\else
  \newcommand{\ian}[1]{}
  \newcommand{\before}[1]{}
  \newcommand{\tak}[1]{}
\fi
\newcommand{\blue}[1]{{\textcolor{blue}{#1}}}

\newcommand{\Erase}[1]{\if0{#1}\fi}
\newcommand{\Add}[1]{\textcolor{black}{#1}}
\newcommand{\Addafter}[1]{\textcolor{black}{#1}}
\newcommand{\AddAdd}[1]{\textcolor{black}{#1}}

\usepackage[noadjust]{cite}

\usepackage{setspace}
\usepackage{amssymb}

\makeatletter
\newcommand{\mylabel}[2]{%
\protected@write \@auxout {}{\string \newlabel {#1}{{#2}{}}}}
\makeatother

\usepackage{bigdelim}
\usepackage{lipsum}

\begin{document}

\def\Title{Data-driven Cloud Clustering via a Rotationally Invariant Autoencoder}

\title{\Title}

\author{Takuya Kurihana,~\IEEEmembership{Student Member,~IEEE,}
        Elisabeth Moyer,        
        Rebecca Willett,~\IEEEmembership{Senior Member,~IEEE,}
        Davis Gilton,
        and~Ian Foster,~\IEEEmembership{Fellow,~IEEE}
\thanks{This work was supported in part by the AI for Science program of the Center for Data and Computing at the University of Chicago; by the Center for Robust Decision-making on Climate and Energy Policy (RDCEP), under NSF grant no.\ SES-1463644; and by the U.S.\ Department of Energy, Office of Science, Advanced Scientific Computing Research, under Contract DE-AC02-06CH11357. R.~Willett and D.~Gilton are partially supported by AFOSR FA9550-18-1-0166, DOE DE-AC02-06CH11357, NSF OAC-1934637, and NSF DMS-1930049.}
\thanks{T.~Kurihana, R.~Willett, and I.~Foster are with the University of Chicago, Department of Computer Science, Chicago, IL 60637 USA (emails: tkurihana@uchicago.edu; willett@g.uchicago.edu; foster@uchicago.edu).}
\thanks{E.~Moyer is with the University of Chicago, Department of Geophysical Sciences, Chicago, IL 60637, USA (email: moyer@uchicago.edu)}
\thanks{R.~Willett is also with the University of Chicago, Department of Statistics, Chicago, IL 60637, USA.}
\thanks{D.~Gilton is with the University of Wisconsin - Madison, Department of Electrical and Computer Engineering, Madison, WI 53706, USA (email: gilton@wisc.edu).}
}

\markboth{IEEE Transactions on Geoscience and Remote Sensing}%
{Kurihana \MakeLowercase{\textit{et al.}}: \Title}

\maketitle

\label{abstract}
\begin{abstract}
Advanced satellite-born remote sensing instruments produce high-resolution multispectral data for much of the globe at a daily cadence.  These datasets open up the possibility of improved understanding of cloud dynamics and feedback, which remain the biggest source of uncertainty in global climate model projections. As a step towards answering these questions, we describe an automated rotation-invariant cloud clustering (RICC) method that leverages deep learning autoencoder technology to organize cloud imagery within large datasets in an unsupervised fashion, free from assumptions about predefined classes. We describe both the design and implementation of this method and its evaluation, which uses a sequence of testing protocols to determine whether the resulting clusters: (1) are physically reasonable, (i.e., embody scientifically relevant distinctions); (2) capture information on spatial distributions, such as textures;  (3) are cohesive and separable in latent space; and (4) are rotationally invariant, (i.e., insensitive to the orientation of an image). Results obtained when these evaluation protocols are applied to RICC outputs suggest that the resultant novel cloud clusters capture meaningful aspects of cloud physics, are appropriately spatially coherent, and are invariant to orientations of input images. Our results support the possibility of using an unsupervised data-driven approach for automated clustering and pattern discovery in cloud imagery.
\end{abstract}


%
\IEEEpeerreviewmaketitle

\section{Introduction}
\label{sec:intro}

\IEEEPARstart{A}{dvanced}
satellite-born remote sensing instruments such as the Moderate Resolution Imaging Spectroradiometer (MODIS) aboard the Terra and Aqua satellites have produced a goldmine of large observational datasets 
covering much of the globe for several decades at a daily cadence.
These multispectral and hyper-spectral datasets open up the possibility of improved understanding of cloud dynamics and feedbacks, which remain the biggest source of uncertainty in global climate model projections~\cite{stephens2005cloud}.
Scientists would like to determine, for example, whether cloud types, the distribution of different types, and the properties associated with different types have changed over time.

Automated cloud classification methods have evolved since the 1970s from 
simple statistical discriminators~\cite{Greaves70techdev,Shenk1976AMC} to various forms of machine learning 
\cite{Bankert1994CloudCO,Lee1990ANN, Welch1992PolarCA, Welch1988CloudFC, cloudsat2008,Schiffer1983TheIS},
including deep neural networks~\cite{Zhang2018CloudNetGC,Rasp2019CombiningCA,Zantedeschi2019CumuloAD,marais2020leveraging}.
Most of the latter attempts have involved supervised learning, whereby a classifier is provided with many cloud images labeled as cumulus, cirrus, etc., from which it learns the relevant features associated with those labels.
However, supervised cloud classification approaches have limitations: 
in addition to requiring large numbers of annotated examples, 
they can only identify
cloud classes for which labeled examples are provided.

Unsupervised learning approaches, which work without labeled examples to find patterns in data, have seen less use in cloud analysis, but have the potentially significant advantage from a scientific perspective of enabling data-driven exploration of large datasets, free from assumptions about predefined classes.
\Addafter{The combination of}
\AddAdd{dimensionality-reduction and clustering the resulting}
\Addafter{low-dimensional representation is a}
\AddAdd{widely-used}
\Addafter{unsupervised learning approach in pattern recognition~\cite{Caron2018DeepCF}.}
Such approaches have been applied to cloud analysis by 
Visa et al.~\cite{Visa1995NeuralNB} and
Tian et al.~\cite{Tian1999ASO} in the 1990s, 
and in 2019 by Kurihana et al.~\cite{kurihana2019cloud} and Denby~\cite{denby2019unsuper}. 
\Erase{However, while promising, these preliminary studies reveal underexplored associations between known cloud categories and their cloud clusters,
encouraging the development of their ability to learn spatial structures.}
We discuss unsupervised approaches further in Section~\ref{sec:approach}.


\AddAdd{One particular unsupervised deep learning method that leverages dimensionality reduction,}
\AddAdd{the \emph{autoencoders}~\cite{kramer1991nonlinear,goodfellow2016deep},}
\Addafter{has been widely used in image processing.}
\Erase{We present here a new approach to unsupervised cloud clustering based on \emph{autoencoders},a learning method widely used in image processing~\cite{Caron2018DeepCF}.}
An autoencoder combines an encoder, decoder, and loss function to learn both to \emph{encode} input images into a compact lower-dimensional \emph{latent space} representation
and to \emph{decode} that representation to output images, in a way that minimizes the difference between input and output: the \emph{reconstruction loss}.
If the 
latent representation both 1) preserves input image features that are important for the target application, while discarding unimportant features,
and 2) maps images that are similar (from the perspective of
the target application)
to nearby locations in latent space,
then clustering of compact representations can be used for classification. 
These properties of the latent representation and thus the resulting clusters
depend on the design of the encoder, decoder, and loss function.

\Addafter{However, while promising, preliminary unsupervised cloud clustering studies reveal underexplored associations between known cloud categories and cloud clusters.
They have not addressed the interpretation of discovered unknown patterns~\cite{denby2019unsuper} and/or remove spatial texture features of cloud imagery when feeding their algorithms~\cite{Tian1999ASO,schuddeboom2018regional}.
Furthermore, these studies have not addressed the important problem of rotation dependence. Identical clouds can be formed in any orientation and thus
algorithms should not distinguish input images based solely on their orientations~\cite{Matsuo2017TransformIA}.}

\Addafter{Another difficulty with the unsupervised approach is that}
a lack of ground truth against which to compare the output of any unsupervised learning system can make evaluation a challenging task.
The ultimate metric is whether the clusters produced by the system, when applied to a set of cloud images, are meaningful and useful.
More specifically, we want clusters that: 
(1) are \emph{physically reasonable} (i.e., embody scientifically relevant distinctions);
(2) capture information on \emph{spatial distributions}, such as textures;
(3) are \emph{separable} (i.e., are cohesive in latent space and separated from each other); 
(4) are \emph{rotationally invariant} (i.e., insensitive to the orientation of an image); and
(5) are \emph{stable} (i.e., produce similar or identical clusters) when different subsets of the data are used.
Criteria 2 and 5 apply to many autoencoder applications;
the others are, to varying degrees, specific to cloud clustering.

We present in this paper a new \textbf{rotation-invariant cloud clustering} (RICC) autoencoder framework 
capable of automating the clustering of cloud patterns and textures without any assumptions concerning artificial cloud categories.
This framework combines five elements to provide this capability:
(1) MODIS data as a source of satellite radiance data, 
from which we extract
cloud features that we show to be well associated with physical metrics;
(2) convolutional neural networks (CNNs)~\cite{Simonyan2015VeryDC,krizhevsky2017imagenet} to extract useful representations of spatial patterns from images;
(3) a loss function that uses transform-invariant techniques~\cite{Matsuo2017TransformIA} to produce latent space representations that are independent of input image orientations;
(4) an autoencoder training protocol that learns from satellite data without introducing biases; and
(5) hierarchical agglomerative clustering (HAC)~\cite{Johnson1967HAC} to extract novel clusters from the latent representation.




A conventional autoencoder trained on cloud images easily produces both high--quality outputs and latent representations that, when clustered, group into what may look like sensible classes. 
However, careful examination reveals that several of the criteria listed above---in particular, criteria 1, 2, and 4---are not satisfied. 
To permit quantitative and qualitative evaluation of autoencoder quality, we develop protocols for evaluating whether a particular design satisfies our criteria.

Application of these protocols to our RICC system demonstrates that it does indeed satisfy four of our five criteria. (The exception is stability, which we have not yet addressed, for reasons discussed later.)
The reader might be concerned that other, perhaps simpler, autoencoder designs could be used to achieve the same effect.
While it is not feasible to evaluate all possible designs,
we define and evaluate two promising alternatives, 
and explain why each fails to satisfy one or more of our criteria.

Our major contributions are:
\begin{itemize}
  \item The design of an unsupervised autoencoder framework, RICC, for grouping cloud images in multispectral satellite images into clusters that capture shape, texture, and other important properties.
  \item The design of protocols for evaluating the scientific utility of the classes produced by RICC and other unsupervised cloud clustering approaches.
  \item 
  A detailed evaluation of RICC and other cloud clustering approaches based on our evaluation protocols.
\end{itemize}

A note about terminology: we refer to RICC as a \textit{clustering} rather than a \textit{classification} system, because its goal is not to assign clouds to predefined classes but rather to identify groupings of clouds (``clusters'') with similar properties. The resulting clusters might motivate the definition of new classes, but that is not their sole, or even primary, purpose.
As the literature is not consistent in making this distinction,
we use the two terms interchangeably when reporting on other work.  

The remainder of this paper is as follows.
After reviewing approaches to cloud classification in Section~\ref{sec:approach},
we describe in Section~\ref{sec:ricc} our approach to unsupervised cloud clustering. 
Section~\ref{sec:protocols} describes the criteria by which we evaluate unsupervised cloud clustering methods, and tests for validating those criteria.
Section~\ref{sec:alternates} elucidates two alternative autoencoders, one non-rotation-invariant (NRI) and one rotation-aware (RA), that optimize different loss formulations. 
Section~\ref{sec:evals} presents the results of applying the tests of Section~\ref{sec:protocols} to the different autoencoders.
In Section~\ref{sec:experiment}, we investigate to what extent our novel cloud clusters overlap cloud categories predicted by a supervised learning approach.
We conclude in Section~\ref{sec:conclusion}.

\section{Approaches to Cloud Classification}\label{sec:approach}
\begin{wrapfigure}[16]{R}{0.5\columnwidth}
\vspace{1ex}
\centering
\includegraphics[width=\columnwidth,trim=0 0 0 0mm,clip]{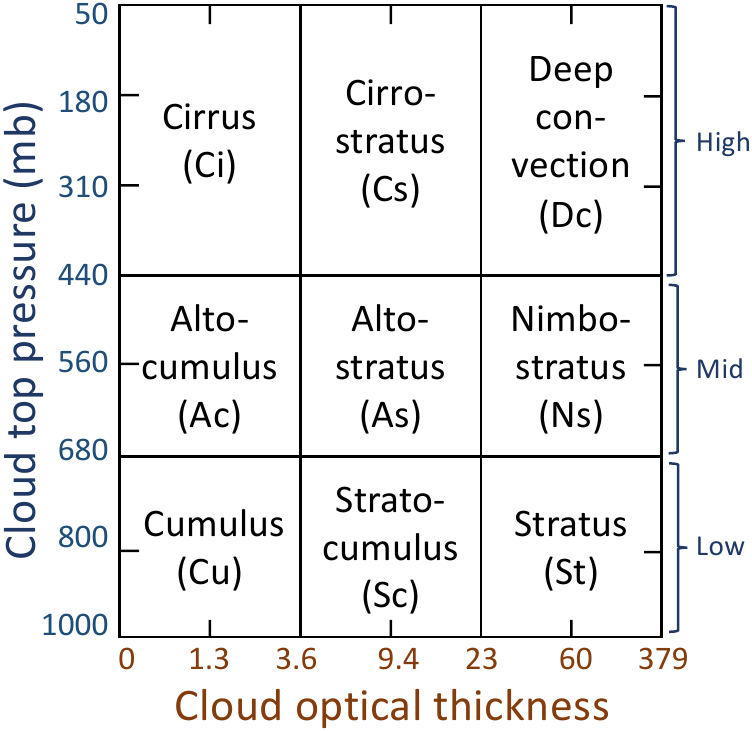}
\vspace{-1ex}
\caption{ISCCP cloud classifications. Cloud categories are defined by combinations of satellite-measured optical thickness and cloud top pressure.\label{fig:clouds}}
\end{wrapfigure}

Cloud classification is concerned with dividing the various forms of clouds into deterministic or nondeterministic groups.
Over the past two centuries, meteorologists have established various cloud classification schemes with from eight to dozens of types and subtypes, 
based on cloud texture, height, and thickness, as well as their surrounding environment.
One influential deterministic cloud classification is that established by the International Satellite Cloud Climatology Project (ISCCP)~\cite{isccp1991}.
ISCCP defines nine cloud types according to optical thickness and cloud top pressure (see Fig.~\ref{fig:clouds}), 
as measured by a global network of geostationary and polar orbit meteorological satellites.

The launch of multi-band geostationary and other meteorological satellites in the 1970s~\cite{Greaves70techdev} provided climate researchers with much richer observational data for understanding cloud feedback. 
Multi-spectral satellite sensors produce hundreds of gigabytes of observations, overwhelming human experts seeking to classify and analyze clouds to diagnose spatial-temporal effects,
and motivating work on automated cloud classification. 

In the following, we first review supervised approaches to cloud classification and their limitations, and then describe preliminary studies involving unsupervised approaches.

\subsection{Supervised Learning for Cloud Classification}
Most work to date on automated cloud classification, and in particular early work, has involved supervised learning,
whereby a classifier is provided
with many images of clouds labeled with the classes that one seeks to recognize
(e.g., cumulus, cirrus). 
The classifier then learns the relevant features associated with those labels.
Such approaches have the significant advantage of being easy to evaluate,
assuming that a reliable and sufficiently large collection of ground truth data is available.
Collections of \AddAdd{cloud images} manually labeled with ISCCP classification are widely used as ground-truth data for supervised cloud classification methods.

Early work on machine-based cloud classification algorithms integrated simple statistics with machine learning algorithms~\cite{Bankert1994CloudCO,Lee1990ANN, Welch1992PolarCA, Welch1988CloudFC}, and combined textual and cloud physical features~\cite{cloudsat2008,Schiffer1983TheIS}.
More recent work~\cite{Rasp2019CombiningCA, Xie2019SegCloudAN, Ye2017DeepCloudGC, Zhang2018CloudNetGC, marais2020leveraging} takes advantage of state-of-the-art deep learning technologies reinforced by increasingly powerful modern computing hardware
to achieve high classification accuracy in extracting relative features from images in cloud classification.

In particular, recent machine learning work has leveraged convolutional neural networks (CNNs),
\AddAdd{which has exhibited impressive classification performance in image recognition~\cite{Simonyan2015VeryDC,krizhevsky2017imagenet}}
Shi et al.~\cite{Shi2017DeepCA} 
using a pre-trained CNN plus a support vector machine to
classify two ground-based cloud observation products into both five (different pattern and thickness) and seven categories (WMO categories).
This approach demonstrated that the use of a deep neural network with convolutional layers can improve cloud classification performance.
Zhang et al.~\cite{Zhang2018CloudNetGC} developed CloudNet, a CNN inspired by a VGG architecture~\cite{Simonyan2015VeryDC}, to classify the 10 WMO cloud classes plus contrails from ground-based observations. 
This work established that deep neural networks are capable of discriminating among the 
10 WMO classes
without the use of customized features.

Two recent studies employed new labeled cloud datasets plus CNNs to study shallow marine clouds, which\AddAdd{,} while commonly classified as stratocumulus or stratus  according to the standard cloud classification\AddAdd{,} actually have rich diversity and different physical feedbacks.
Rasp et al.~\cite{Rasp2019CombiningCA} leveraged pixel-based CNNs to classify shallow marine clouds
into new four categories;
Stevens et al.~\cite{stevens2019} identified four recurrent categories of mesoscale organization in data from MODIS Terra and Aqua satellites.
Both studies used human labelers to classify cloud images according to texture, and found that the resulting custom annotations revealed physically meaningful cloud regimes when they then compared temperature, relative humidity, and vertical velocity across the four classes.

\subsection{Limitations of Supervised Learning Approaches}

Supervised cloud classification has drawbacks.
First, it requires a large annotated dataset.
Second, because human-defined cloud classes are only well defined for classic examples,
which account for just a small fraction of large satellite datasets~\cite{rossow1993comparison, hahn2001isccp},
supervised approaches fall short when used to classify diverse real-world data.
Third,
as labels are restricted to prior assumptions,
they cannot identify cloud types that were not specified in the training data but that might be relevant to climate research.

These difficulties have been highlighted by several studies. 
When Wood and Hartmann~\cite{Wood2006SpatialVO} investigated the relationship between the complex textures of low clouds and the physical characteristics found within common mesoscale (1--100~km) cloud organizations, they found that four frequent mesoscale cloud patterns associated with the existence of open- and closed-cell structures, often classified as stratus or stratocumulus
in the ISCCP classification, occur in different geographical regions and have different distributions of liquid water path, suggesting that the spatial variability of low clouds is underrepresented in the standard cloud types.

\subsection{Unsupervised Learning Approaches}\label{sec:unsup}

Unsupervised learning for cloud classification was first explored by 
Visa et al.~\cite{Visa1995NeuralNB} and
Tian et al.~\cite{Tian1999ASO} 
as an alternative to orthodox classification methods that depended on ad hoc thresholds and decision trees.
Those early works applied a self-organized map~\cite{kohonen1990self}, which preserves a topological relation when mapping a high-dimensional space to a low-dimensional space, to classify extracted features by singular value decomposition (SVD), and selected features by the input statistic. 
Tian et al.~\cite{Tian1999ASO} concluded that unsupervised learning may not always match the standard cloud types because its classifications smooth boundaries between independent cloud classes, 
leading to a classification accuracy (relative to human-defined cloud classes) that falls behind that of supervised learning.
Their results rejected the use of unsupervised learning as a promising method.

More recently, concern about the inability of artificial cloud classes to capture the diversity of clouds has spurred increased interest in the use of state-of-the-art unsupervised learning using CNNs to discover novel patterns. 
As the convolutional filters used in CNNs can learn spatial patterns,
it seems probable that a CNN-based classifier should be able to capture more relevant physics than the classifiers used in earlier work~\cite{Visa1995NeuralNB,Tian1999ASO}, 
which eliminated spatial information by flattening and learned representative statistical values based purely on mean cloud properties.

Kurihana et al.~\cite{kurihana2019cloud} used a convolutional autoencoder to capture both the spatial structure of cloud images and relevant physics in the latent representation.
That preliminary study trained the autoencoder with a compound loss function (see Appendix~\ref{sec:CNRI}) using visible to thermal bands of MODIS radiances, and proposed an early version of the five criteria described in Section~\ref{sec:protocols} as a basis for evaluation.
Results showed that both the physical association and spatially cohesive cloud cluster criteria were satisfied, but the work did not address the rotation dependency to input orientation and did not fully explore associations with established cloud categories.

In another relevant study,
Denby~\cite{denby2019unsuper} applied representation learning techniques to classify mesoscale cloud organizations. 
In representation learning, 
CNNs are trained to embed meaningful input data features in the latent representation, which is then used for a clustering or regression task.
Denby trained a residual neural network, ResNet-34, on GOES-16 images
to formulate the representation, 
and then applied hierarchical clustering to the latent representation of unseen GOES-16 images.
This work differs from Kurihana et al.~\cite{kurihana2019cloud} and the present paper, in its approach to learning spatial features.
Denby uses only the encoder and loss function parts of an autoencoder to learn how to embed spatial features in the original image into the latent representation.
By contrast, we train a full encoder--loss function--decoder triad to learn differences in shapes of input images.
The resulting clusters differentiate cloud images based on the strength of radiation in visible and infrared channels (GOES-16 bands 1 and 9); the triplet loss applied to the ResNet-34 network encourages the separation of the dissimilar texture of images.
However, their example classes still distribute similar cloud images across several clusters. 
For example, they report (Figure 2 in Denby~\cite{denby2019unsuper}) that cellular stratocumulus clouds are distributed among at least four classes,
indicating a limitation in their ability to classify small patches of large cloud structures.
Their approach is also limited in its evaluation, 
in that it does not investigate whether the mapping of clouds to clusters is aligned with physical properties---a major focus of our work.



\section{The RICC Autoencoder Method}\label{sec:ricc}
\begin{figure*}
    \centering
    \includegraphics[width=0.95\textwidth]{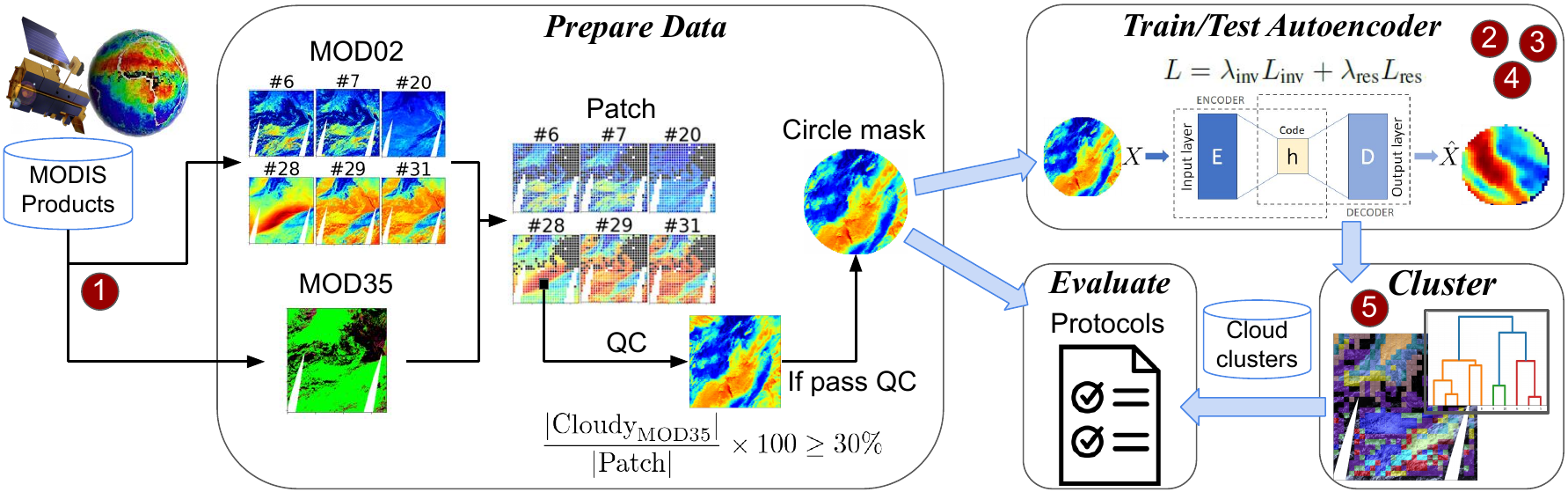}
    \caption{The RICC system workflow works with six discrete spectral bands from the MODIS Level-1B 1~km calibrated radiances product associated with physical and radiative cloud properties, from the visible to thermal infrared. Its implementation comprises four principal steps, as follows, that together implement the five elements listed in Section~\ref{sec:ricc}, as indicated by circled numbers.
    The \textbf{Prepare Data} step subdivides each large MODIS swath into smaller image units (\emph{patches}), to which are applied quality control (cloudy pixels per patch $>$30$\%$) and circular masking for stable optimization.
    The \textbf{Train/Test Autoencoder} step ingests these patches into an autoencoder training process that uses the rotation-invariant loss function to obtain autoencoder weights that produce a lower-dimensional representation with a) the rotation-invariant feature needed to map different orientations of otherwise identical input images into a uniform orientation, and b) the spatial feature needed to restore structural patterns in inputs with high fidelity.
    The \textbf{Cluster} step applies agglomerative clustering to extract novel clusters from the latent representation. 
    Finally, the \textbf{Evaluate} step applies protocols designed to determine whether the clusters produced by the system are meaningful and useful when applied to a set of test cloud patches. \Add{We summarize pseudocode for the overall RICC framework spanning in Section~\ref{sec:ricc} and Section~\ref{sec:evals} in Appendix~\ref{sec:ripseudocode}.} }
    \label{fig:flow_ricc_system}
\end{figure*}

The RICC autoencoder workflow, shown in Fig.~\ref{fig:flow_ricc_system}, 
implements a RICC autoencoder framework that combines five elements
to achieve unsupervised clustering of cloud patterns and textures without any assumptions concerning artificial cloud categories:
(1) MODIS data as a source of satellite radiance data, 
from which we extract
cloud features that are well correlated with physical metrics.
(2) A convolutional autoencoder architecture as a means of extracting useful representations of spatial patterns in images.
(3) A loss function that uses transform-invariant techniques to produce latent space representations that are independent of input origin orientation.
(4) An autoencoder training protocol that learns from satellite data without introducing biases.
(5) HAC clustering to extract novel clusters from the latent representation.
We describe each element in turn.

\subsection{Use of MODIS Satellite Data}\label{sec:modis}

We use MODIS products~\cite{justice1998moderate} from the Terra satellite instruments
to train and test our autoencoder-based unsupervised learning framework for cloud clustering.
We choose MODIS as its instruments have captured data with approximately 2330~km by 2030~km spatial coverage every five minutes since 2000, providing a variety of products for analysis.
In Sections~\ref{sec:physicalreasonableness} and~\ref{sec:experiment}, we also use data 
from the Aqua satellite instrument in a comparison with another
supervised cloud classification study~\cite{Zantedeschi2019CumuloAD}.  

MODIS provides a variety of products, from raw radiance observations to derived cloud physical parameters, all aligned to the same geolocation points on the ground.
We use three such products in this work, MOD02, MOD35, and MOD06:
\begin{itemize}
    \item 
MODIS level 1B calibrated radiance imagery at 1~km resolution (MOD021KM and MYD021KM; hereafter \textbf{MOD02})~\cite{mod021km}, which we use for autoencoder training and testing \Add{after subdividing into smaller geographical units};
\item 
MODIS35 level 2 cloud flag product at 1~km resolution (MOD35L2 and MYD35L2; hereafter, \textbf{MOD35})~\cite{Ackerman2008CloudDW,Frey2008CloudDW} aligned on MOD02 data, 
which we use for quality control; and
\item 
MODIS06 level 2 cloud product at 1~km resolution (MOD06L2 and MYD06L2; hereafter, \textbf{MOD06})~\cite{baum12, steven17}, which we use to evaluate to what extent clustering results obtained via our autoencoder, when trained only on radiance data, reflect cloud physical properties.
\end{itemize}

Note that we do not use the MOD06 variables as inputs to our autoencoders.
This is because
we want to perform clustering free from any assumptions captured in those variables.
Furthermore, the MOD06 product is estimated at pixel resolution from MOD02 via various theoretical bases and thermodynamics, 
which can truncate spatial information concerning certain scales of cloud organization.

\begin{table*}
\centering
\caption{The 12 MODIS products used to train and evaluate our autoencoders and clustering method. Source: NASA Earthdata.}\label{tab:products}
\begin{small}
\begin{tabular}{C{1.1cm}p{6.2cm}C{0.9cm}p{4cm}p{0.1cm}p{3.1cm}}
\hline
 Product & Description &Band& Primary Use  && Our Use \\ \hline
 MOD02   & Shortwave infrared (1.230-1.250 $\mu$m)         & 5 & Land/cloud/aerosols properties & & Replaces band 6 in \S\ref{sec:experiment} \\
         & Shortwave infrared (1.628-1.652 $\mu$m)         & 6 & Land/cloud/aerosols properties & \rdelim\}{6}{3cm}& \multirow{6}{3cm}{Six bands used to train autoencoders} \\
         & Shortwave infrared (2.105–2.155 $\mu$m)         & 7 & Land/cloud/aerosols properties &&  \\
         & Longwave thermal infrared (3.660–3.840 $\mu$m)  & 20& Surface/cloud temperature  && \\
         & Longwave thermal infrared (7.175–7.475 $\mu$m)  & 28& Cirrus clouds water vapor   &&\\
         & Longwave thermal infrared (8.400–8.700 $\mu$m)  & 29& Cloud properties  && \\
         & Longwave thermal infrared (10.780–11.280 $\mu$m)& 31& Surface/cloud temperature   &&\\ \hline
  MOD35  & Cloud mask             &  & Cloud pixel detection   &&\\ \hline
  MOD06  & Cloud optical thickness&  & Thickness of cloud   &\rdelim\}{4}{*}& \multirow{4}{3cm}{Used to evaluate clusters based on MOD02-trained autoencoder } \\
         & Cloud top pressure     &  & Pressure at cloud top   &&\\
         & Cloud phase infrared   &  & Cloud particle phase   &&\\
         & Cloud effective radius &  & Radius of cloud droplet &&\\ \hline
\end{tabular}
\end{small}
\end{table*}

\subsubsection{Details on MOD02 Bands}
The MOD02 product has 36 spectral bands in the range 0.4--14.4 $\mu$m (i.e., visible to thermal infrared) from the year 2000 to the present. 
We list the MOD02 radiance bands used in this work in Table~\ref{tab:products}.
Except as noted in the next paragraph, we use
in our patches the six bands listed in rows 2--7.
We use these bands,
which range from the shortwave to longwave,
because they are important for the retrieval algorithms used to compute the MOD06 variables listed in Table~\ref{tab:products}.
\AddAdd{They are also relatively uncorrelated with bands in distant wavelengths, representing distinct physical properties.}
\Add{This pre-selection of six bands also facilitates practical implementation by reducing the computing resources required for analysis.
The results from Tests~\ref{test11}--\ref{test42} in Section~\ref{sec:evals} provide additional support for this choice of bands. 
Automated band selection, for example via neural networks, would be an interesting topic for future work.}
Specifically, bands 6, 7, and 20 are used to estimate cloud optical properties (e.g., cloud optical thickness and effective radius) and bands 28, 29, and 31 are used to separate high and low clouds and detect cloud phase;
thus, we hope that a model trained on these six bands may learn associations of cloud physical parameters, especially cloud optical thickness and cloud top pressure used for ISCCP cloud classification.
As is common in neural network training, 
we normalized the values in each selected band to the range [0, 1], 
so as to prevent different scales of values in different bands from biasing training. 

Due to a known stripe noise issue at band 6 (1.628--1.652 $\mu$m) in Aqua~\cite{rakwatin2007stripe,gladkova2011quantitative}, we use band 5 (1.230--1.250 $\mu$m: the first row in Table~\ref{tab:products}) as an alternative to band 6 for the study reported in Section~\ref{sec:experiment}. 
We select that band because it is primarily sensitive to cloud optical thickness (the 1.6 $\mu$m and 3.7 $\mu$m bands are used for cloud optical thickness retrievals~\cite{steven17}) and is the closest to band 6.

The MODIS instrument 
generates hierarchical data format (HDF) image files of
2030 pixels $\times$ 1354 pixels $\times$ 36 bands.
To enable efficient learning of cloud features on neural networks, we use a smaller geographical unit, a \emph{patch}, as the element on which we perform clustering.
\AddAdd{Every} 128 pixels $\times$ 128 pixels $\times$ 6 band subset is cropped from a MODIS image.


\begin{figure}
    \centering
    \includegraphics[width=0.9\columnwidth,trim=3mm 3mm 6mm 4mm,clip]{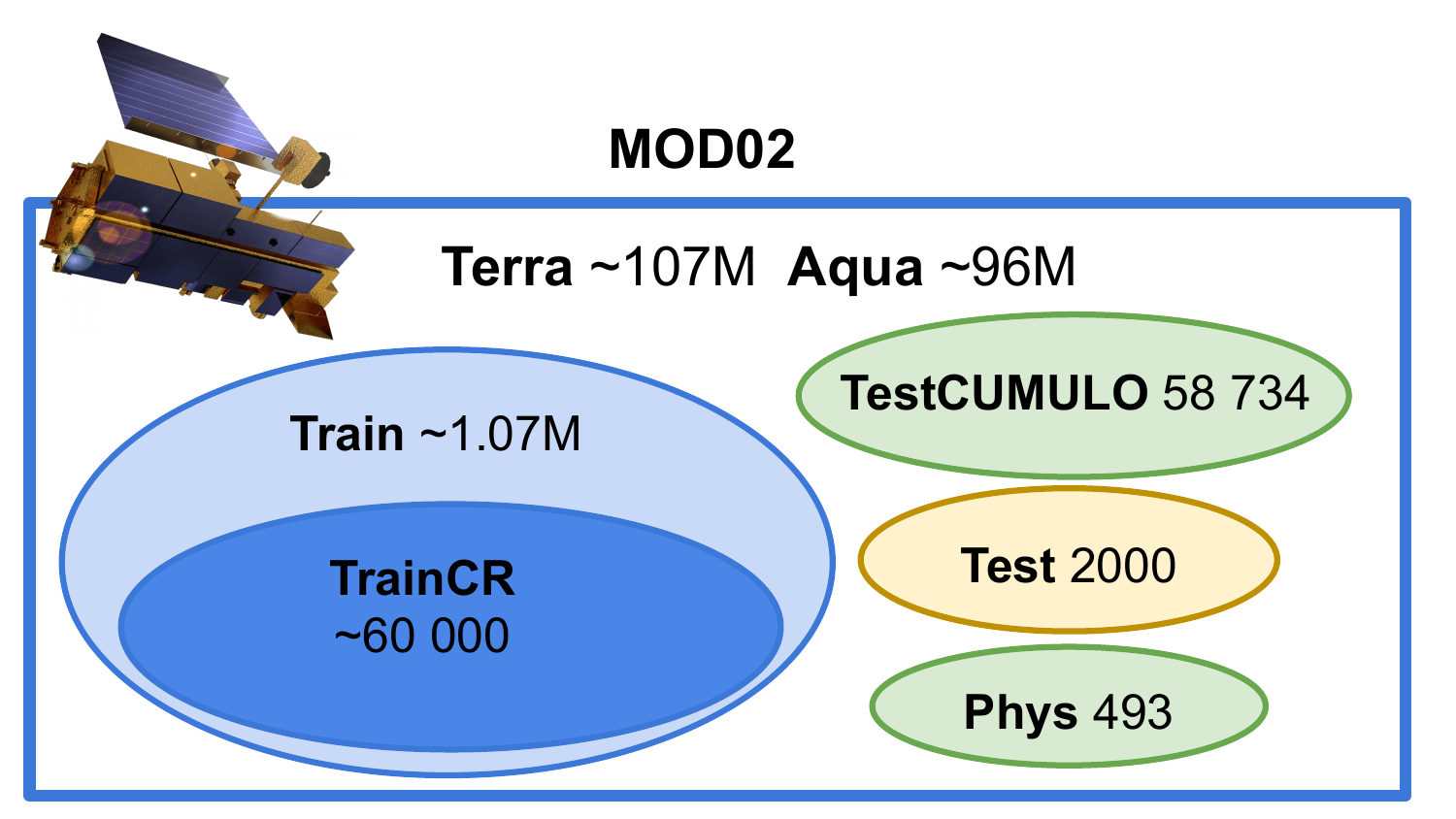}
    \caption{The five shaded ovals (not to scale) each denote a different subset of the full 19-year MODIS MOD02 1~km calibrated radiances dataset that we use in our work.
    For each, we indicate the number of patches.
    Training subsets are on the left and test subsets on the right.
    Of the three test subsets, \texttt{TestCUMULO} and \texttt{Phys} are composed of patches that are adjacent in space and time, while \texttt{Test} is not.}
    \label{fig:data}
\end{figure}

\subsubsection{MOD02 Subsets Used in Evaluation}\label{sec:testdata}
The full 19-year (2000--2018) Terra MOD02 dataset corresponds to about 100 million
valid 128 pixels $\times$ 128 pixels patches of the form just described.
From this set, we extract the following subsets (see also Fig.~\ref{fig:data})
for training and testing. 
\begin{itemize}
    \item \texttt{Train}: \num{1075994} patches used to train the NRI autoencoder. This is about 1\% of the full 19-year (2000--2018) MOD02 (Terra) dataset.
    \item  \texttt{TrainCR}: A \num{60000}-patch subset of  \texttt{Train}, used to train the RA and RI autoencoders.
    \item  \texttt{Test}: \num{2000} patches selected at random from the 19-year MODIS dataset minus those patches already selected for \texttt{Train}, 
    used to test the various autoencoders.
    \item  \texttt{TestCUMULO}: \num{58734} patches corresponding to all daytime swathes of MOD02 (Aqua) on January 1, 2008, which we use in Section~\ref{sec:physicalreasonableness} to investigate associations with cloud physical parameters, and in Section~\ref{sec:experiment} to compare the resulting clusters with a labeled dataset.
    \item \texttt{Phys}: 
    493 patches selected from a region off the coast of California (11--44$^\circ$ N, 144--112$^\circ$ E) on December 1st, 2015, which we use in Sections~\ref{sec:spatialinfotest} and~\ref{sec:test:separable} to investigate associations with cloud physical parameters in a spatially continuous region.
\end{itemize}

We select the latter region as it features marine stratocumulus clouds that form in the stable air off the west coast of the continent, where upwelling of cold water prevents deep convection from the sea surface and stabilize low cloud~\cite{woodStCreview}, and the specific date as it exhibits both low marine stratocumulus, which occur frequently in winter, and high cloud, and thus provides a good test as to whether our framework clearly separates these different cloud objects. Following our quality-control scheme, we remove patches comprised of invalid pixels and patches containing less than 30\% of cloud pixels as detected by the MOD35 cloud flag product.

\subsection{Autoencoder Network Architecture}

\begin{figure}
    \centering
    \includegraphics[width=.8\linewidth]{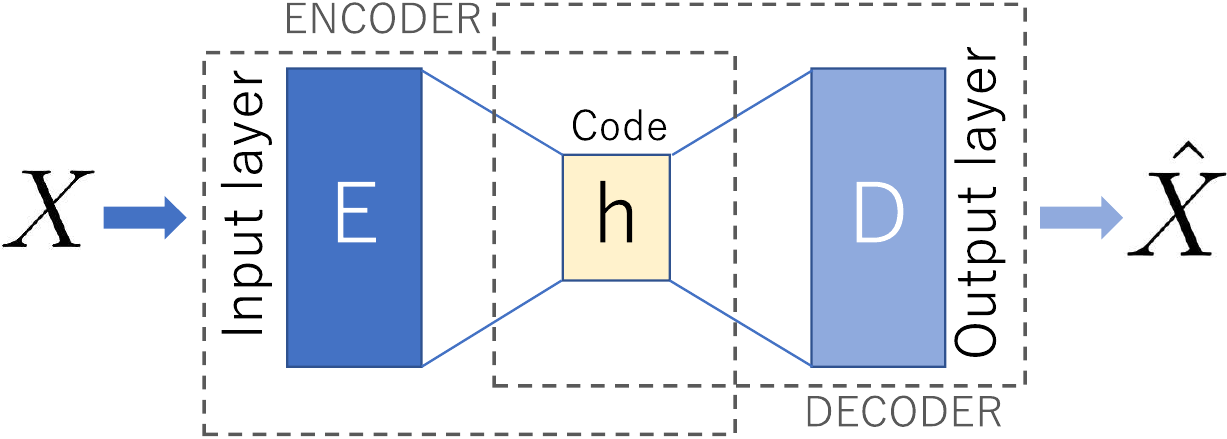}
    \caption{Autoencoder cartoon, showing input $X$, encoder network \texttt{E}, latent representation \texttt{h}, decoder network \texttt{D}, output $\hat{X}$. Adapted from Wikipedia.
    }
    \label{fig:autoencoder}
\end{figure}

The autoencoder~\cite{kramer1991nonlinear,goodfellow2016deep,Hinton2011TransformingA,Hinton94AEscience} is a widely accepted unsupervised learning method that
learns to convert input images first to a compact lower-dimensional ``latent space" representation
and then from that representation to an output, in a way that minimizes the difference between input and output: the reconstruction loss.
An autoencoder typically uses a symmetric encoder-decoder architecture (see Fig.~\ref{fig:autoencoder}), with the encoder $E$ extracting essential information from high-dimensional inputs into a lower-dimensional intermediate layer, the \emph{latent representation},
and the decoder $D$ then approximating the inputs from the latent representation. 
The decoder restores the original data from $h$, a latent representation at the bottleneck layer, where the input data $X = \{x_1, \cdots, x_n\}$ are encoded as $z = E(X)$, such that $\hat{X} = D(z)$. 

Training an autoencoder network generates a model capable of restoring the principal patterns of the original structure in an unsupervised manner.
Because a trained autoencoder retains only relevant features in the latent representation, omitting noise in the inputs, such networks are commonly used as a preprocessing tool~\cite{Caron2018DeepCF} prior to image processing
tasks such as clustering or regression, anomaly detection~\cite{Sakurada14Anom}, \Erase{denoising~\cite{Vincent2008ExtractingAC}}, and inpainting~\cite{Pathak2016ContextEF,Suganuma2018ExploitingTP}. 
In addition, 
autoencoders have been applied successfully to a wide variety of image
recognition problems~\cite{Caron2018DeepCF} via
clustering and classification of the compact representation.
\Add{A variety of autoencoder architectures and learning metrics have been developed; recent popular examples are the variational autoencoder~\cite{Kingma2014AutoEncodingVB}, which approximates a continuous latent space to generate new data, and the denoising autoencoder~\cite{Vincent2008ExtractingAC}, which reconstructs corrupted structures in the original data so that the model gains robustness.
In this study, we use a convolutional autoencoder with orthodox end-to-end training, as our goal is simply to extract features into the low-dimensional embedding.}

\begin{figure}
		\centering
	\includegraphics[width=\textwidth,trim=4mm 3mm 3mm 3mm,clip]{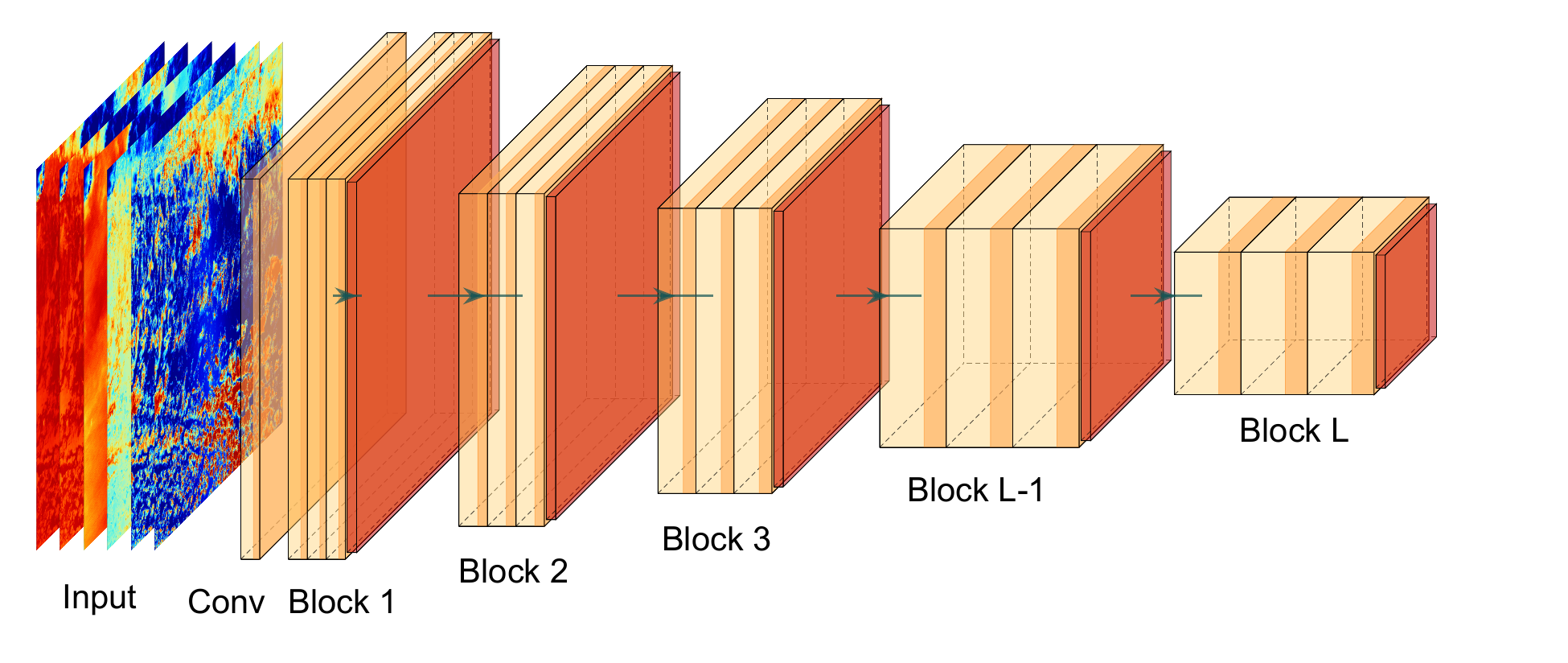}
	\caption{The RICC (and L2 and RA: see below) encoder consists of L blocks, each with three convolutional layers (orange) activated by leaky ReLU, and with batch normalization (red) applied at the final convolutional layer in each block before activation.}\label{fig:network:ricc}
\end{figure}

We summarize the detailed configuration of our encoder in Fig.~\ref{fig:network:ricc} and Table~\ref{model-table}.
The decoder (not shown) has the same structure, but mirrored.
We follow standard practice in integrating CNNs into our autoencoder architecture
so as to preserve the spatial structure in the input images and to detect the edges of input cloud objects.
We stack the convolutional layers into \textit{blocks},
each consisting of three convolutional layers of the same size, plus batch normalization~\cite{Ioffe2015BatchNA} at the end of the block.
(Three layers are required to allow the bottleneck representation to learn the structure of the entire input image.)

We use a 3$\times$3 filter, following Simonyan et al.~\cite{Simonyan2015VeryDC} who proposed that the smaller odd-number kernel size with multiple convolutional layers can learn entire features in its inputs.
We halve the resolution of the incoming image but double the filter size each time that we move from one convolutional block to the next,
using max pooling with stride 2 in the NRI and RA autoencoders.
However, as the pooling operation can trigger a loss of spatial structure in CNNs~\cite{Hinton2011TransformingA}, 
we alternate the max-pooling layer with a stride convolution layer, as stride convolution with increased stride performs as accurately as max pooling in image recognition tasks~\cite{springenberg2014striving}.
We use a leaky rectified linear unit (leaky ReLU), a refinement of ReLU~\cite{Nair2010RectifiedLU}, for the activation function, $f(x) = \max(0.3x,x)$.
We use stochastic gradient descent~\cite{lecun1998gradient} as our optimization method.
\Add{We present training implementation details in Section~\ref{sec:training-protocol}, and results of training loss of our proposed autoencoder in Fig.~\ref{fig:gridsearch_loss_cloud} in Appendix~\ref{sec:valid_cloud}.
We further investigate features \AddAdd{learned} by our neural network in Appendix~\ref{sec:ricode}.
}

\begin{table}[thb]
\caption{The layers used in the RICC autoencoder's encoder}
\label{model-table}
\centering
   \begin{tabular}{l|r|r|r}
    \hline
    Layer & Resolution & Filters & Operation\\
    \hline\hline
    Input   &  128 x128&     6& Resize\\ \hline
    Conv    &   32 x 32&     6& Conv x 1\\ \hline
    Block 1 &   16 x 16&    32& Conv x 3\\ \hline
    Block 2 &   16 x 16&    64& Conv x 3\\ \hline
    Block 3 &     8 x 8&   128& Conv x 3\\ \hline
    Block 4 &     4 x 4&   256& Conv x 3\\ \hline
    Block 5 &     2 x 2&  512& Conv x 3\\
    \hline
    \end{tabular}
\vskip -0.1in
\end{table} 

\subsection{Rotation-Invariant Loss Function}\label{sec:ri:loss}

\begin{figure}
    \centering
    \subfloat[Conventional autoencoder]{\includegraphics[clip, width=0.8\textwidth]{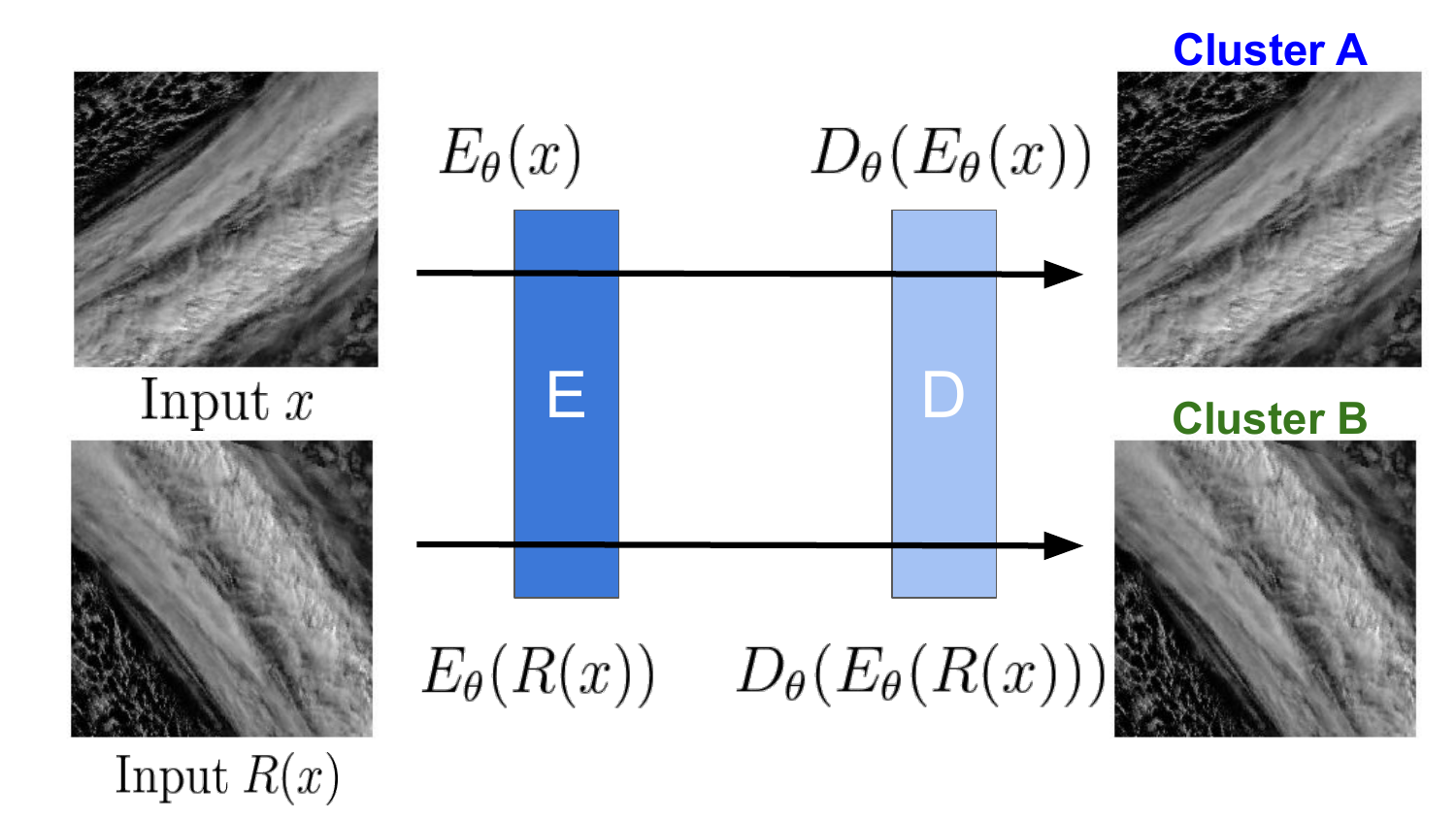}
    \label{fig:conceptOrdAE}}
    \\
    \subfloat[Rotation-invariant autoencoder]{\includegraphics[clip, width=0.8\textwidth]{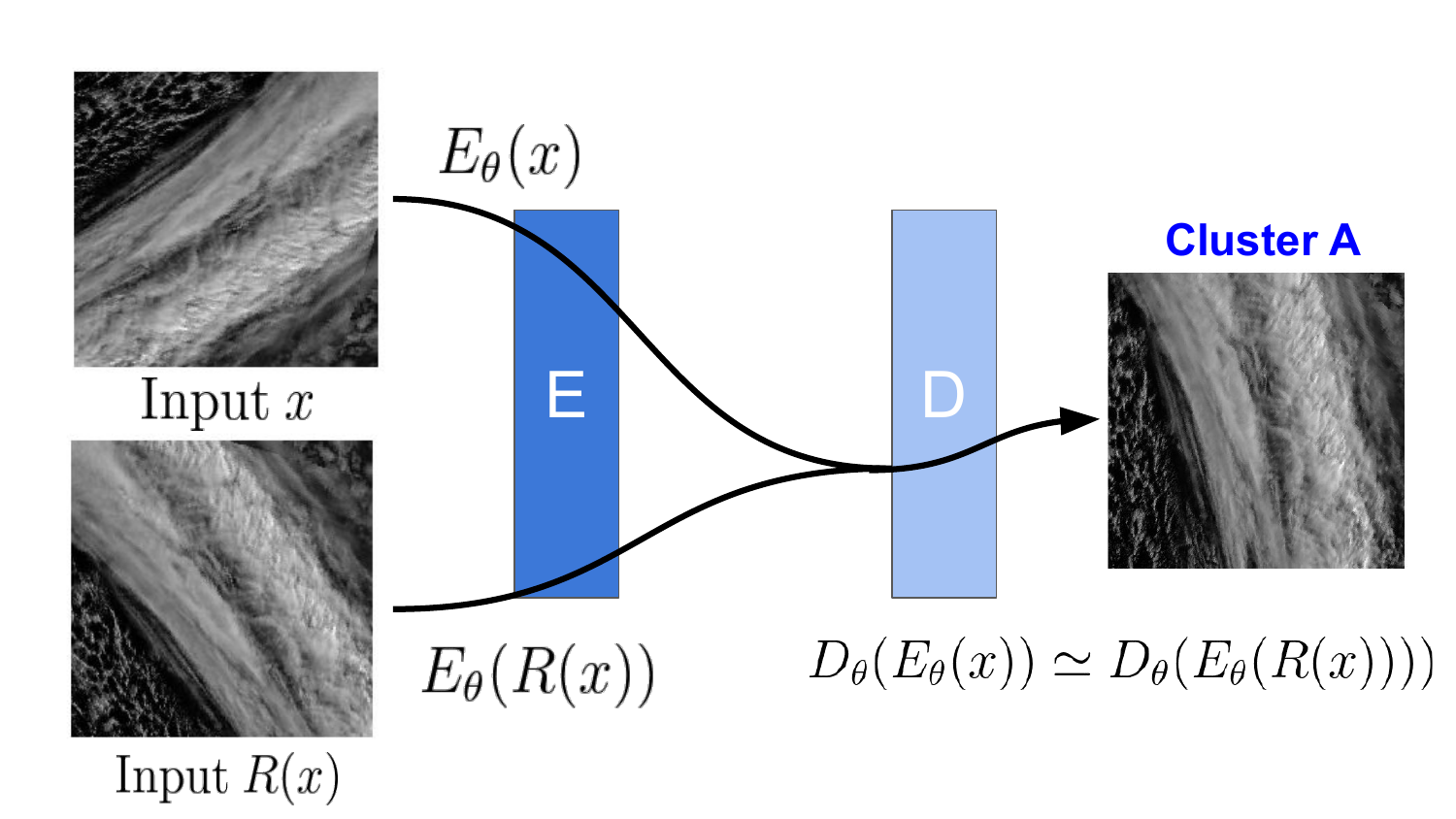}
    \label{fig:conceptRI}}
    \caption{\Add{Illustration of the unsupervised learning process when using (a) conventional autoencoder with Equation~\ref{standardloss} vs.\ (b) rotation-invariant autoencoder with Equation~\ref{eq:ri_loss}. 
    Because the conventional autoencoder preserves orientation in the latent representation, the two input images are reconstructed perfectly but are placed in different clusters, A and B. In contrast, the rotation-invariant autoencoder produces a rotation-invariant latent representation that allows for similar two reconstructions and permits clustering to place them both in the same cluster, A. 
 }}\label{fig:concept-AEs}
\end{figure}

The loss function is used during autoencoder training to quantify the difference between an original and restored image;
the optimization process then works to minimize that difference.
In a conventional autoencoder, this function might be:
\begin{equation} \label{standardloss}
    L(\theta) = \sum_{x \in S} || x - D_{\theta}(E_{\theta}(x))  ||^p_p ,
\end{equation}
where $S$ is a set of training inputs; 
$\theta$ is the encoder and decoder parameters, for which values are to be set via training;
$x$ and $D_{\theta}(E_{\theta}(x))$ are an input in $S$ and its restoration, respectively;
and $|| \cdot ||^p_p$ denotes the L$_p$ loss, the $p$th power of the $p$-norm of the inputs and restorations.

\Add{However, as we shall see in Fig. \ref{fig:concept-AEs}, this simple loss function does not preserve important features in cloud images.
Optimization of Equation~\ref{standardloss} considers that an image $x$ and the rotated image $R(x)$ are different representations, as shown in Fig. \ref{fig:conceptOrdAE}; thus, clustering may place them in different clusters.
Such rotation dependency is suboptimal for cloud analysis because any particular cloud type can occur in different orientations, and for that reason, it is not desirable for autoencoders to assign cloud types that depend solely on orientation.
Our rotation-invariant loss function is designed to generate a similar latent representation for clouds with similar morphology, regardless of their orientation, so that they are placed into the same cluster (Fig. \ref{fig:conceptRI}).}

RICC instead adapts the loss function used in the shifted transform invariant autoencoder of Matsuo et al.~\cite{Matsuo2017TransformIA} to obtain a loss function $L$ that
combines both a rotation-invariant loss, $L_{\text{inv}}$,
to learn the rotation-invariance needed to map different orientations of otherwise identical input images into a uniform orientation, and 
a restoration loss, $L_{\text{res}}$, to learn
the spatial information needed to restore structural patterns in inputs with high fidelity:
\begin{equation}\label{eq:ri_loss}
    L = \lambda_{ \text{inv}} L_{\text{inv}} + \lambda_{ \text{res}} L_{\text{res}},
\end{equation}
where the scalars $\lambda_{\text{inv}}$ and $\lambda_{\text{res}}$ are the weights of the respective loss terms. 
\Add{Note that both parameters are important; 
an extensive parameter search, described in Appendix~\ref{sec:valid_cloud} and depicted in Fig. \ref{fig:gridsearch_result_all}, shows that the autoencoder performance depends on the values of each parameter, not solely their ratio.}
\Addafter{Empirically we observe that the numerical optimization methods used to train the network (which rely on other hyperparameters such as learning rate schedules or the variance of initial values) benefit from the added flexibility of the second hyperparameter.}
We describe the two loss terms in turn.

\textit{Rotation-Invariant Loss}:
Matsuo et al.~\cite{Matsuo2017TransformIA} define a transform variance term that sums,
over both the minibatch $S$ and a set of shift parameters,
the differences between the restored input and restored translated input:
\[
     L_{\text{matsuo}}(\theta) = \sum_{x \in S}\sum_{T \in {\cal T}} {|| D_{\theta}(E_{\theta}(x)) - D_{\theta}(E_{\theta}(T(x)))||^2_2},
\]
where ${\cal T}$ is a set of shift transform operators, 
each of which translates objects in an input by a different shift vector.

We modify their formulation
by replacing their shift transform operator $T$
with a scalar rotation operator $R$ and
by computing the average, rather than the sum, of the differences for each set of $N$ rotated images. The loss term is then: 
\begin{equation}\label{tinv}
     L_{\text{inv}}(\theta) = \frac{1}{N}\sum_{x \in S}\sum_{R\in {\cal R}} {|| D_{\theta}(E_{\theta}(x)) - D_{\theta}(E_{\theta}(R(x)))||^2_2},
\end{equation}
where ${\cal R}$ is a set of rotation operators, 
each of which rotates an input by a different angle.

$L_{ \text{inv}}$ computes, for each image in a minibatch, 
the difference between the restored original and all restored rotations.
Thus, the minimization of Equation~(\ref{tinv}) by the optimization process
produces similar latent representations for an image, regardless of its orientation.

The second modification facilitates tuning of the $\lambda$ parameters 
(see Appendix~\ref{sec:valid_cloud}) 
by balancing the rotation-invariant loss term,
which is now a sum of differences over the minibatch $S$, 
and the restoration loss term, 
which as we define next, computes the sum of minimum differences over the minibatch.

\textit{Restoration Loss}: To enable a network to learn spatial substructure in images, we also adapt the restoration error term of Matsuo et al.~\cite{Matsuo2017TransformIA}
to minimize the difference between the restoration, $D_{\theta}(E_{\theta}(x))$,
of an original image $x$, and the original image when rotated by an operator $R \in {\cal R}$,
so that optimization obtains values $\theta$ that preserve spatial structure
in the input while allowing for transformation to a canonical orientation.
\begin{equation}\label{tres}
    L_{\text{res}}(\theta) = \sum_{x \in S} \displaystyle \min_{R \in {\cal R}} {|| R(x) - D_{\theta}(E_{\theta}(x)) ||^2_2} .
\end{equation}

Matsuo et al.~\cite{Matsuo2017TransformIA} also include a sparsity term, with the goal of minimizing the number of neurons required to achieve translation invariance in their latent representation. For simplicity of optimization, we exclude that term.

\subsection{Training Protocol}\label{sec:training-protocol}

Our training protocol optimizes weights until the rotation-invariant loss converges on the training data. 
We apply a hyperparameter search process, described in Appendix~\ref{sec:valid_cloud}, to balance parameters between the transform-invariant and restoration terms so as to achieve both rotation-invariant and spatial features.
\AddAdd{But, we do not expect our autoencoder to be robust to changes in hyperparameters.
The results of clustering can change if we change the hyperparameters of autoencoders.}
We trained autoencoders on our training patches for 100 epochs using stochastic
gradient descent with a learning rate of 10$^{-2}$ on an NVIDIA V100 GPU in the University of Chicago’s Midway compute cluster.
We used a batch size of 16 so as to accommodate data in memory during the iterative image rotations.

\subsection{Clustering Method}\label{sec:HAC}

\Erase{The final step in our method development is to cluster the latent representations produced by the autoencoder for a set of cloud images.}
\Add{The final step in our method development is to cluster the latent representations produced by the trained autoencoder for a set of cloud images, in order to identify semantically meaningful cloud clusters.}
\Addafter{While other clustering algorithms, for instance, non-negative matrix factorization (NMF)~\cite{lee1999learning}, can approximate input data into a low-dimensional matrix} \AddAdd{(i.e.,} 
\Addafter{produce dimensionally reduced representation similar to autoencoder) and can be used for clustering, applications of NMF~\cite{wang2018detecting, wang2020robust} have not \AddAdd{addressed the issue of rotation-invariance} within their training process. 
Therefore, \AddAdd{we address rotation-invariant and dimensionality reduction separately for the autoencoder, and group} similar latent representations based on their distances for clustering in our proposed approach.}
We apply hierarchical agglomerative clustering (HAC)~\cite{Johnson1967HAC} to merge pairs of clusters from bottom to top, minimizing at each step the linkage distance among merging clusters. 
We use Ward's method~\cite{Ward1963HierarchicalGT},
which minimizes the variance of merging clusters, as the linkage metric. 
Suppose that we have two clusters $C_A$ and $C_B$ containing data points $X = \{x_{i_1}, \cdots, x_{i_N}\}$ for cluster $i \in \{A,B\}$. HAC with Ward's method computes the linkage distance as:
\begin{equation}
    d(C_A, C_B) = E(C_A \cup C_B ) - E(C_1) - E(C_2),
\end{equation}
where $E(C_i) =\sum_{x \in C_i} (d(x, c_i))$ for the centroid $c_i = \sum_{X \in C_i} \frac{x}{|C_i|}$, denotes the sum of the weighted square distances between all data points in the cluster and the centroid $c_i$.

We use HAC as our clustering method because its initialization strategy of repeatedly merging data points has been shown to give more stable results than other methods that use random starting centroids~\cite{macqueen1967kmeans}.
We have shown in previous work~\cite{Jenkins2019DevelopingUL} that HAC clustering results outperform those obtained with other common clustering algorithms.
\Add{We work with 12 clusters here}\AddAdd{,}
\Add{as we have found that the clustering agreement score defined in Section~\ref{sec:protocols} typically stabilizes
for 10 or more clusters, and 12 or more clusters are commonly used in unsupervised cloud classification studies~\cite{denby2019unsuper,schuddeboom2018regional}. 
We leave methods for determining an optimal number of clusters for future work.
}

\section{Evaluating Unsupervised Cloud Clustering}\label{sec:protocols}
An inherent difficulty in validating unsupervised classification methods is that there is no ground truth against which to evaluate the outcomes.
Instead, we evaluate the classes produced by the method using a series of protocols. 
These protocols include, in addition to conventional image quality measures,
measurements of the degree to which various rearrangements of the pixels in images lead to the same cluster assignments (to evaluate whether latent representations capture spatial patterns);
measurements of the degree to which the same image maps to the same class when rotated
(to evaluate whether latent representations are rotation-invariant);
and measurements of the physical properties associated with different classes.
In the introduction, we noted five criteria that an automated cloud clustering method must satisfy for the purpose~\cite{kurihana2019cloud}.
Here we expand upon these descriptions and introduce the tests that we use to evaluate four of these criteria, based on the quantitative and qualitative \AddAdd{requirements}\Erase{norms} listed in Table~\ref{tab:autoencoders}. 
For Test \ref{test11}, we use the largest test dataset~\texttt{TestCUMULO}; for Tests \ref{test21}--\ref{test42}, we use the smaller test datasets~\texttt{Test} and~\texttt{Phys}.

\subsection{Criterion 1: Physically Reasonable Clusters}\label{sec:phys}

We expect the clusters produced by a clustering method to be \emph{physically reasonable}, with scientifically relevant distinctions.
To this end, we define a test that
examines quantitative differences among cloud physics parameters.
Results are in Section~\ref{sec:physicalreasonableness}.

\textbf{Test \ref{test11}: Cloud physics parameters.}\mylabel{test11}{1}
We define reasonableness in terms of whether each cluster produced by the unsupervised clustering method is associated with distinct distributions of four selected retrieved cloud physics parameters from the MOD06 product:
cloud optical thickness (COT), cloud top phase (CPH), cloud top pressure (CTP), and cloud effective radius (CER).
We select COT and CTP because these two parameters are used for ISCCP cloud classification to evaluate the compatibility between novel types of clouds and the established ISCCP cloud classes~\cite{isccp1991}.
We select CPH and CER in order to consider differences in cloud properties that are not considered in the nine ISCCP classes.
For instance, comparing CTP and CPH allows us to determine whether high CTP values and the CPH liquid phase appear concurrently in the same cluster, as we would expect, because high CTP represents lower altitude clouds for which cloud particles should be in the liquid phase.
We expect our clustering system to learn complex combinations of radiances, enabling our trained autoencoders to reflect information from more physical variables than does ISCCP's classification.

We apply the reasonableness test as follows.
We first verify that the values for each parameter are not randomly distributed,
which would contradict cloud physics. 
Then, for each parameter, 
we examine whether different clusters show different distributions.
We do this by computing for each cluster pair the inter-cluster correlation
and then computing the median of the resulting inter-cluster correlation coefficients. 
If for at least one of the four parameters this median is less than 0.6, 
\AddAdd{a cutoff commonly used to indicate the empirical threshold of no strong correlation~\cite{hinkle2003applied} in a variety of natural sciences,}
then we conclude that the clusters do indeed group patches based on physical properties. \Add{A median inter-cluster histogram correlation of less than 0.6 suggests that half of the histogram pairs do not perfectly match, and}
\AddAdd{thus we declare that the cloud clusters have distinct patterns.}

Note that we train our autoencoders only on the MOD02 input radiances,
not on the MOD06 COT, CPH, CTP, and CER parameters.
Thus, this test determines whether our training and clustering process is able to embed into the latent representation the distinctions that are recorded by the MOD06 parameters.
Recall from Section~\ref{sec:modis} that there are six radiance bands.
Three of these bands (6, 7, 20, at 1.6, 2.1, 3.7 $\mu$m, respectively) are used in the algorithm that estimates cloud optical properties (e.g., cloud optical thickness and effective radius), and the other three (28, 29, 31, at 7.3, 8.5, 11 $\mu$m) are used to separate high and low clouds and to detect cloud phase.

\subsection{Criterion 2: Spatial Distribution}\label{sec:test:spatial}
We expect the clusters produced by the unsupervised clustering method to capture information on cloud \emph{spatial distributions}. This means that they should not be reproducible by using only mean properties over the target area.
To this end, we define three tests, with the results in Section~\ref{sec:spatialinfotest}.

\textbf{Test 2.1: Spatial coherence.} \mylabel{test21}{2.1}
We evaluate whether the clusters produced by an autoencoder demonstrate more spatially coherent assignments than those obtained by clustering patch-mean cloud parameters.
When clustering without an autoencoder, 
we apply HAC to the patch-mean values of COT, CTP, CPH, and CER.

\textbf{Test 2.2: Smoothing.}\mylabel{test22}{2.2}
We examine how the cluster assignments for cloud images change when we alter the spatial resolution of the images via smoothing.

\textbf{Test 2.3: Scrambling.} \mylabel{test23}{2.3}
We examine how cluster assignments scramble pixels in patches so as to remove spatial patterns while preserving the distribution of values.

If the cluster assignments do \emph{not} change in the latter two tests,
we conclude that our autoencoder is not learning spatial information, 
because the encoder generates similar representations when the input images
are transformed to remove spatial information.

\paragraph{Adjusted mutual information (AMI) score}
We use the AMI score~\cite{Nguyen2014AMI},
which adjusts the mutual information (MI) score to account for chance,
to measure the extent to which cluster assignments agree for inputs of different spatial resolutions.
Given two clustering assignments $U$ and $V$, the AMI is computed as:
\begin{equation}
    \text{AMI}(U,V) =\frac{\text{MI}(U,V)-\mathbb{E}(\text{MI}(U,V))}{\text{avg} \{H(U), H(V)\}- \mathbb{E}(\text{MI}(U,V)) } , 
\end{equation}
where $H(\cdot)$ depicts entropy, which formally defines with probability $P(u)$ as 
\begin{equation}
    H(U) = -\displaystyle \sum_{u\in U} P(u) \log{}P(u) 
\end{equation}
and $\text{MI}(U,V)$ is the mutual information between clustering assignments $U$ and $V$, as determined by their joint distribution $P(u,v)$ and their respective probabilities $P(u)$ and $P(v)$:
\begin{equation}
    \text{MI}(U,V) = \displaystyle \sum_{u\in U} \displaystyle \sum_{v\in V} P(u,v) \log{}\frac{P(u,v)}{P(u) P(v)} .
\end{equation}

The AMI of two sets of cluster assignments is 1 if the assignments match perfectly, 
regardless of the assigned labels, and 0 if there is no match.
If the AMI score between the clustering results obtained for two different spatial resolutions is low, we conclude that the trained autoencoder has learned spatial information in its latent representation;
if the agreement score is high, we conclude that it
has failed to capture spatial patterns and thus is likely instead encoding information 
about the distribution of pixel values in input images.  

\begin{figure}[htbp]
\centering
\hrule
\vspace{1ex}
\begin{minipage}[c]{0.70\textwidth}
\begin{singlespace}
\begin{tabbing}
\=aaa\=aaa\=aaa\=\kill
\> \blue{\# Smooth a single image}\\ 
\> \texttt{smooth(x, k):}\\
\>\> \texttt{p$_{\{(i,j),(i,j+1), ..., (i+k-1,j+k-1)\}}$}\\
\>\>\>\> \texttt{$\leftarrow$ Average(x$_{\{(i,j),(i,j+1), ..., (i+k-1,j+k-1)\}}$}\\
\>\> \texttt{return(p)}\\
\\
\>\blue{\# Compare cluster assignments for different kernel sizes}\\
\>\texttt{for kernel size k in 1..9:}\\
\>\> \blue{\# Smooth each patch for kernel size $k$}\\
\>\> \texttt{P$_k$ = \{ smooth(x, k) for x in X$_{\text{holdout}}$ \}}\\
\>\> \blue{\# Encode each smoothed and scrambled patch}\\
\>\> \texttt{Z$_k^{sc}$ = \{ encode(x) for x in P$_k$ \}}\\
\>\> \blue{\# Cluster resulting latent representations}\\
\>\> \texttt{C$_k^{sc}$ = Cluster(P$_k$, \#cluster)}\\
\>\> \blue{\# Determine agreement between C$_1$ and C$_k$}\\
\>\> \texttt{compute AMI(C$_1$, C$_k$)}
\end{tabbing}
\end{singlespace}
\end{minipage}
\begin{minipage}[t]{0.35\textwidth}
\includegraphics[width=\linewidth,trim=198mm 57mm 53mm 135mm,clip]{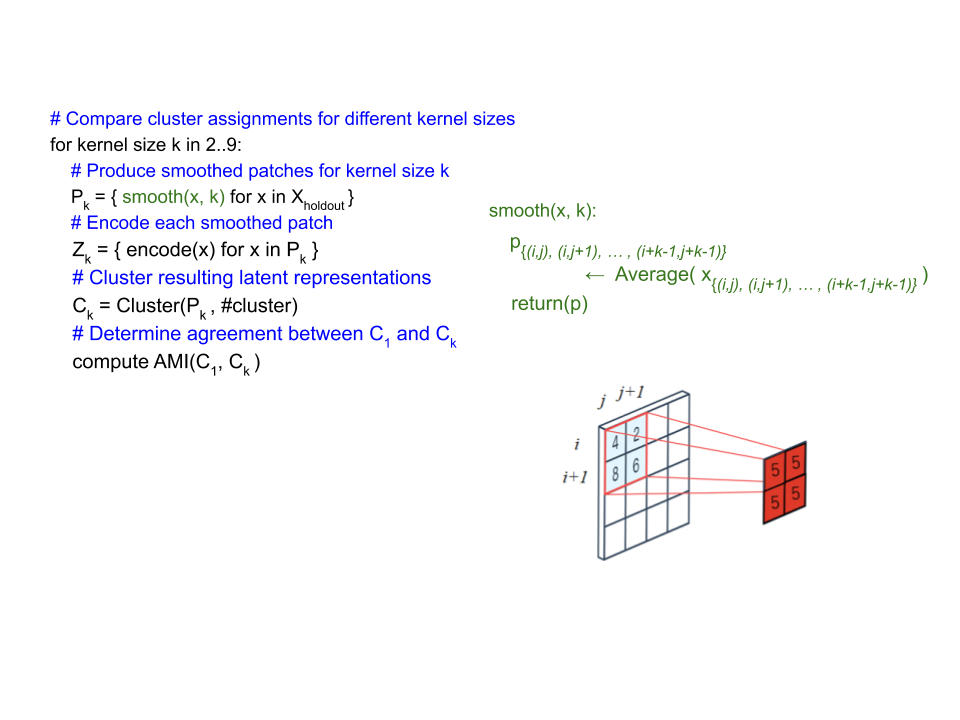}
\end{minipage}
\vspace{1ex}
\hrule
\vspace{-1ex}
    \caption{Pseudocode for the smoothing protocol. The \texttt{smooth} function computes an average over a local window $k \times k$, as depicted in the figure.
    }
    \label{fig:smooth}
\end{figure}

\paragraph{Smoothing test implementation}
Fig.~\ref{fig:smooth} provides the pseudocode for the smoothing test. 
We use the \texttt{Test} dataset described in Section~\ref{sec:testdata}, 
which contains \num{2000} holdout patches unseen during training.
Let $k$ be the kernel size used to convolve the image with a boxcar filter to produce the smoothed version.
The smoothing process computes an average value for each pixel region in a patch.
For instance, for $k = 2$, it sets each pixel
at $(i,j)$, $(i+1,j)$, $(i,j+1)$ and $(i+1,j+1)$, 
for $i \in \{1, 3, \cdots N-1\}$ and $j \in \{1, 3, \cdots N-1\}$, where $N$ is the patch size, to be the average value of those pixels,
leaving any remaining border pixels 
(e.g., if $N=5$ and $k=2$, those with $i=5$ or $j=5$) unchanged.

Let $P_k$ be the patches obtained when each of
the holdout patches $X_{\text{holdout}}$ is smoothed over $k \times k$ pixels.
We then encode each smoothed patch in $P_k$ with the trained encoder, giving $Z_k$ as the corresponding set of latent representations, and apply HAC to those latent representations to obtain a set of 12 clusters $C_k$.
\Erase{(We work with 12 clusters here as we have found that AMI scores typically stabilize
for 10 or more clusters.)}
Lower AMI scores then indicate that the autoencoder has mapped different spatial structures within the input images into different latent representations.
Finally, we determine agreement between $C_1$, obtained from the unsmoothed $X_{\text{holdout}}$, and $C_k$ by computing AMI$(C_1, C_k)$.
We perform this process for each $k\in\{2,\cdots,9\}$.

\begin{figure}[htbp]
    \centering
\hrule
\vspace{1ex}
\begin{minipage}[c]{0.7\textwidth}
\begin{singlespace}
\begin{tabbing}
\=aaa\=aaa\=aaa\=aaa\=\kill
\> \blue{\# Scramble a single image}\\ 
\> \texttt{scramble(x):}\\
\>\> \texttt{indicesA = [(1,1), (1,2), ...,}\\
\>\>\>\> \texttt{(height(x), width(x))]}\\
\>\> \texttt{indicesB = random\_shuffle(indicesA)}\\
\>\> \texttt{for (i,j),(m,l) in (indicesA,indicesB):}\\
\>\>\> \texttt{p$_{\text{(i,j)}}$ = x$_{\text{(l,m)}}$}\\
\>\>\> \texttt{p$_{\text{(l,m)}}$ = x$_{\text{(i,j)}}$}\\
\>\> \texttt{return(p)}\\
\\
\>\blue{\# Compare cluster assignments for different kernel sizes}\\
\>\texttt{for kernel size k in 1..9:}\\
\>\> \blue{\# Smooth each patch for kernel size $k$: see Fig.~\ref{fig:smooth}}\\
\>\> \texttt{P$_k$ = \{ smooth(x, k) for x in X$_{\text{holdout}}$ \}}\\
\>\> \blue{\# Scramble each smoothed patch}\\
\>\> \texttt{P$_k^{sc}$ = \{ scramble(x) for x in P$_k$ \}}\\
\>\> \blue{\# Encode each smoothed and scrambled patch}\\
\>\> \texttt{Z$_k^{sc}$ = \{ encode(x) for x in P$_k^{sc}$ \}}\\
\>\> \blue{\# Cluster resulting latent representations}\\
\>\> \texttt{C$_k^{sc}$ = Cluster(Z$_k^{sc}$, \#cluster)}\\
\>\> \blue{\# Determine agreement between C$_1$ and C$_k^{sc}$}\\
\>\> \texttt{compute AMI(C$_1$, C$_k^{sc}$)}
\end{tabbing}
\end{singlespace}
\end{minipage}
\begin{minipage}[t]{0.3\textwidth}
\includegraphics[width=\linewidth,trim=202mm 5mm 38mm 152mm,clip]{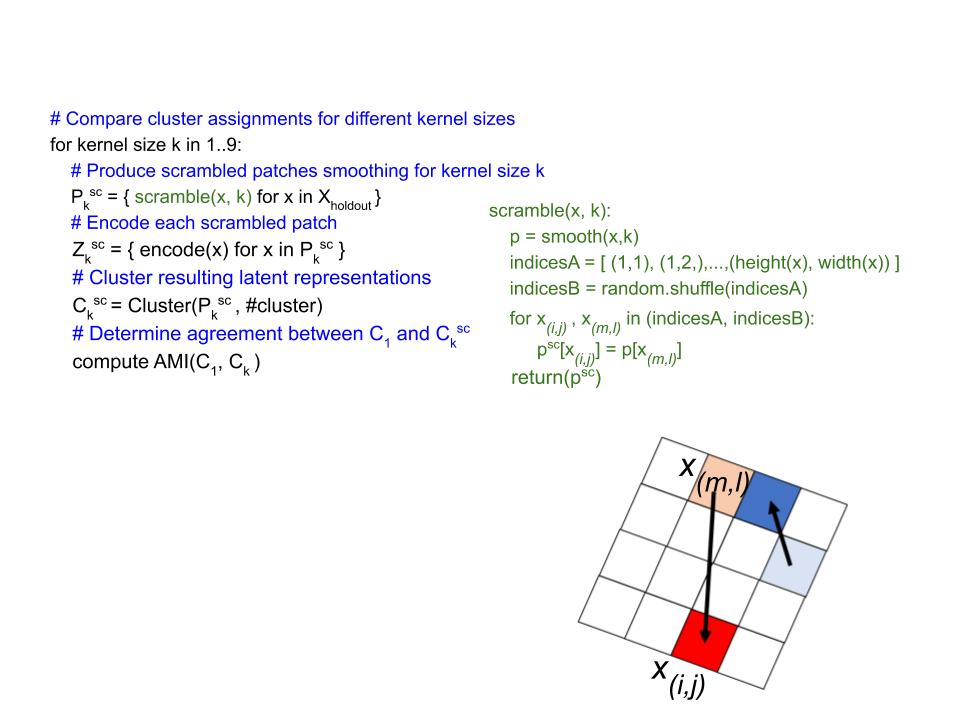}
\end{minipage}
\vspace{1ex}
\hrule
\vspace{-1ex}
    \caption{Pseudocode for the scrambling protocol. 
    The scramble function shuffles the values of randomly selected pixel pairs, %
    as depicted in the figure.
    }
    \label{fig:scramble}
\end{figure}

\paragraph{Scrambling test implementation}
Fig.~\ref{fig:scramble} presents the scrambling test protocol in pseudocode form.
As with the Smoothing Test, we compare the cluster assignments that result when the autoencoder is trained 
with different parameters and 
on images that are smoothed with different kernel sizes, $k\in\{1,2,\cdots,9\}$.
However, in this test, we also scramble the pixels of each image after smoothing
by applying a random permutation to the smoothed pixels,
a process that we repeat (with the same random permutation) for each channel.
%

Again, suppose that we use holdout patches $X_{\text{holdout}}$ in evaluating the smoothing protocol. 
For each kernel size $k\in\{1,2,\cdots,9\}$, 
we smooth each image in $X_{\text{holdout}}$ by $k$
and then scramble its pixels,
giving $P_k$.
We then compute the latent representation on $Z^{sc}_k$ by encoding $P^{sc}_k$ with a trained encoder,
and cluster the latent representations obtained for the different images to obtain clusters $C^{sc}_k$.
We assess the agreement of $C_k$ and $C^{sc}_k$
by computing AMI$(C_k, C^{sc}_k)$.


\subsection{Criterion 3: Separable Clusters}\label{sec:test:separable}
We expect the clusters produced by an unsupervised clustering method to be \emph{separable}:
(i.e., both cohesive in latent space and separated from each other).
We define one test, with results in Section~\ref{sec:results:sep}.

\textbf{Test \ref{test31}: Spatial organization.}\mylabel{test31}{3}
We use t-distributed Stochastic Neighbor Embedding (t-SNE)~\cite{Maaten2008VisualizingDU} (see Appendix~\ref{sec:tSNE} for formulation)
to examine the spatial organization of the latent representation when projected onto a two-dimensional map, and thus to determine whether similar (dissimilar) clusters, as defined by the physical parameter distributions, are projected to closer (more distant) locations in the embedding space.


\subsection{Criterion 4: Rotation Invariance}\label{sec:test:rotate}

We expect the classes produced by an unsupervised clustering method to be \emph{rotationally invariant},
(i.e., to place a cloud image into the same cluster regardless of its orientation).
regardless of the orientation of the image in which they appear. 
We define two tests of increasing difficulty, with results presented in Section~\ref{sec:spatialinfotest}.

\textbf{Test 4.1: MNIST rotation-invariance.} We evaluate rotation-invariance on simple images containing well-defined structures, namely, the MNIST database~\cite{deng2012mnist}.\mylabel{test41}{4.1}


\textbf{Test 4.2: Multi-cluster rotation-invariance.} We evaluate whether our RI autoencoder groups similar types of clouds into the same cluster regardless of image orientation.\mylabel{test42}{4.2}

\paragraph{MNIST rotation-invariance}
This test evaluates rotation-invariance on simple images containing well-defined structures: specifically, on images from the MNIST database~\cite{deng2012mnist}.
This database of \num{70000} images of handwritten digits, 
partitioned into an \num{60000}-image training set and a \num{10000}-image test set,
is commonly used for training image processing systems~\cite{deng2012mnist}.
Each image is a 28$\times$28 array of 8-bit gray-scale levels, 
plus an associated integer label.

We train the autoencoder on the MNIST training set,
apply the trained autoencoder to images from the test set,
and evaluate whether the resulting outputs
reproduce the input images with mapping to a single canonical orientation.
Specifically, we sample 40 images for every digit class,
create four replicas of each image, 
and rotate each of the resulting 10$\times$40$\times$4=1600 images
at random.
We then verify whether
the 40$\times$4=160 outputs for each digit class map to a consistent orientation.

\paragraph{Multi-cluster rotation-invariance}
We use the 2000 \texttt{Test} patches (see Section~\ref{sec:testdata}) not considered during training as holdout patches. 
We make 11 copies of each patch in this set. 
We rotate every 30$^{\circ}$; thus, the ideal result should return the same cluster label for both the original patches and the rotated copies.
We then implement HAC clustering for from \num{4} to \num{2000} clusters. 
The AMI score should be close to \num{1} for \num{2000} clusters, because our \texttt{Test} dataset has \num{2000} patches, meaning that each patch and its replications can be placed in a unique cluster.

\subsection{Criterion 5: Stable Clusters}

We expect the clusters produced by an unsupervised clustering method to be \emph{stable}, (i.e., to be identical or at least similar to those produced) when the method is trained on different subsets of the training data.
Cluster stability is a difficult problem,
and there is not yet agreement in the community as to how to define it~\cite{ben2006sober,von2010clustering}.
Thus, we do not consider this criterion in this study.

\section{Alternate Autoencoders}\label{sec:alternates}
We compare our RI autoencoder with two alternative autoencoders, each with a different loss function: non-rotation-invariant (\emph{NRI}), and rotation aware (\emph{RA}), described in detail in Appendices~\ref{sec:CNRI} and \ref{sec:RA}.

The NRI autoencoder has the network architecture of Fig.~\ref{fig:cnri},
and Equation~(\ref{eq:cnri}) as its loss function.
Minimizing 
Equation~(\ref{standardloss}) with $p=1,2$, the encoder embeds spatial features, (i.e., the structure and orientation of input images), into the latent representation,
from which the decoder approximates the original inputs.
However, because the loss function does not reward the separate capture of structure and orientation, the  
autoencoder 
produces different latent representations for images that differ only in their rotation.

\begin{figure}[btph]
    \begin{center}
\includegraphics[width=\columnwidth,trim=1mm 1mm 1mm 1mm,clip]{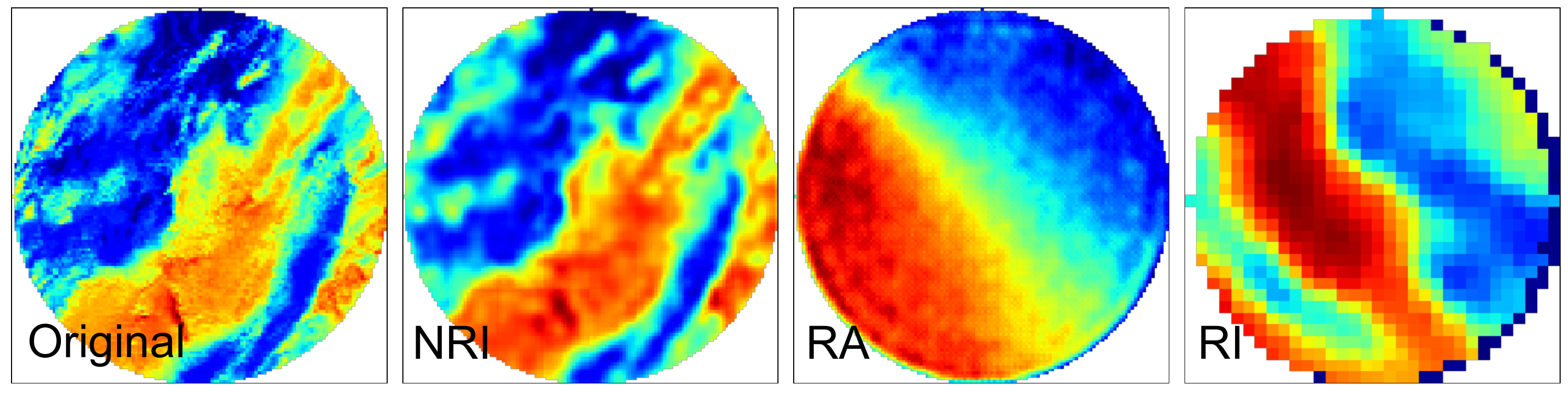}
    \end{center}
    \vspace{-2ex}
    \caption{Example results obtained when autoencoders are applied to a cloud image. \Addafter{We visualize one band (band 6) \AddAdd{across the six bands used}.} From left to right: Original image with circular mask, and results when processed with NRI, RA, and RI autoencoders.
    The NRI autoencoder does not produce a canonical rotation and the RA autoencoder does not preserve patterns in the original; the RI autoencoder does both.
    }
    \label{fig:collapse}
\end{figure}

The RA autoencoder has the network architecture of Fig.~\ref{fig:network:ricc}.
Its loss function, Equation~(\ref{eq:RA}), addresses the rotation-invariance problem, but produces reconstructed cloud images with low fidelity (see Fig.~\ref{fig:collapse}).
The RA autoencoder optimizes simultaneously for both to learn spatial patterns and to produce identical latent representations for an image and various rotations of that image.
The introduction of the RA autoencoder allows us to compare latent representations produced by autoencoders with and without learning better spatial patterns.    

\section{Evaluation Results}\label{sec:evals}
We now evaluate the performance of autoencoders from the perspective of the criteria C1--C4 introduced in Section~\ref{sec:protocols}.
\AddAdd{We now evaluate the performance of the RI, NRI, and RA autoencoders from the perspective of the criteria C1–C4 introduced in Section~\ref{sec:protocols}.
The results, summarized in Table~\ref{tab:autoencoders}, allow us both to evaluate our RICC (RI autoencoder plus HAC clustering) method in absolute terms (we show that it satisfies all four criteria) and to understand the importance of the rotation invariant term in the loss function
(the NRI autoencoder, which lacks that term, fails the tests defined for criterion C4)
and of a latent representation that preserves spatial structure
(the RA autoencoder, which includes the rotation invariant term but does a poor job of preserving spatial structure, fails test~\ref{test42}).}

\begin{table*}
\centering
\caption{The three autoencoders that we consider in this paper and the results obtained when we evaluate each with the seven tests that we defined in Section~\ref{sec:protocols} to assess four of the five criteria introduced in Section~\ref{sec:intro}.
}\label{tab:autoencoders}
\begin{tabular}{|p{1.6cm}| p{3.2cm} | p{1.5cm} | p{4.2cm} | C{1.5cm} | C{1.5cm}| C{1.5cm}|} 
\hline
 \multicolumn{4}{|c|}{} & \multicolumn{3}{c|}{\textbf{Autoencoder}}   \\ 
\multicolumn{4}{|c|}{}  &   \textbf{NRI} & \textbf{RA} & \textbf{RI} \\ \hline\hline

\multicolumn{4}{|r|}{\textbf{Network shown in:}}   & Fig.~\ref{fig:cnri} & Fig.~\ref{fig:network:ricc} & Fig.~\ref{fig:network:ricc}\\ \hline
\multicolumn{4}{|r|}{\multirow{1}{2.2cm}{\\\textbf{Loss function:}}} 
    & $L_1 + L_2 +  L_{\text{high-pass}} + L_\text{MS-SSIM}$ 
    & \multirow{1}{2cm}{\centering $L_{\textrm{\textrm{agn}}}$\\ +\\ $\lambda \cdot  L_{\textrm{\textrm{inv}}'}$}  
    & \multirow{3}{1.8cm}{\centering $\lambda_{ \text{inv}} L_{\text{inv}} + \lambda_{ \text{res}} L_{\text{res}}$} \\ \hline
\multicolumn{4}{|r|}{\textbf{Equation reference in text:}}  
            &  (\ref{eq:cnri})
            &  (\ref{eq:RA})
            &  (\ref{eq:ri_loss})\\ \hline\hline

\textbf{Criterion} & \textbf{Test} & \AddAdd{\textbf{Test data}} & \textbf{Pass/Fail \AddAdd{requirement}} & \multicolumn{3}{c|}{\textbf{Passed: Yes/No}} \\ \hline
Physically reasonable
    & \ref{test11}: Cloud physics \hspace{5ex} parameters & \Add{\texttt{Test}- \texttt{CUMULO}}, \Add{\texttt{Phys}} & Non-random dist.; median inter-cluster correlation $<$ 0.6&  \multirow{2}{*}{Y} & \multirow{2}{*}{Y} & \multirow{2}{*}{Y} \\ \hline
   
\multirow{3}{2.5cm}{Spatial distribution}
    & \ref{test21}: Spatial coherence & \Add{\texttt{Phys}} & Spatially coherent clusters 
                                    & Y & Y & Y \\ \cline{2-7}
    & \ref{test22}: Smoothing  & \Add{\texttt{Phys}} & Low AMI score & Y & N & Y  \\ \cline{2-7}
    & \ref{test23}: Scrambling & \Add{\texttt{Phys}} & Low AMI score & Y & N & Y \\
\hline
Separable   & \ref{test31}: Separable clusters & \Add{\texttt{Phys}} & No crowding structure & Y & Y & Y \\ 
\hline
\multirow{2}{2.5cm}{Rotationally invariant} & \ref{test41}: MNIST images & \Add{\texttt{MNIST}} & Single canonical orientation & N & Y & Y  \\ \cline{2-7} 
 & \ref{test42}: Multi-cluster &  \Add{\texttt{Test}}  & AMI score 
                closer to \num{1.0} &  N & N & Y \\ \hline

\end{tabular}
\end{table*}

\subsection{Results for Criterion 1: Physical Reasonableness}\label{sec:physicalreasonableness}
\begin{figure*}
    \centering
     \includegraphics[width=\linewidth,]{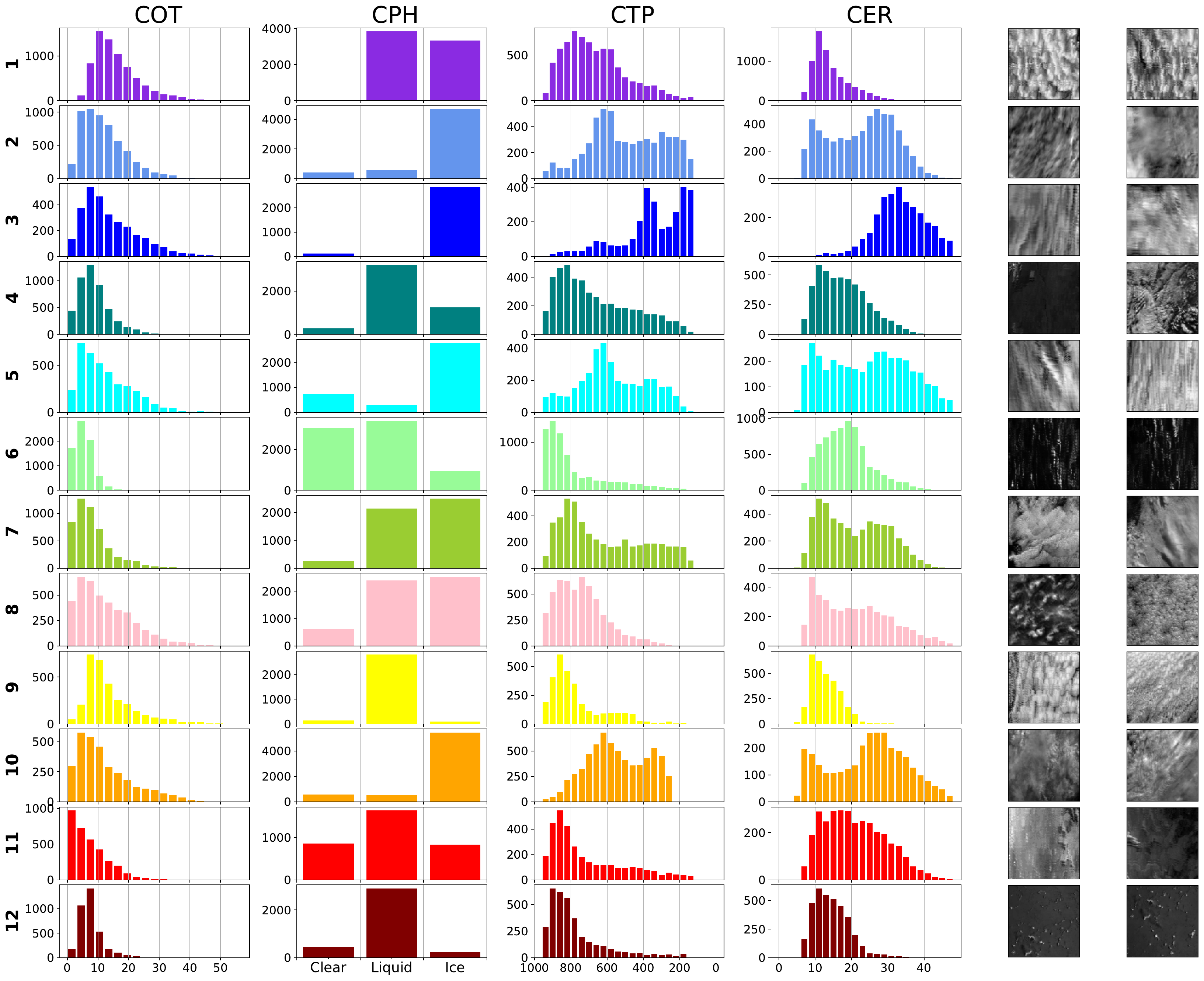}
    \caption{\emph{Test~\ref{test11}: Cloud physics parameters, applied to RI autoencoder}. The first four columns show histograms of the patch-mean values of four derived cloud physical parameters---COT (no unit), CPH (clear sky, liquid, or ice), CTP (hPa), and CER ($\mu$m)---for each of the 12 clusters produced by the RI autoencoder for the \texttt{TestCUMULO} dataset. The last two columns show, for the two patches closest to the centroid, the raw visible image (band 5), with white to grey indicating cloud and black either ocean or land without cloud.
    We make the following observations about these results.
    1) The distributions of each parameter \emph{within} clusters do not contradict cloud physics (i.e., they are not random) and furthermore show distinct differences \emph{across} clusters, a result supported quantitatively by median inter-cluster correlation coefficients, as discussed in the text. 
    2) In clusters for which the COT histogram shows larger values, the sample images are cloudier, as we would expect.
    3) Clusters \#7 and \#8 both have roughly similar proportions of liquid and ice phases, but differ in COT and CTP, which in \#8 are wider and narrower, respectively, than in \#7. It seems likely that \#7 is a mixture of two cloud types--low altitude liquid and high altitude ice crystal clouds---while \#8 aggregates more than two cloud types.
    }
    \label{fig:physicalreasonable}
\end{figure*}
\begin{figure*}
    \centering
     \includegraphics[width=\linewidth,]{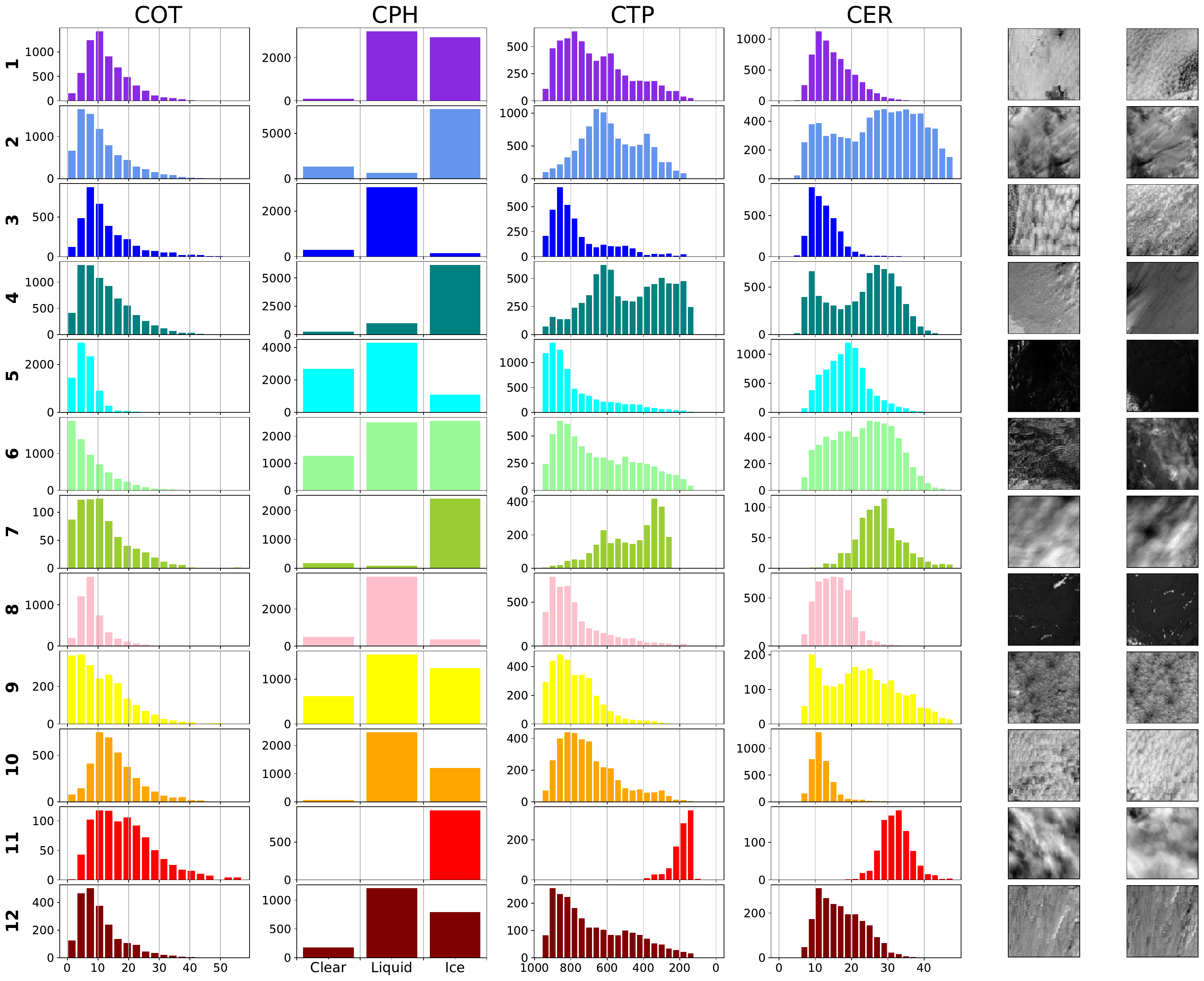}
    \caption{\Addafter{\emph{Test~\ref{test11}: cloud physics parameters, applied to NRI autoencoder}. As in Fig. \ref{fig:physicalreasonable} but for NRICC.
    We make the following observations about these results.
    1) The distributions of each parameter \emph{within} clusters do not contradict cloud physics and \emph{across} clusters show distinct differences supported by Table~\ref{tab:stats:physicalreasonable}.
    2) \AddAdd{Similar to} RICC in Fig. \ref{fig:physicalreasonable}, cloudier patches associate with larger values of COT. 
    3) Compared to two patches from RICC in Fig. \ref{fig:physicalreasonable} closest to the centroid, it seems that the example two patches (clusters \#2, \#7, \#9, \#10, \#11, \#12) from NRICC are more likely to show identical orientation, indicating a NRICC exhibits rotation dependence.}}
    \label{fig:nri_physicalreasonable}
\end{figure*}

\begin{table}
\centering
\caption{Test~\ref{test11}: Median of correlation coefficients for four cloud physics parameters and for RI, RA, and NRA autoencoders. Correlation coefficients $<$0.6 in boldface. AE=autoencoder.}\label{tab:stats:physicalreasonable}
\centering
\begin{tabular}{|p{0.6cm}| C{2.6cm}|l*{3}{C{.7cm}}|}
\hline
 AE & {Parameter(s)} & COT &  CPH & CTP & CER \\ \hline\hline
 RI & $\lambda_{\text{inv}}, \lambda_{\text{res}} = 32,80$ & 0.842 & \textbf{0.591} & \textbf{0.461} & 0.692\\ 
 RA& $\lambda = 1 $ & 0.835 & \textbf{0.588} & \textbf{0.438} & 0.661\\ 
 NRI& -- & 0.829 & 0.629 & \textbf{0.425} & \textbf{0.484}  \\  \hline
\end{tabular}
\end{table}

\begin{figure*}
    \centering
    \begin{minipage}{.49\columnwidth}
        \subfloat[NRICC.\label{fig:dendrogram-nri}]{\includegraphics[width=0.96\textwidth,]{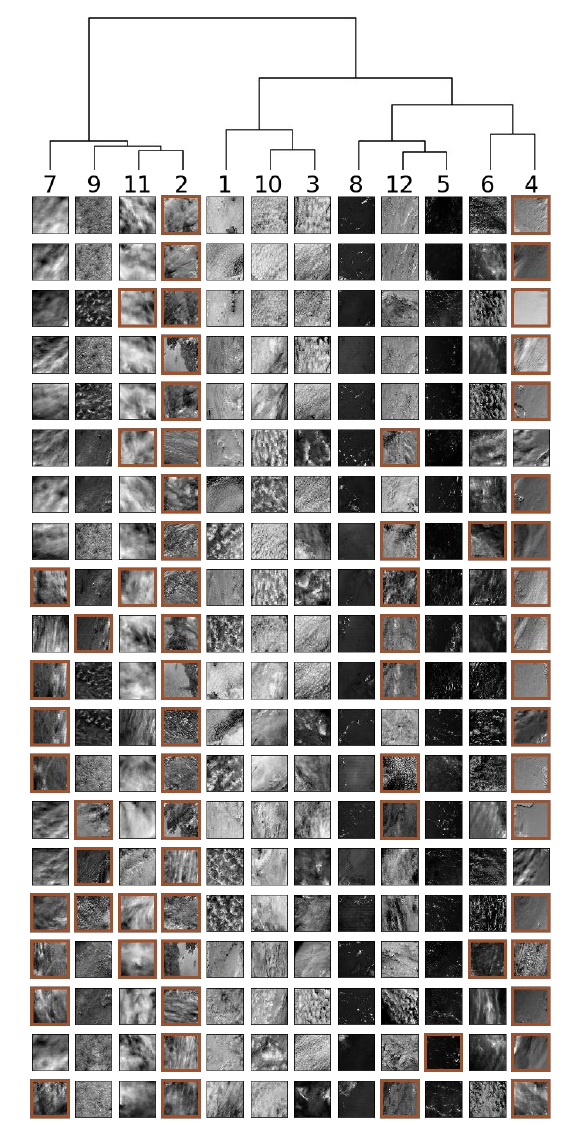}}
    \end{minipage}
    \begin{minipage}{.49\columnwidth}
        \subfloat[RICC.\label{fig:dendrogram}]{\includegraphics[width=0.96\textwidth,]{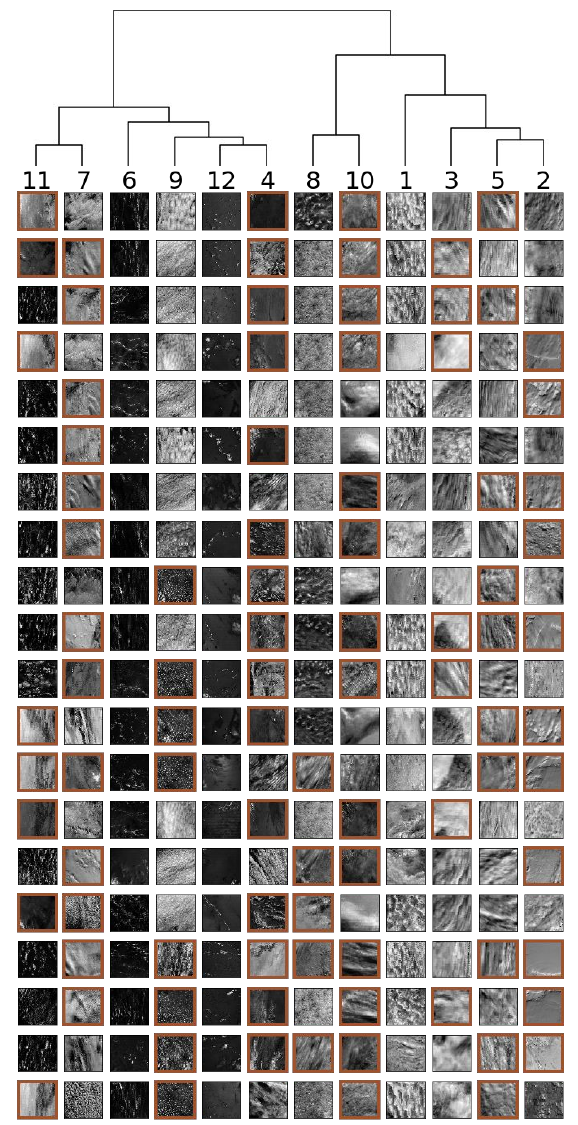}}
    \end{minipage}
    \caption{
    \Addafter{Dendrogram representation of the HAC tree structure generated by (a) NRICC and (b) RICC} for the \texttt{TestCUMULO} dataset of January 1, 2008, with 12 clusters. Cluster numbers are shown at the dendrogram leaf nodes; the column below each number shows MOD02 raw visible (band 5) images for the 20 patches closest to the centroid of that column's cluster, 
    ordered according to their distance to the centroid, with the topmost patch being the closest.
    \Addafter{(Note that these clusters of (a) NRICC are the same as in Fig.~\ref{fig:nri_physicalreasonable}, and 
    clusters of (b) RICC are Fig.~\ref{fig:physicalreasonable}. 
    The top two patches for each cluster are identical to those shown in the last two columns of Fig.~\ref{fig:nri_physicalreasonable} and Fig.~\ref{fig:physicalreasonable} respectively.)}
    White-to-grey colors indicate cloud pixels; black denotes a non-cloud (ocean or land) pixel; and patches with a brown frame contain land clouds.
    We see that RICC clusters differentiate between ocean clouds (cluster \#6 is 96.6\% ocean cloud patches) and land clouds (cluster \#10 is 61.0\% land cloud patches), a differentiation based on texture and radiances that may capture different cloud formation processes.
    \Addafter{We observe the identical bias of distribution of ocean clouds (cluster \#5 is 95.0\% ocean cloud patches) and land clouds (cluster \#4 is 62.6\% land cloud patches) in NRICC clusters.}
    Furthermore, the example patches from RICC shown for the different clusters show evidence of rotation invariance, of grouping based on similarity of texture, and
    distinct differences in the density of cloud pixels and in circular/uniform/streak structure.
    Comparison with Fig.~\ref{fig:physicalreasonable} shows that clusters with sparse clouds (\#6, \#12) have a lower standard deviation in their optical thickness than do clusters with homogeneous texture (\#1, \#2, \#3, \#5).
    \Addafter{Clusters from NRICC also show the morphological similarity of texture, and distinct difference \AddAdd{in} density of cloud pixels which correspond \AddAdd{to} distributions of optical thickness in Fig.~\ref{fig:nri_physicalreasonable}.
    However, the example patches from clusters \#7 and \#11 from NRICC indicate evidence of rotation dependence based on the orientation of \AddAdd{textures.} Both clusters capture thick (tail distribution of optical thickness), relatively larger size of ice-crystal clouds (higher mean of effective radius and dominant ice phase) at high altitudes (peak of top pressure $>$ 400 hPa), but patches from \#7 tend to be tilted clockwise, \AddAdd{while} patches from \#11 are rotated anti-clockwise. 
    Results match the illustration in Fig.~\ref{fig:concept-AEs}, and indicate the \AddAdd{limitations} of NRICC optimizing our original loss function in Equation~(\ref{eq:cnri}) and \AddAdd{benefits} of RICC learning our proposed rotation-invariant loss function in Equation~(\ref{eq:ri_loss}).}
    }
\end{figure*}

We apply Test~\ref{test11}, introduced in Section~\ref{sec:phys},
to evaluate the physical reasonableness of the cluster labels assigned by our trained autoencoders. 
Specifically, we investigate whether cloud physical parameters (i.e., the retrieved physical parameters computed by MODIS scientists from cloud properties), are both consistent within clusters and distinct among clusters. We use the largest test dataset, \texttt{TestCUMULO}, for this purpose.
\Add{Although the purpose of Test~\ref{test11} is to account for physical reasonableness, the distinct combinations of physical properties seen in Fig. \ref{fig:physicalreasonable} also \AddAdd{contribute} to our interpretation of the clusters identified by the RICC framework.} 
\Addafter{Fig. \ref{fig:nri_physicalreasonable} also contributes to our interpretation of results clusters from the NRI cloud clustering (NRICC) framework.}

Fig.~\ref{fig:physicalreasonable} shows, for each of the 12 clusters generated by RICC for the \texttt{TestCUMULO} dataset, the distributions of patch-mean values for four derived physical parameters, plus the two closest patches to the cluster's centroid. 
We work with 12 clusters because we have seen previously that when clustering \texttt{Phys}, AMI scores stabilize~\cite{kurihana2019cloud} for cluster counts of 10 and higher.
Examination of these distributions shows that clusters are clearly associated with meaningful physical cloud attributes.
For example, high-altitude cirrus clouds have low cloud top pressure and are composed of ice crystals,
features that are captured in clusters \#3 and \#10, which both have a dominant ice phase, lower peak values for cloud top pressure (CTP), and relatively larger cloud effective radius (CER). 
Cluster \#3 shows similar distributions for cloud optical thickness (COT) and CPH to cluster \#10, 
but differs in CTP (narrower range) and CER (a single, smaller peak),
suggesting subtle physical differences.
As a second example of parameter value distributions having a reasonable association in terms of cloud physics, stratocumulus is low-altitude clouds with liquid droplets and a middle range of optical thickness: features seen in cluster \#9.

As discussed in Section~\ref{sec:phys}, we use the median inter-cluster correlation as a quantitative measure of the diversity of parameter distributions across clusters.
Our results are in Table~\ref{tab:stats:physicalreasonable}.
For the RI autoencoder (first row), this measure has values 0.591 and 0.461 for the CPH and CTP parameters, respectively, 
suggesting that these parameters are differently distributed in different clusters, as we desire.
For the RA and NRI autoencoders (rows 2 and 3), this measure is also below 0.6 in the case of the CTP parameter, with values of 0.438 and 0.425, respectively.
In summary, we conclude that all three autoencoders, while trained on data that contain only information about cloud textures and radiances, produce clusters that reflect intrinsic differences in physical parameters.

We can also see in Fig.~\ref{fig:physicalreasonable} a clear association between COT distributions and displayed patches.
In clusters \#1, \#2, and \#3, the displayed patches show dense clouds (i.e., dominant white and grey pixels) and COT has a long tail distribution;
in clusters \#6 and \#12, the displayed patches show sparse clouds (i.e., mostly black pixels) and COT has a narrow distribution and smaller peak values. 
\Addafter{We can see in Fig. \ref{fig:nri_physicalreasonable} the association between COT distributions and dense/sparse clouds on patches from clusters produced by NRICC.
Cloudier cloud patches in clusters \#1, \#2, \#3, \#10, and \#11 show tail distribution in COT; 
in clusters \#5 and \#8, sparse cloud patches show \AddAdd{a narrow} distribution peaked at low COT values.}

To examine the cluster-texture associations in more detail, we show in Fig.~\ref{fig:dendrogram-nri} and Fig.~\ref{fig:dendrogram} a hierarchical \emph{dendrogram}, created by merging pairs of clusters that are close in the latent space. 
In the resulting graph, the height of each node is proportional to the intergroup dissimilarity between its two offspring subgraphs or leaf nodes.
We also show, below each cluster number, images for the 20 patches that are closest to that cluster's centroid.
We find in Fig. \ref{fig:dendrogram} that clusters from RICC with dense cloud textures (\#1, \#2, \#3, \#5) also have large COT values with large standard deviations, while
clusters \#6 and \#12 again show lower COT peak values and standard deviation; their example patches are primarily sparse clouds over the ocean.
\Addafter{Similarly, we find in Fig. \ref{fig:dendrogram-nri} that NRICC demonstrates the association between cloud texture and distributions of COT observed in RICC above;
clusters with sparse texture clouds (clusters \#5 and \#8) show lower COT values with narrower distribution;
clusters with dense cloud textures (\#1, \#2, \#7, \#9, \#10, and \#11) have larger COT values with larger standard deviations.}
\Add{The dendrogram highlights both inter- and intra-cluster similarities and dissimilarities. We observe that patches from the same cluster, and to a lesser extent those from nearby clusters, show evidence of rotation-invariance and grouping based on similarity of \AddAdd{texture.} Patches from more distant clusters show distinct differences in the density of cloud pixels and in circular/uniform/streak structures.}

Interestingly, while clusters \#7 and \#8 from RICC have similar distributions in Fig.~\ref{fig:physicalreasonable}, they are some distance apart in Fig.~\ref{fig:dendrogram} and show different textures in the example images.
To examine this phenomenon further, we computed the percentage of ice and liquid pixels within each patch in the two clusters. 
We find that while cluster \#8 has a broad range [0\%--100\%] of ice and liquid percentage, in cluster \#7 most patches have $>$80\% of either ice or liquid pixels, suggesting that cluster \#8 amalgamates multiple cloud types with different physical and texture features.

\Addafter{We further compare the resultant dendrogram from RICC and NRICC to inspect the impact on the cluster-texture association of clustering assignments without rotation-invariant term.
The two example patches \AddAdd{closest to their centroids} in Fig. ~\ref{fig:nri_physicalreasonable} suggest that they are more likely to show identical orientation than example patches from RICC.}
\AddAdd{The NRICC dendrogram suggests, by comparison of the twenty patches closest to their centroids, that NRICC shows rotation dependence.
In particular, cluster \#7 depicts patches tilted clockwise, and cluster \#11 has patches rotated anti-clockwise.}
\Addafter{We further investigate the cluster-rotation dependency in Test~\ref{test42}.
}

\subsection{Results for Criterion 2\AddAdd{:} Spatial Distribution}\label{sec:spatialinfotest}

\begin{figure*}
    \centering
    \includegraphics[width=\textwidth,trim=0mm 28.mm 0mm 32.5mm,clip]{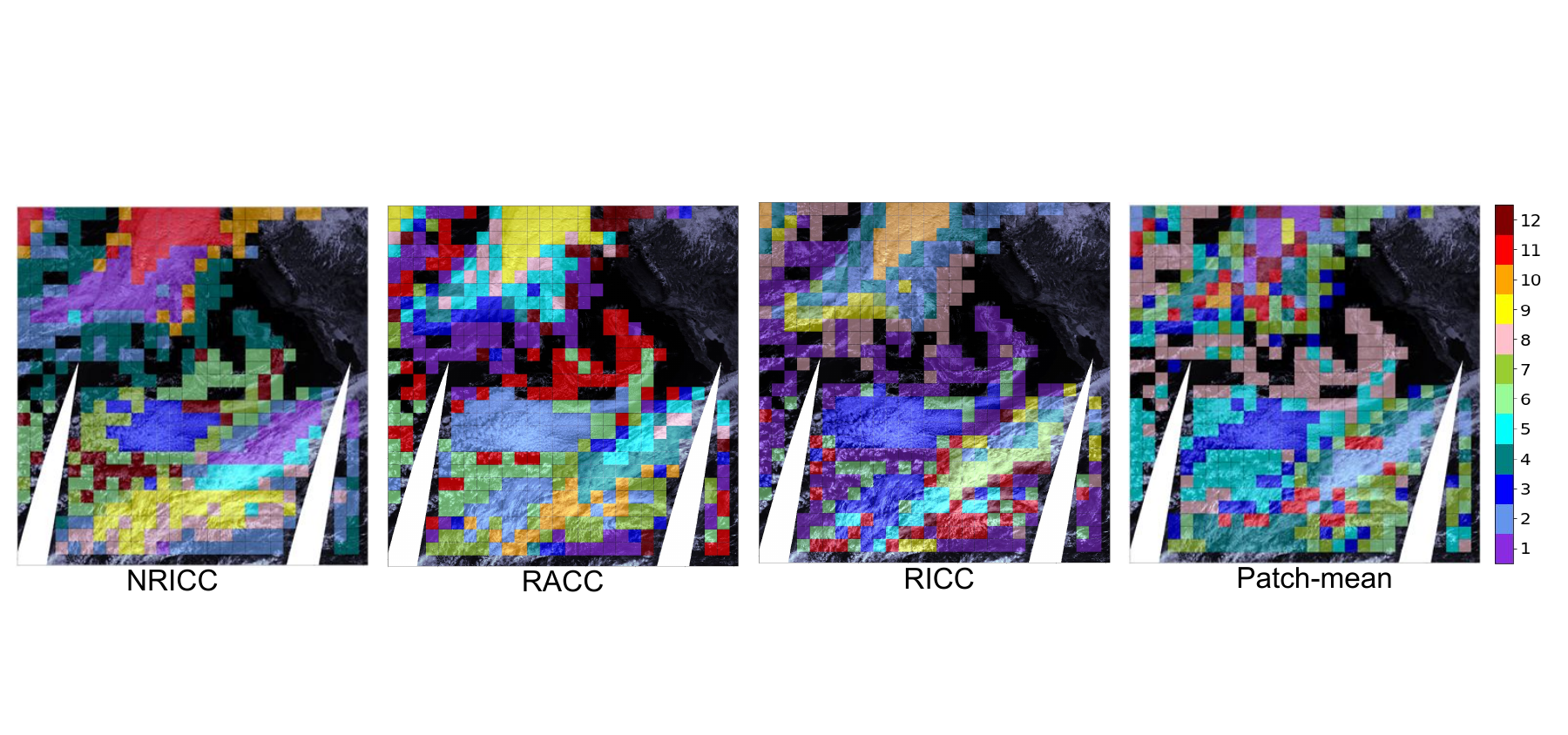}
    \caption{\emph{Test~\ref{test21}: Spatial coherence, applied to NRI, RA, and RI autoencoders}.
    On the same MODIS swath image (\texttt{Phys}, with 493 patches), we show:
    Left, center: Clusters produced by NRICC, RACC, and RICC, respectively; 
    right: Clusters produced by HAC applied to patch-mean values of COT, CTP, CWP, and CER cloud physical parameters.
    The background raw visible image (band 1) is provided for context only; it is not used in training the autoencoders.
    In each case, 12 clusters are produced.
    The color bar shows cluster number; white means no data due to orbital coverage gaps or invalid data, and black means patches with $<30\%$ cloud pixels. All three autoencoder-based approaches produce spatially coherent cluster assignments.}
    \label{fig:coherence}
\end{figure*}

We apply the three tests described in Section~\ref{sec:test:spatial} to the NRICC, RA cloud clustering (RACC), and RICC frameworks.

\paragraph{Results for Test \ref{test21}\AddAdd{:} Spatial Coherence}
We use the 493-patch MODIS swath \texttt{Phys} to investigate spatial coherence.
We show in Fig.~\ref{fig:coherence} four different clusterings of this dataset: 
three with (left, centers) and one without (right) an autoencoder.
We see that all three autoencoder-based approaches
(none of which is aware of patch locations in the swath)
produce more spatially cohesive cluster assignments than does a clustering based only on mean physics parameters.

The \texttt{Phys} dataset used here to verify spatial coherence is disjoint from the \texttt{TestCUMULO} dataset used in Section~\ref{sec:physicalreasonableness} to verify physical reasonableness. Thus, as a further test, 
reported in Appendix~\ref{sec:appned_c1physdata}, we also verify that the clusters produced by RICC for \texttt{Phys} are physically reasonable.

\paragraph{Results for Test \ref{test22}\AddAdd{:} Spatial Resolution}
This test examines whether and how cluster assignments change when we alter the spatial resolution of images via smoothing.
Table~\ref{tab:spacialtest} summarizes the results obtained. 
Each row in the table shows the results for a different autoencoder,
while the columns correspond to different kernel sizes. 
The notation tells that the horizontal resolution of the inputs for the RI autoencoder is 32$\times$32 and that of inputs for the RA and NRI autoencoders are 128$\times$128, so that we need to apply different kernel sizes to smooth the same percentage of pixels in the images.
We see that the best RI autoencoder, 
obtained with $\lambda_{\text{res}}$~=~80, $\lambda_{\text{inv}}$~=~32,
achieves the lowest agreement score, 0.577,
showing that it is indeed capturing spatial patterns.
\AddAdd{The agreement scores of the NRI autoencoder are always larger than that of the optimal RI autoencoder,
indicating that the difference can be accounted for with and without rotation-invariance.}

\begin{table*}
\small\centering
\caption{Spatial distribution tests \ref{test22} and \ref{test23}, applied to NRI, RA, and RI autoencoders.
AMI scores are given for different kernel sizes, with the largest value in each column being in boldface.
}\label{tab:spacialtest}
\scriptsize\centering
\begin{tabular}{|p{1.2cm}| C{2cm}|l*{7}{C{.4cm}}|l*{8}{C{.4cm}}|}
\hline
 \multicolumn{2}{|c|}{}   &  \multicolumn{8}{c|}{Test \ref{test22}: Smoothing} &  \multicolumn{9}{c|}{Test \ref{test23}: Scrambling} \\ \hline
 \multirow{2}{*}{{Autoencoder}} & \multirow{2}{*}{Parameter(s)}  &  \multicolumn{17}{c|}{Kernel size}  \\ 
 &  &2(8)&3(12)&4(16)&5(20)&6(24)&7(28)&8(32)&9(36) &1(1)&2(8)&3(12)&4(16)&5(20)&6(24)&7(28)&8(32)&9(36) \\ \hline\hline
 \multirow{3}{*}{RI} & $\lambda_{\text{inv}}, \lambda_{\text{res}} = 32,80$&0.630&\textbf{0.607}&0.664&\textbf{0.604}&\textbf{0.624}&\textbf{0.577}&\textbf{0.580}&\textbf{0.582}&0.636&0.612&\textbf{0.601}&\textbf{0.612}&\textbf{0.584}&\textbf{0.619}&0.600&0.625&\textbf{0.589} \\ 
 & $\lambda_{\text{inv}}, \lambda_{\text{res}} = 0,80$ &0.669&0.665&0.675&0.631&0.661&0.624&0.597&0.611&\textbf{0.631}&0.629&0.637&0.641&0.626&0.678&\textbf{0.585}&\textbf{0.590}&0.603 \\ 
 & $\lambda_{\text{inv}},\lambda_{\text{res}} = 3.2,8$ &\textbf{0.626}&0.658&\textbf{0.640}&0.644&0.654&0.661&0.608& 0.610&0.642&0.624&0.627&0.646&0.608&0.633&\textbf{0.585}&0.606&0.593\\ 
 {RA}& $\lambda = 1 $ &0.743&0.748&0.705&0.648&0.760&0.696&0.727&0.691&0.709&0.704&0.722&0.741&0.661&0.770&0.676&0.758&0.749  \\ 
 {NRI}& -- & 0.645&0.657&0.692&0.658&0.673&0.693&0.674&0.627&0.654&\textbf{0.610}&0.652&0.618&0.637&0.623&0.668&0.702&0.642  \\  \hline
\end{tabular}
\end{table*}

\paragraph{Results for Test \ref{test23}\AddAdd{:} Scrambling}
This text examines how cluster assignments change when we scramble image pixels so as to remove spatial patterns while preserving the distribution of pixel values.
We evaluate clustering agreements for inputs of original and scrambled patches. 
A low agreement score indicates that the trained autoencoder is encoding information about spatial patterns in the latent representation;
a high agreement score shows that it is not.
Table~\ref{tab:spacialtest} shows results.
We see that the RI autoencoder, under the optimal parameter combination, outperforms other autoencoders in learning spatial patterns,
achieving the lowest score, 0.584.
\Addafter{We observe that \AddAdd{the} NRI autoencoder and a suboptimal RI autoencoder ($\lambda_{\text{inv}}$~=0, $\lambda_{\text{res}}$~=~80), both of them are trained without rotation-invariant term, also achieve the lowest agreement score for four kernel sizes in the scrambling test.}
\AddAdd{Results suggest that the autoencoder with our original loss function (i.e., loss function that optimizes only reconstruction term) also results in learning spatial patterns.}

\subsection{Results for Criterion 3: Separable Clusters}\label{sec:results:sep}
\begin{figure}
    \centering
    \includegraphics[width=\textwidth]{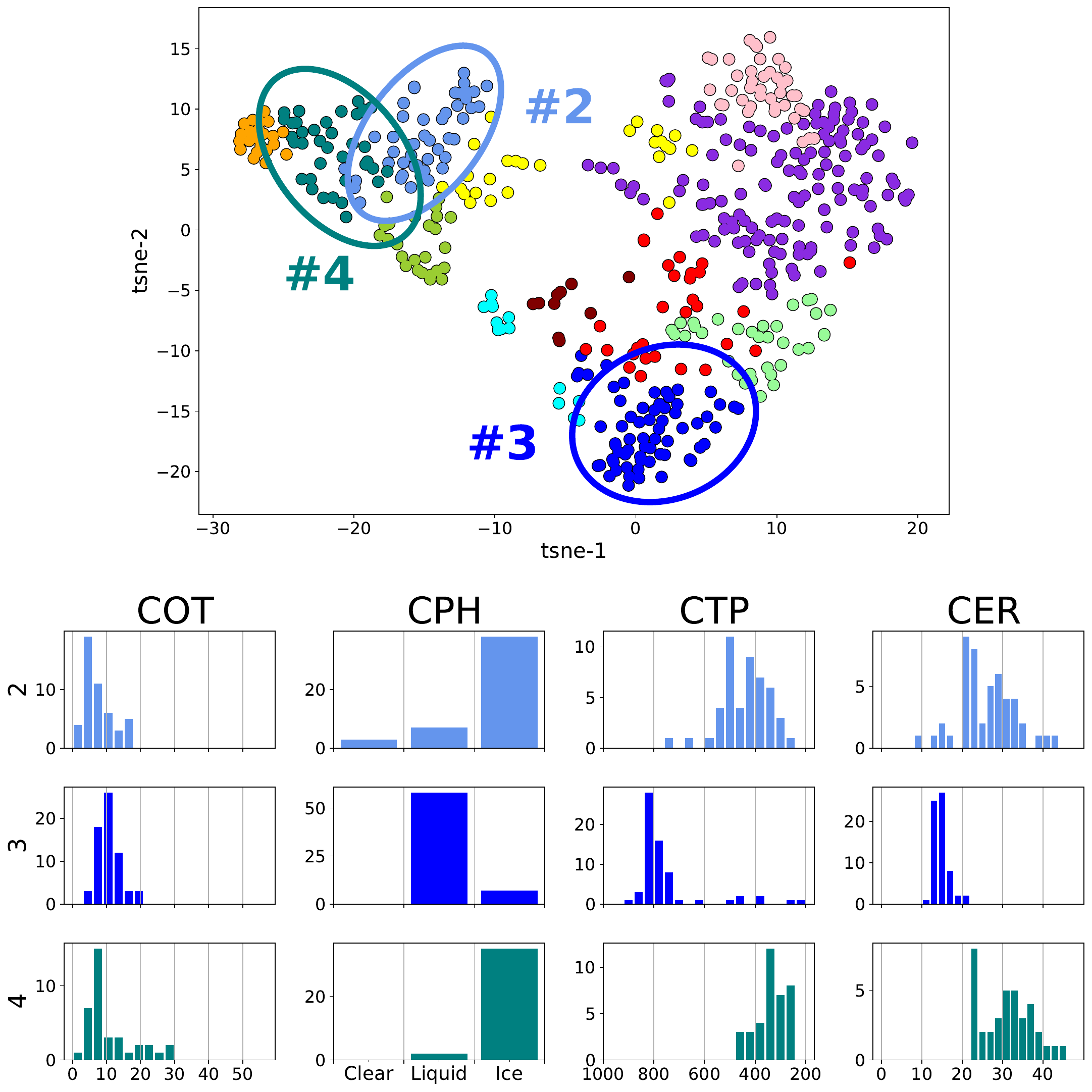}
    \vspace{-2ex}
    \caption{\emph{Test \ref{test31}: spatial organization, applied to RI autoencoder}. Top: t-SNE visualization of the latent representations of \texttt{Test} patches. 
    Patches in each cluster are projected near to each other and distant from patches in other clusters. 
    Bottom: Histograms of path-mean values of four derived cloud physics parameters---optical thickness, phase, cloud top pressure, and effective radius---for the three clusters circled in the top panel.
    (We select these three clusters to highlight the association between distance in 2D tSNE space and physical parameters; see Fig.~\ref{fig:physicalreasonablePhysdata} in Appendix~\ref{sec:appned_c1physdata} for histograms of the other nine clusters.)
    Closer clusters (\#2, \#4) demonstrate similar distributions in their four histograms than do distant clusters (e.g., \#2, \#3), indicating that RICC clusters reflect nonlinear interrelations among selected spectral bands.
    }
    \label{fig:separable}
\end{figure}
We apply Test~\ref{test31}, described in Section~\ref{sec:test:separable}, to RICC,
applying t-SNE to the latent representations produced by the RI autoencoder for the 493 patches in \texttt{Phys}.

Results are in Fig.~\ref{fig:separable}.
The upper panel shows the two-dimensional t-SNE projection,
with each color representing one of the 12 clusters.
We see that the clusters are, for the most part, 
homogeneous and distinct, 
indicating that clustering achieves good separability.
The lower panel provides histograms for four physical properties for clusters \#2, \#3, and \#4. 
We choose to highlight these clusters here because clusters \#2 and \#3 capture the common physical features of high-altitude cirrus and stratocumulus, respectively (demonstrated in Appendix~\ref{sec:appned_c1physdata} in Fig.~\ref{fig:physicalreasonablePhysdata}), 
and \#2 is one of the three clusters closest to \#4.
We see that the two highlighted clusters that are adjacent in the projected map (\#2 and \#4) have similar physical properties, whereas the first highlighted cluster (\#3), which is distant from both, has more dissimilar physical properties.
We observed in previous work a similar trend for the NRI autoencoder on the same dataset~\cite{kurihana2019cloud}, concluding that NRICC and RICC systems can approximate non-linear physical relationships across the selected six bands.

Recall that each cloud physical parameter provided in the MOD06 product is estimated via theoretically based algorithms~\cite{Ackerman2008CloudDW,baum12,steven17} that combine radiance and brightness values from spectral bands that are sensitive to physical features.
For example, bands 6 and 7 correspond to the atmospheric absorption range of H$_2$O, and thus are used for the computation of COT and CER~\cite{baum12}. 
Similarly, the separation of the liquid and ice phases in CPH is computed from the brightnesses in bands 29 and 31, 
due to those bands' sensitivity to the ice phase. 

\begin{figure*}
	\centering
	\begin{minipage}{.33\columnwidth}
		\centering
		\subfloat[NRI autoencoder.\label{fig:mnist-img1}]{
		    \includegraphics[width=.9\textwidth]{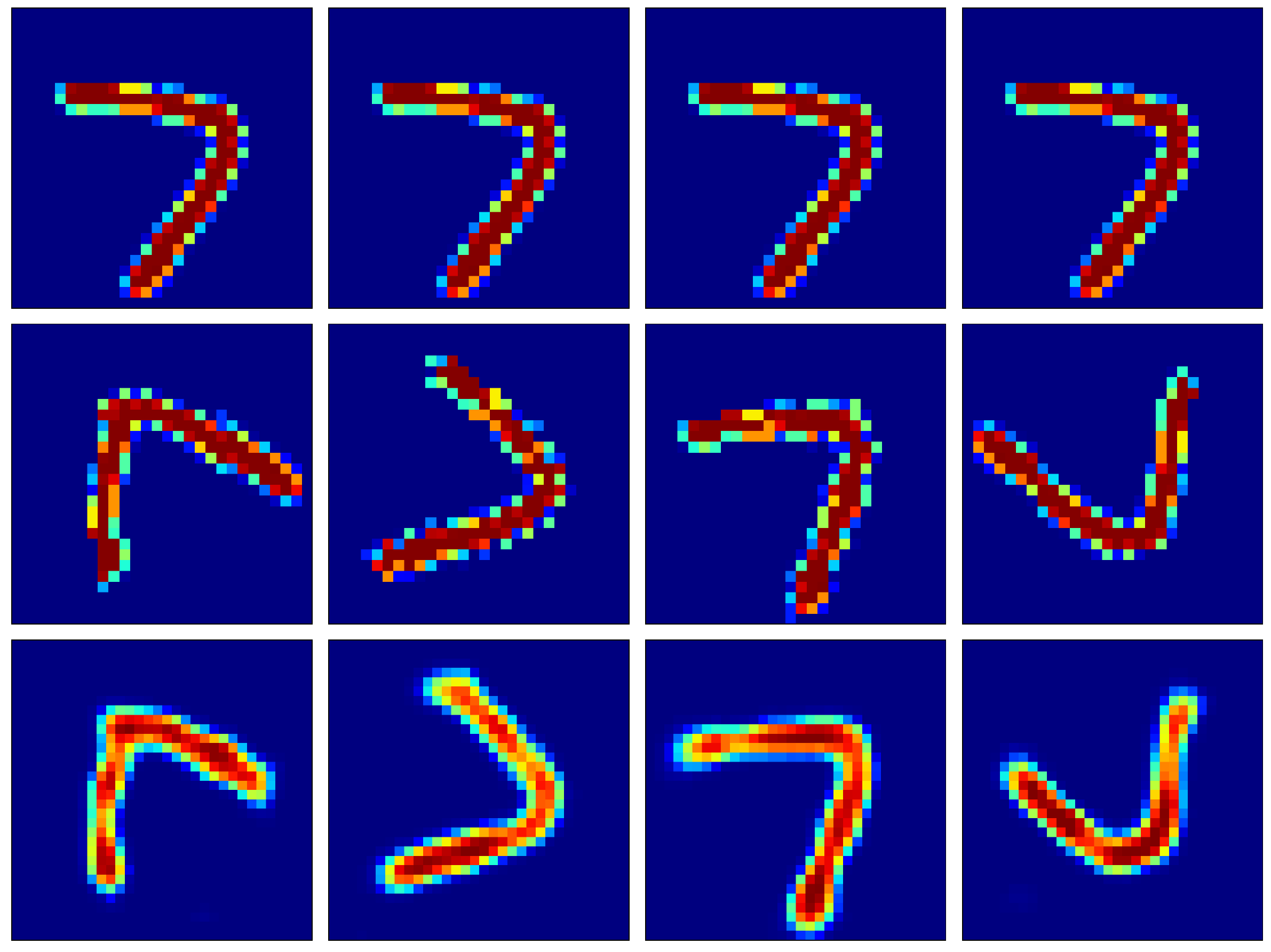}
		}
	\end{minipage}%
	\begin{minipage}{.33\columnwidth}
		\centering
		\subfloat[RA autoencoder.\label{fig:mnist-img2}]{
		    \includegraphics[width=.9\textwidth]{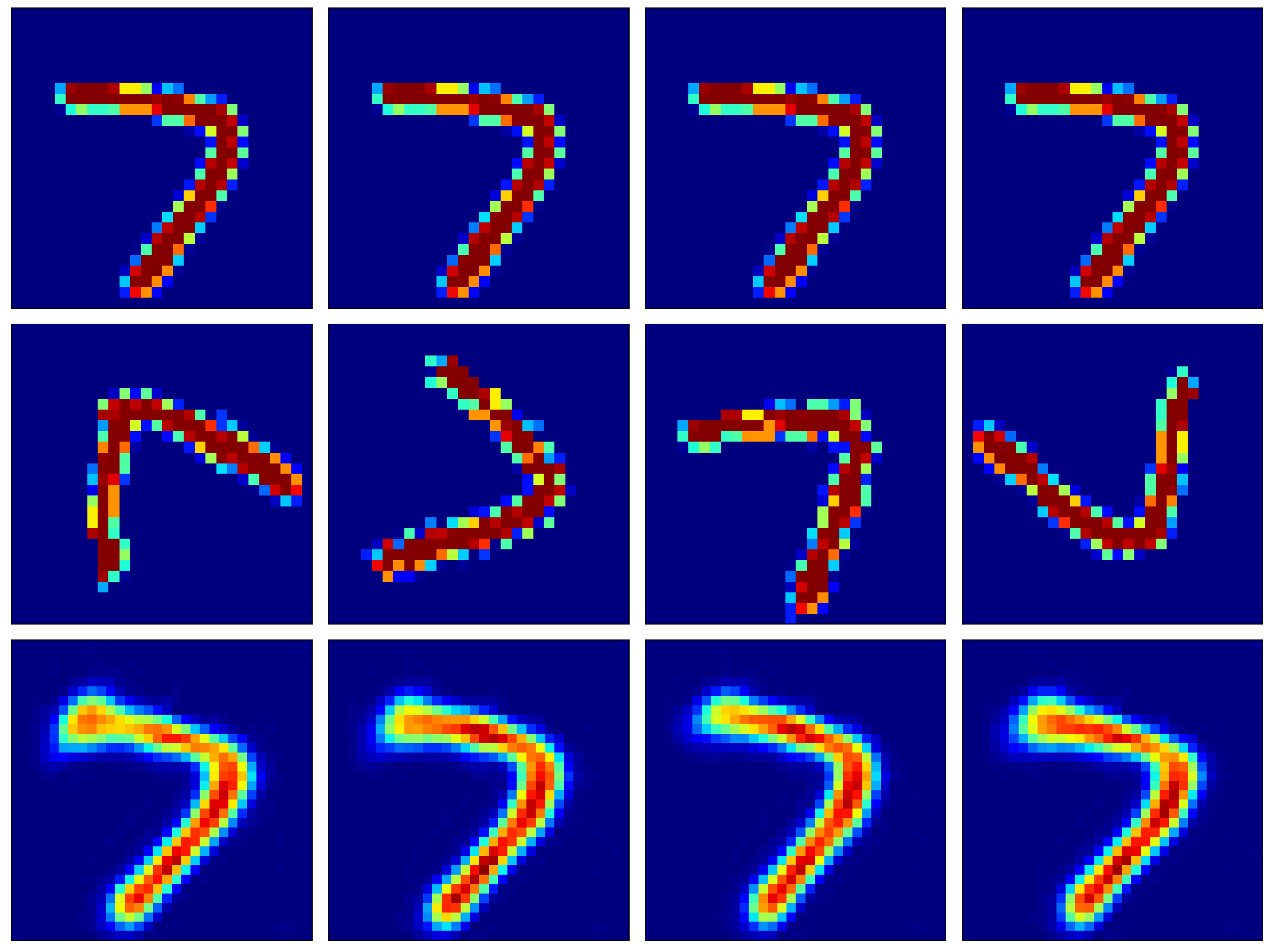}
		}
	\end{minipage}
	\begin{minipage}{.33\columnwidth}
		\centering
		\subfloat[RI autoencoder.\label{fig:mnist-img3}]{
		    \includegraphics[width=.9\textwidth]{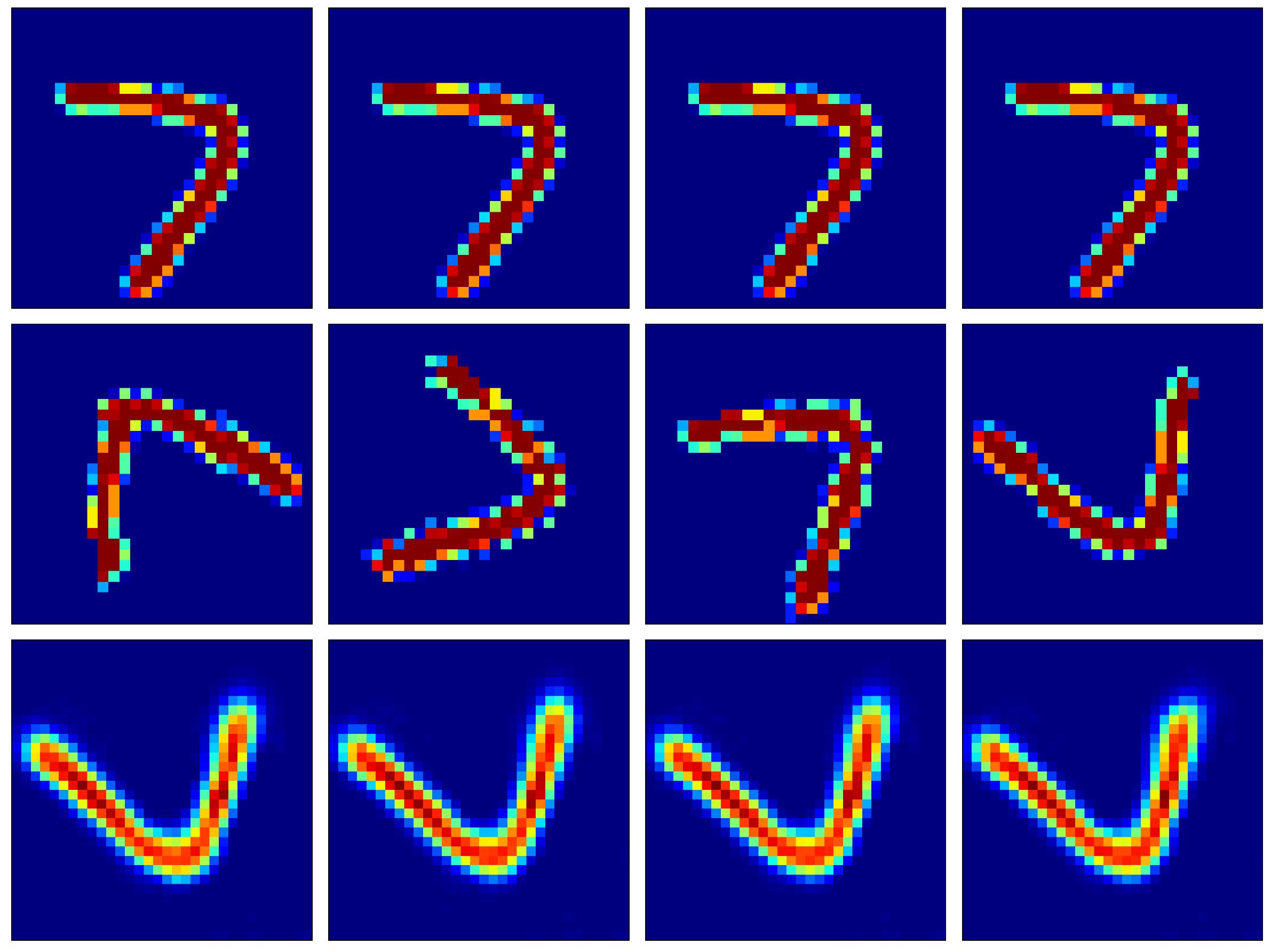}
		}
	\end{minipage}
	\caption{A single instance of the digit class `7' from MNIST (top row), four random rotations of that image (middle row), and restorations of those rotated images (bottom row). 
    (a) The NRI autoencoder yields images that display no canonical rotation.
    (b, c) The RA and RI autoencoders both produce restored images in a uniform canonical rotation.}\label{fig:mnist_images}
\end{figure*}
\begin{figure}[htbp]
    \centering
    \includegraphics[width=0.9\columnwidth,trim=1mm 1mm 1mm 1mm,clip]{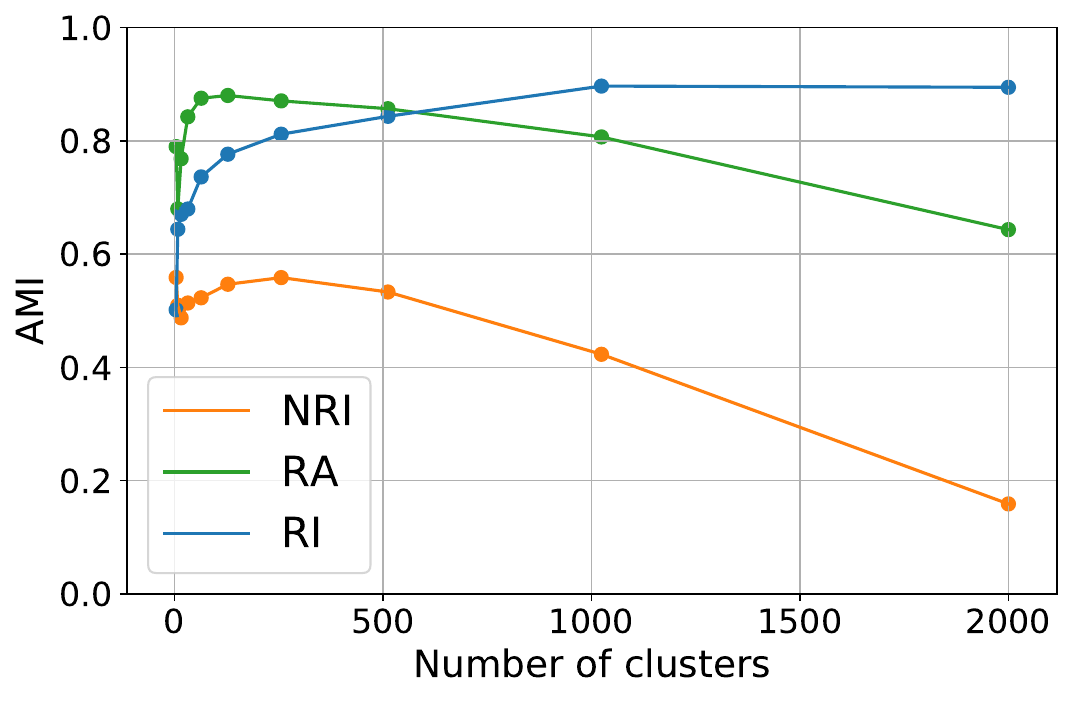}
    \vspace{-2ex}
    \caption{\emph{Test \ref{test42}: multi-cluster, applied to NRI, RA, and RI autoencoders}. Clustering agreement scores (AMI) 
    on the \texttt{Test} dataset, 
    for from 4 to 2000 clusters. The AMI curve for RICC converges at 0.9, meaning that RICC autoencoder produces rotation-invariant latent representation.  
    }
    \label{fig:ri_multi}
\end{figure}
\subsection{Results for Criterion 4: Rotation-Invariance}\label{sec:multicluster}

We apply the two tests described in Section~\ref{sec:test:rotate}.

\paragraph{Results for Test \ref{test41}: MNIST Rotation-Invariance}
Fig.~\ref{fig:mnist_images} shows restorations of a digit image `7' as produced by three different autoencoders: the NRI autoencoder of Equation~(\ref{eq:cnri}); the best RA autoencoder with $\lambda$ = 0.1 (Fig.~\ref{fig:mnist-img2}); and the best RI autoencoder with $\lambda_{\text{inv}}$ = 10, $\lambda_{\text{res}}$ = 10 (Fig.~\ref{fig:mnist-img3}).
We see that in this MNIST test, the RA and RI autoencoders both produce outputs with a single canonical rotation, as well as preserving input patterns with high fidelity.

\paragraph{Results for Test \ref{test42}\AddAdd{:} Multi-cluster}
This test is designed to verify
whether RICC groups similar clouds into the same cluster regardless of image orientation.
We use the 2000 \texttt{Test} patches (see Section~\ref{sec:testdata}) as holdout patches.
We make 11 copies of each patch in our holdout set, with and without rotations.
We rotate every 30$^{\circ}$; thus, the ideal result should return the same cluster label for both the original patches and the rotated copies.
We then implement HAC clustering for from \num{4} to \num{2000} clusters. 
We would expect the AMI score to be close to \num{1} for \num{2000} clusters because our \texttt{Test} dataset has \num{2000} patches, meaning that each patch and its replications can be placed in a unique cluster.
Fig.~\ref{fig:ri_multi} plots AMI scores as a function of the number of clusters. 
The AMI curve for the RICC (blue) converges to 0.9, indicating that the clustering result is agnostic to the orientation of clouds in the holdout set.
In contrast, the AMI curve for RACC (green) achieves 0.88 at the number of cluster 128 but decreases to 0.64, telling us that the latent representation insufficiently learns spatial features in inputs, rendering clustering assignments at random among the similar structure of cloud images.
We see that the agreement score for the NRICC decreases as the number of clusters increases, suggesting that the NRI autoencoder's latent representation is influenced by the rotation of images even if they are for the same types of clouds.
Therefore, we conclude that RICC is functional to process a real image dataset when input orientation does not matter for pattern recognition.

\section{Comparison to a Labeled Dataset}\label{sec:experiment}
\begin{figure}
    \centering
    \includegraphics[width=\linewidth]{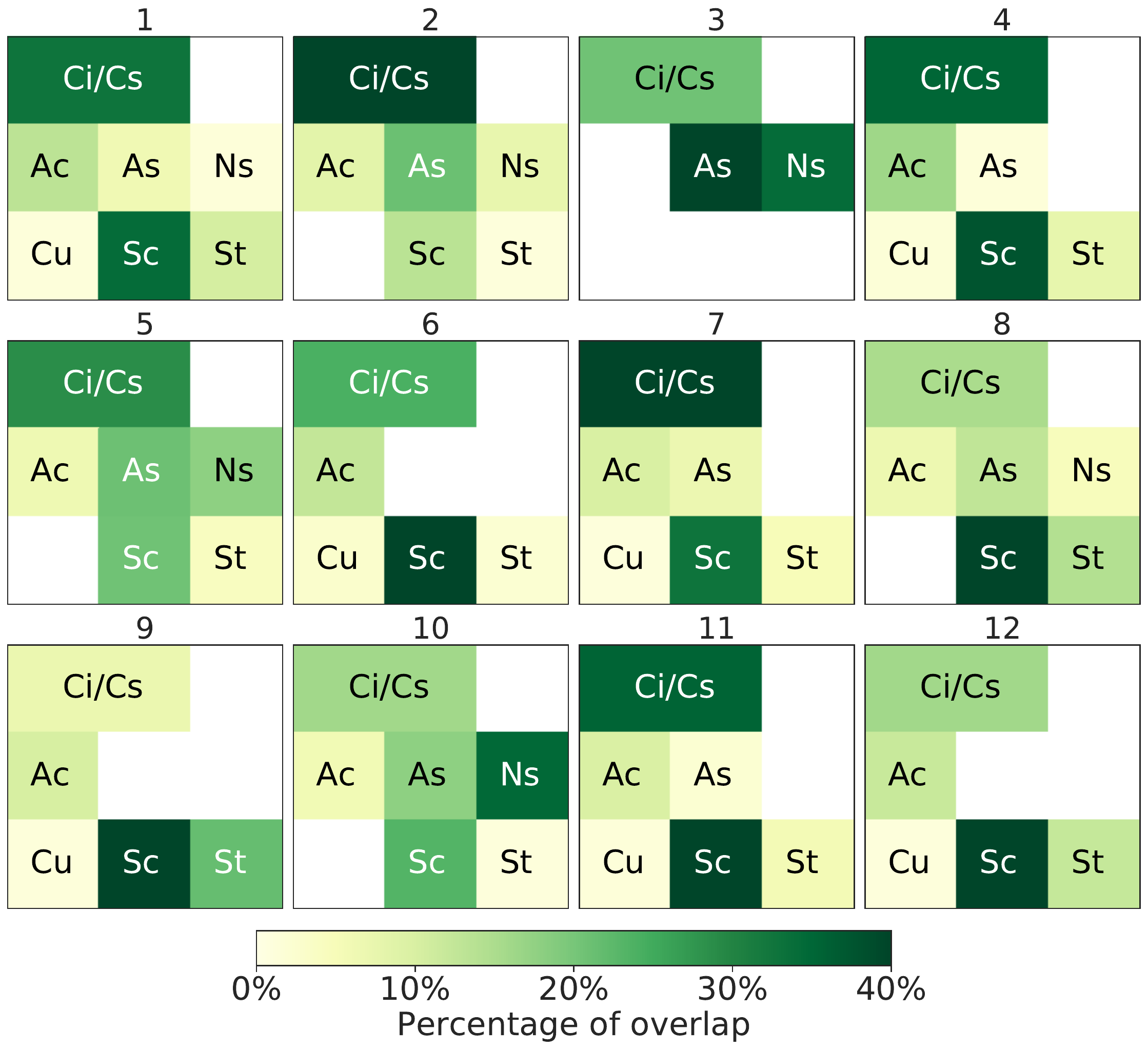}
    \vspace{-2ex}
    \caption{Heatmap histograms for eight of the nine ISCCP cloud categories~\cite{isccp1991} (see Fig.~\ref{fig:clouds}) when RICC is used to generate 12 clusters. 
    Note that Ci and Cs are combined and that deep convection is not included.
    A white panel represents a cluster with $<$1\% of pixels in that category.
    Most clusters contain one of primarily low- (Cu, Sc, St), middle- (Ac, As, Ns), or high-altitude (Ci-Cs, Dc) cloud categories.
    Clusters 3, 5, and 10 also contain thick clouds (Ns).
    The figure shows that cloud clusters produced by our unsupervised approach are compatible with labels produced via supervised approaches.
    It also suggests that more than 12 clusters are needed to achieve clearer separation of high and low clouds in clusters 1, 4, 7, and 11.
    }
    \label{fig:isccp12}
\end{figure}


Previous unsupervised learning studies have not investigated in any detail the association between clusters produced by unsupervised learning and assignments of established cloud categories generated by supervised learning methods.
In this final study we undertake such an investigation by
comparing RICC clusters 
with cloud categories from a supervised cloud classifier developed by Zantedeschi et al.~\cite{Zantedeschi2019CumuloAD}. 
\Add{Note that the purpose of this comparison analysis is to investigate the distribution of \AddAdd{the} overlap of our cloud clusters on labeled \AddAdd{known} cloud categories, rather than to measure the replication of their labels using the RICC framework via common metrics such as accuracy.}

Zantedeschi et al.\ build their supervised cloud classifier by using data from the 2B-CLDCLASS-lidar product~\cite{sassen2008global,cloudsat2008}, which combines CloudSat CPR and CALIPSO lidar measurements, as their baseline. 
To prepare a training set, they  
assign a cloud category label to each January 2008 MODIS MOD02 1~km resolution pixel from the Aqua instrument (essentially equivalent to the data considered in our work) that overlaps with 2B-CLDCLASS-lidar data for January 2008.
Specifically, they assign to each overlapped MOD02 pixel 
one of the nine ISCCP cloud categories, 
with cirrus and cirrostratus (Ci and Cs) combined for a total of eight unique classes,
based on the most frequent cloud category found in the 2B-CLDCLASS-lidar vertical layers at that pixel.
The resulting training set, which combines MODIS spectral radiances and cloud properties plus 2B-CLDCLASS-lidar-based labels, is relatively small,
as the 2B-CLDCLASS-lidar pixel-width `tracks' cover just 0.07\% of the MOD02 data. 
They then use these data
to train a hybrid invertible residual network~\cite{Nalisnick2019HybridMW} (IResNet) as a pixel-level cloud classifier.
Finally, they apply this classifier to predict labels for all pixels in the January 2008 MODIS MOD02 1~km resolution data.
It is this resulting dataset, which they name CUMULO, that we compare against RICC.

We first reproduce Zantedeschi et al.'s work by training their 
IResNet classifier on \num{2098755} 3 pixels $\times$ 3 pixels \textit{tiles}, labeled via 2B-CLDCLASS-lidar as just described, from the January 2--31, 2008 MODIS MOD02 1~km Aqua dataset.
We test the trained model on \num{47360560} equivalent tiles from January 1, 2008,
obtaining 89.0\% validation accuracy, similar to the 90.9\% reported by Zantedeschi et al.

We then apply RICC to the \num{58734} 128$\times$128 MOD02 patches for 
January 1, 2008, to obtain RICC clusters for the area that is covered by the \num{2098755}-tile IResNet test set.
We use a cluster count of 12 in the subsequent comparison analysis.
To permit comparison of IResNet pixel-level and RICC patch-level clustering, 
we map RICC clusters to the pixel level by 
identifying for each MOD02 pixel the patch(es) that contain it
(due to overlaps among patches, one pixel can be in up to four patches)
and for each such patch, the cluster in which it places the pixel,
and then assigning to the pixel either the cluster in which it is placed most frequently,
or if there is no single most frequent cluster, the first cluster encountered.
In this way, we obtain for each pixel contained on January 1, 2008, MOD02 dataset two classifications, an ISCCP class from IResNet and a cluster number from RICC. 

Then, for each of the 12 RICC-assigned clusters,
we examine the ISCCP labels assigned to pixels in that cluster,
with results shown in Fig.~\ref{fig:isccp12}.
We see that eight of the 12 clusters have a dominant ($>$40$\%$) CUMULO category.
Furthermore, a majority of the clusters contain primarily of either low- (Cu, Sc, St), middle- (Ac, As, Ns), or high-altitude (Ci-Cs, Dc) cloud categories.
\AddAdd{Clusters \#3, \#5, and \#10 consist of patches containing thick clouds (Ns).}
\Add{Thus, the 12 clusters produced by RICC are compatible with known cloud categories produced via supervised approaches.} 
Exceptions are clusters \#1, \#4, \#7, and \#11, which combine the high-altitude Ci-Cs and low-altitude Sc categories within the same cluster---a phenomenon
that we attribute to the high frequency with which the trained classifier assigns the Sc category.
\Add{These results \AddAdd{suggest} that more than 12 clusters are needed to achieve a clear separation of high and low clouds--a topic for future work.}

As an additional physical-based evaluation, we revisit 
Fig.~\ref{fig:physicalreasonable} histograms of the COT, CPH, CTP, and CER variables for the 12 clusters.
CTP and CER show distinct peaks in different clusters;
COT shows more compact distributions, with either shorter or longer tails; and
CPH shows different proportions of pixel-based cloud phases.
The CTP distributions suggest that our novel cloud clusters are able to distinguish low, middle, and high altitude clouds, as different clusters have clear peaks at around 950, 500, and 200 hPa, respectively.
The CER distributions for clusters \#2, \#3, and \#10 present distinct higher peaks (more than 30$\mu$m), suggesting that our clusters further distinguish higher and middle altitude clouds according to whether they consist predominantly of larger cloud droplets.
We see similar trends between clusters \#2 and \#7 in Fig.~\ref{fig:isccp12} (both select high clouds Ci/Cs ), whereas CPH and CTP histograms show clear differences in their distributions (cluster \#2 has a dominant ice phase in the CPH's distribution, and distinct middle/high ($<$680 hPa) patch-based mean values), 
demonstrating that RICC learns the additional physical parameters required for detailed clustering.

The COT distributions have less clearly distinguished peaks than CTP and CER: most clusters have a peak at around 5--10. 
Nevertheless, the COT distributions are well associated with the separability of clouds as a function of thickness.
For example, cluster \#3 is composed of about 44$\%$ of As cloud pixels and 35$\%$ of Ns cloud pixels, and their COT distributions in COT histogram are long tail, meaning that our novel clusters differ in their optical thickness.
Note that the overall mean value for COT in~\texttt{TestCUMULO} is \num{11.86} over the parameter range [0,100], suggesting that subtle differences among the distributions can explain the separation between the thick (Ns, St) and thin (Ci, Ac, Cu) categories.

Cluster \#7 is distinguished by its lack of a single dominant ISCCP category:
it includes significant amounts of both Ci and Sc.
To investigate, we compute the patch-based mean and standard deviations of CTP 
and create histograms of each (not shown in this paper) to study their distributions. 
The histogram of standard deviations shows a bimodal distribution, 
with one peak at 100 hPa and another at 270 hPa;
the histogram of means shows a long tail distribution with a peak at 800 hPa.
This result suggests that cluster~\#7 combines both multi-class patches
(i.e., ones containing both high and low clouds) and patches
that are dominant in low clouds.


\section{Conclusions}\label{sec:conclusion}

We have presented an unsupervised data-driven cloud clustering framework that addresses the previously unsolved rotation dependence problem in cloud image classification 
by combining a convolutional autoencoder with a rotation-invariant loss function that decouples the restoration and transform variance terms to learn both spatial and rotation features simultaneously.
We apply seven test protocols to verify that this rotation-invariant cloud clustering (RICC) system yields physically reasonable and spatially coherent clusters, and learns spatial and rotation-invariant features.
We also demonstrate that it yields cloud clusters that match reasonably well with established cloud categories as estimated via a supervised approach.
These results support the possibility of using an unsupervised data-driven approach for the automation of cloud clustering and pattern discovery, without the prior hypothesis of ground-truth labels.

RICC's innovative combination of data selection, deep convolutional neural network architecture, loss function, autoencoder training protocol, and clustering method allow it to realize practicable unsupervised cloud clustering, a previously unmet goal.
All five features just listed appear to be essential to success.
For instance, an autoencoder of an identical architecture trained only on the MODIS surface reflectance product (i.e., without radiance data) clusters clouds by
albedo and cloud texture rather than cloud properties; in contrast,
RICC's use of visible and thermal bands relative to cloud optical and top properties allows it to both capture different cloud texture and emulate distinct physical feature, 
and thus to achieve better performance than Denby~\cite{denby2019unsuper}.
The results indicate that neural networks reflect nonlinear interrelations among input data, suggested by previous studies with application of neural networks in satellite remote sensing~\cite{krasnopolsky2009neural}.
We also demonstrated that deep convolutional networks are required for our loss function to achieve rotation-invariance: 
both shallow and fully connected networks were insufficient, 
and too compact a representation fails to reconstruct the original cloud structure with sufficient fidelity.
Our proposed evaluation metrics also represent a contribution.
Common evaluation metrics such as L2 loss proved to be insufficient for distinguishing among different autoencoders, giving similar results for all three autoencoders evaluated in this paper. 
In contrast, our quantitative and qualitative evaluation protocols can distinguish useful from non-useful autoencoder approaches, and should prove useful for studying applications of unsupervised learning in cloud classification and more generally in the Earth sciences. 

Overall, our RICC system can be used both to produce clusters matched to a standard cloud classification and to obtain more sophisticated insights into the robustness of those classifications.
Now that we have validated that RICC satisfies our first four criteria, the fifth, \emph{stable} clusters~\cite{von2010clustering} could be a subject of future work. 
\Add{This fifth criterion is expected to support the determination of the optimal number of clusters in our framework.}
We also plan to apply RICC to large quantities of MODIS imagery to investigate 20-year global trends in the distribution of cloud patterns, with the goal of achieving a data-driven diagnosis of cloud organization behaviours.
We also envision re-visiting regional changes in spatial cloud textures examined in previous large-scale studies~\cite{7451236} that demonstrated changes in the radiance and spatial texture changes of the Earth's atmosphere due to natural and anthropogenic variability in the climate system.

\section*{Acknowledgments}

The authors thank the University of Chicago's Research Computing Center for access to computing resources, and Valentina Zantedeschi and co-authors for access to CUMULO labeled dataset and scripts.

\bibliographystyle{IEEEtran}
\bibliography{references}

\begin{appendices}
\counterwithin{figure}{section}
\counterwithin{equation}{section}

\section{RI Autoencoder Hyperparameter Search}\label{sec:valid_cloud}
\begin{figure}[htb!]
    \begin{center}
        \includegraphics[width=\textwidth,trim=1mm 1mm 1mm 1mm,clip]{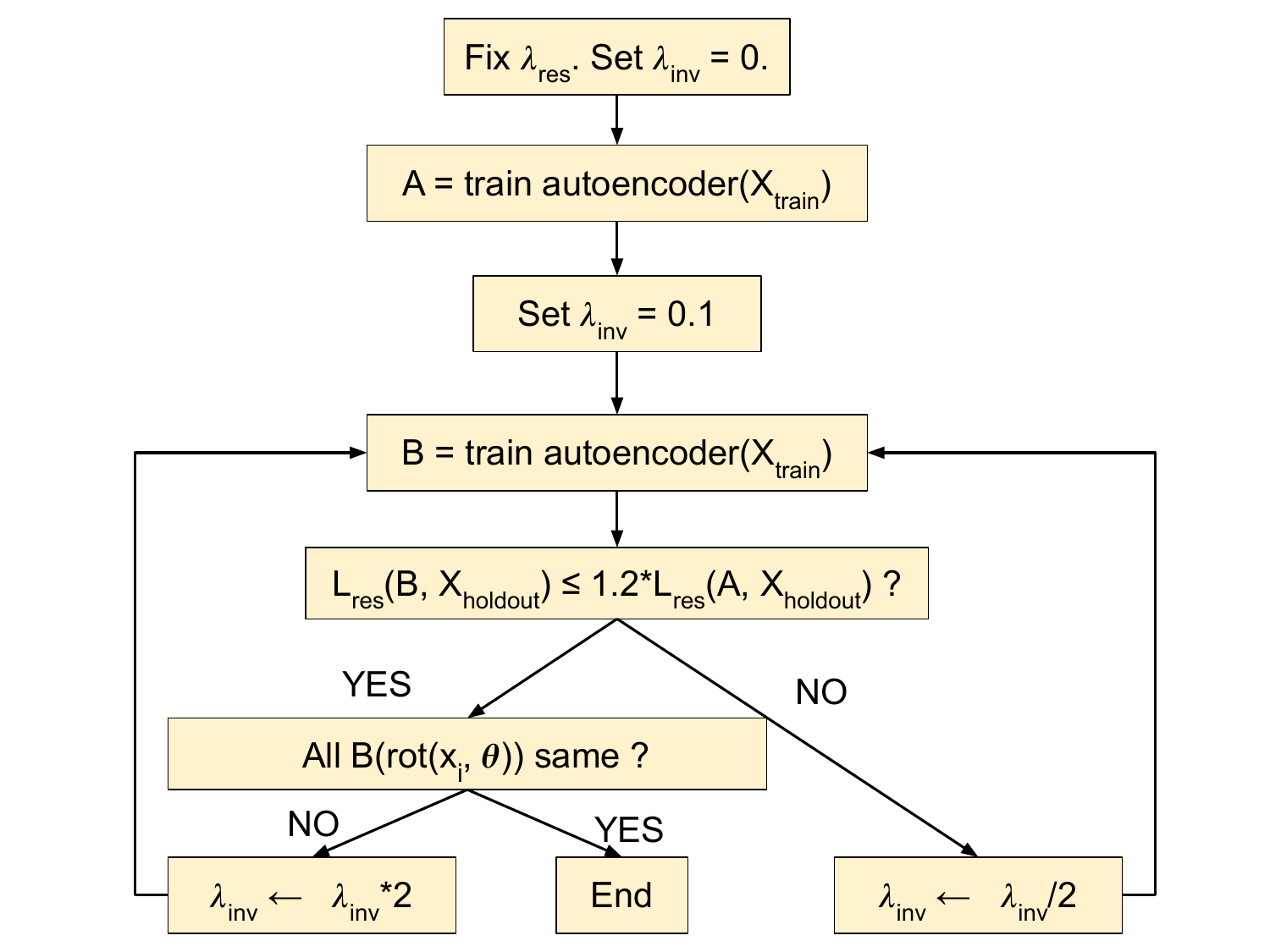}
    \end{center}
    \vspace{-2ex}
    \caption{Flowchart of the parameter search method used to determine values for $\lambda_{ \text{inv}}$ and $\lambda_{ \text{res}}$ for cloud images.}
    \label{gridsearch_protocol}
\end{figure}

The performance of the RI autoencoder on a particular training set is sensitive to the values assigned to the $\lambda$ parameters.
We formulate in Fig.~\ref{gridsearch_protocol} a grid search process for finding an optimal combination of $\lambda_{ \text{inv}}$ and $\lambda_{ \text{res}}$.
(We also test a number of learning rates, for 10$^{-4}$ to 10$^{-2}$, for stochastic gradient descent.)
This process proceeds as follows:

\begin{enumerate}
    \item Fix $\lambda_{\text{res}}$ and learning rate \emph{lr}. Set $\lambda_{ \text{inv}}$ = 0. 
    \item Train the autoencoder to obtain a baseline trained model \textbf{A}; measure its restoration loss $L_{\text{res}}$, on a holdout set $X_{\text{holdout}}$: 
    $L_{\text{res}}(\textbf{A}, X_{\text{holdout}})$.
    \item Set $\lambda_{\text{inv}}$ = 0.1.
    \item Train the autoencoder with the new $\lambda_{\text{inv}}$ to obtain a new trained model \textbf{B}; measure its restoration loss on the same holdout set, giving $L_{\text{res}}(\textbf{B}, X_{\text{holdout}})$.
    \item If the restoration loss of the newly trained model \textbf{B} is less than 20$\%$ larger than that of the baseline \textbf{A}, i.e., if $L_{\text{res}}(\textbf{B}, X_{\text{holdout}}) \leq 1.2 L_{\text{res}}(\textbf{A}, X_{\text{holdout}})$, then double $\lambda_{\text{inv}}$. 
    \item Compare the results produced by \textbf{B} for images $x_i\in X_{\text{holdout}}$, each replicated $N$ times, and with each replicate rotated by a different $\theta$. Observe whether, for each image, all $\textbf{B}(\texttt{rot}(x_i, \theta))$ have the same canonical rotation. 
    (In our implementation, we assume that \emph{N} = 12 and $\theta\in \{0, 30, \cdots, 330\}$.) We verified $\textbf{B}(\texttt{rot}(x_i, \theta))$ based on the cosine similarity and eye-ball observation.
    \item If for all $x_i \in X_{\text{holdout}}$, all $\textbf{B}(\texttt{rot}(x_i, \theta))$ are the same, terminate the search.
    \item Otherwise halve $\lambda_{\text{inv}}$.
\end{enumerate}

\begin{figure*}
	\centering
	\begin{minipage}{.225\columnwidth}
		\centering
		\subfloat[Ratio of restoration losses\label{fig:gridsearch_lossratio}]{
		\includegraphics[width=\textwidth]{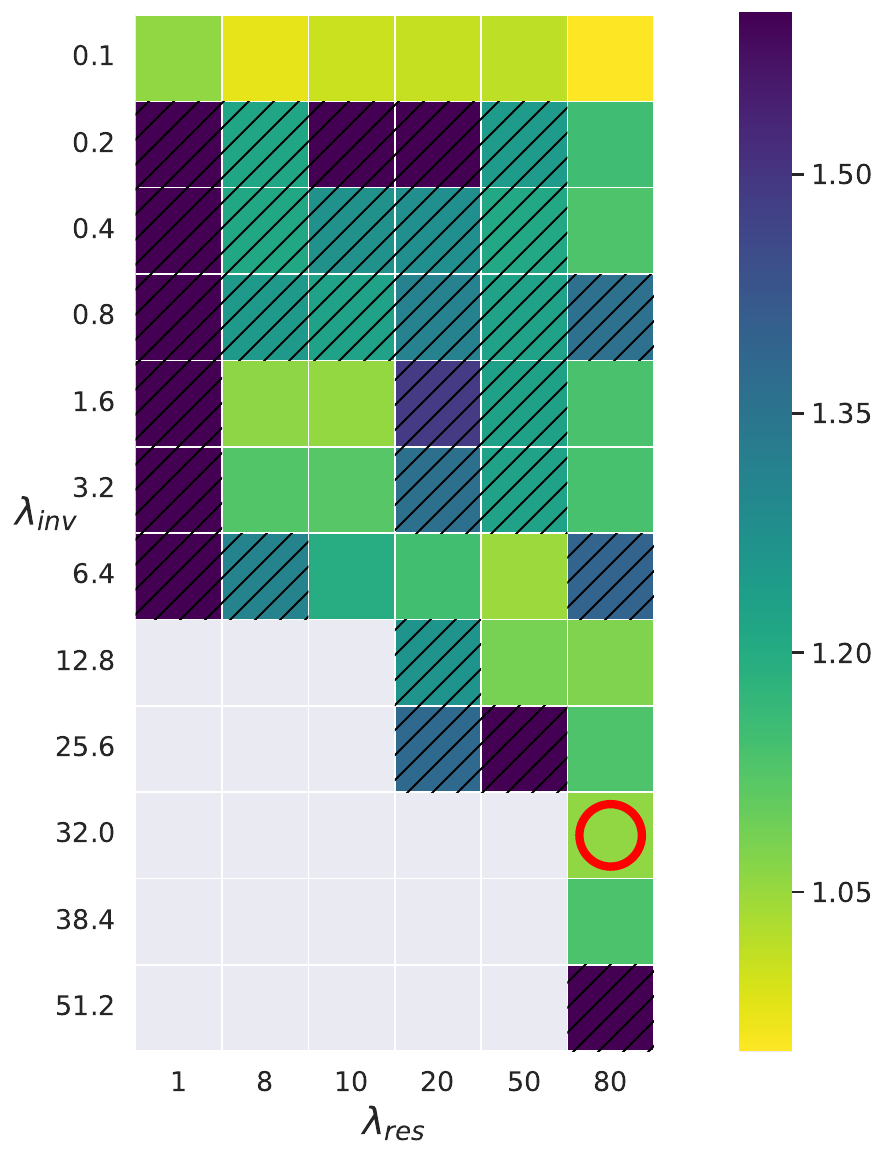}}
	\end{minipage}%
	\hspace{5mm}
	\begin{minipage}{.225\columnwidth}
		\centering
		\subfloat[Similarity of restored images. \label{fig:gridsearch_cossim}]{\includegraphics[width=\textwidth]{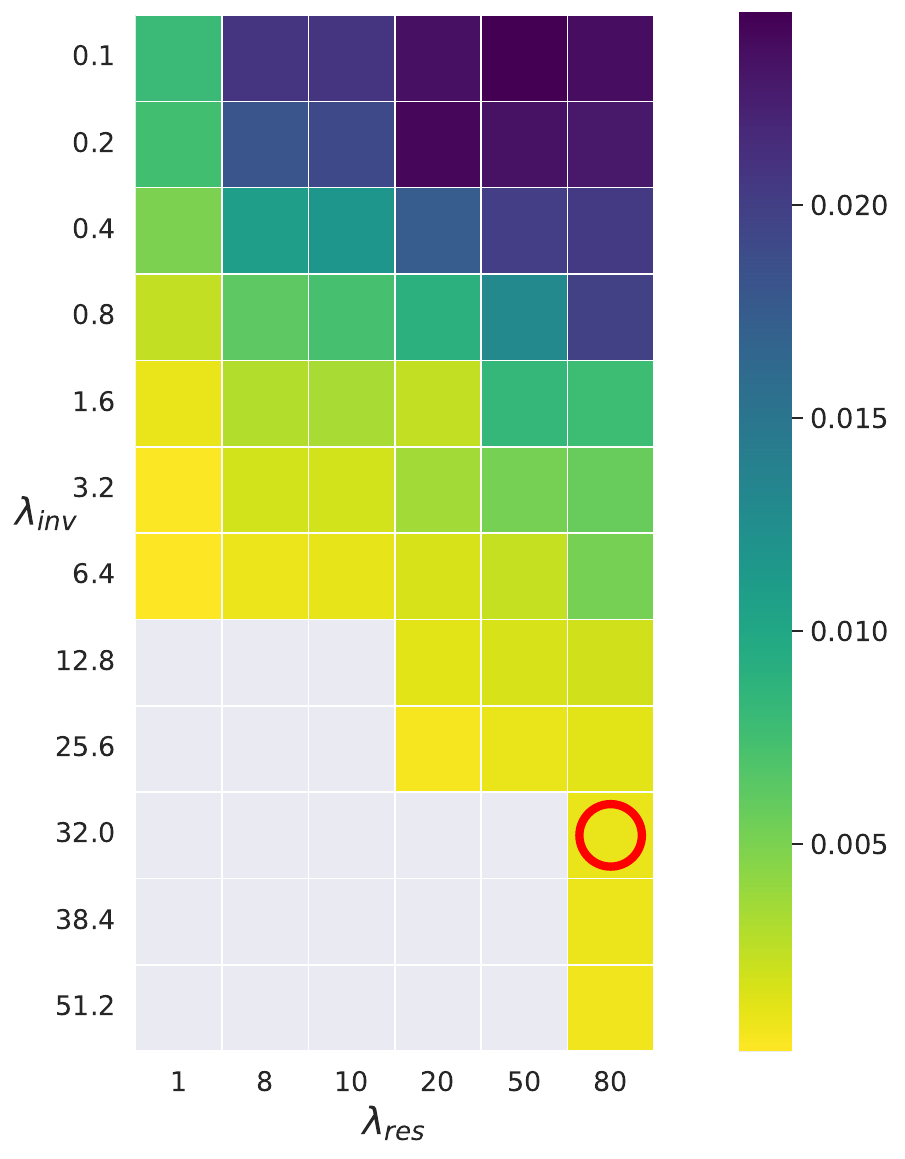}}
	\end{minipage}%
	\hspace{5mm}
	\begin{minipage}{.45\columnwidth}
		\centering
		\subfloat[Training losses for different $\lambda$ combinations.\label{fig:gridsearch_loss_cloud}]{\includegraphics[width=\textwidth]{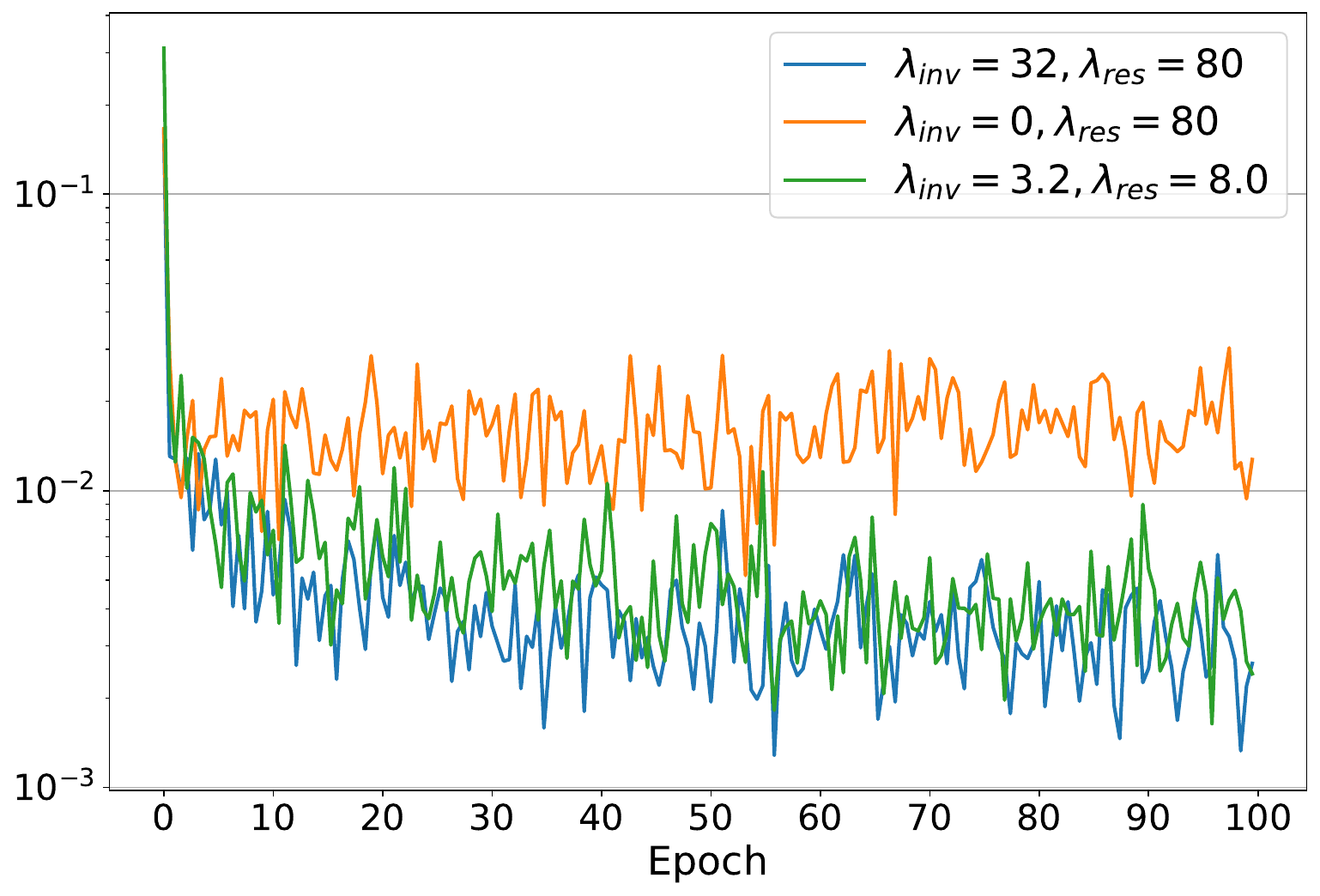}}
	\end{minipage}
	\caption{Results of grid search for RI autoencoder. (a) Ratio of the two restoration losses $L_{\text{res}}(\textbf{B}, X_{\text{holdout}})/ L_{\text{res}}(\textbf{A}, X_{\text{holdout}})$ for different $\lambda$ combinations. (b) Standard deviation of cosine similarity computed on the restoration images from RI autoencoder, when feeding various transformations rotated by $\{0^{\circ}, 30^{\circ},\cdots, 330^{\circ}\}$. We highlighted the optimal combination by a red circle. (c) Training losses for the optimal $\lambda$ combination (blue) and two suboptimal combinations (orange and green) examined in the grid search.  }\label{fig:gridsearch_result_all}
\end{figure*}

We then repeat this procedure for a new $\lambda_{\text{inv}}$ value.

As shown in Fig.~\ref{fig:gridsearch_result_all},
this grid search reveals that the optimal parameter combination is $\lambda_{\text{inv}}$ = 32 and $\lambda_{\text{res}}$ = 80 for \emph{lr} = 10$^{-2}$. 
Other parameter combinations fail to satisfy the restoration error ratio criterion [$L_{\text{res}}(\textbf{B}, X_{\text{holdout}}) > 1.2 L_{\text{res}}(\textbf{A}, X_{\text{holdout}})$] (Fig.~\ref{fig:gridsearch_lossratio}): i.e., the restoration loss in the newly trained model B produces restoration errors more than 20$\%$ larger than those of our baseline, and/or the rotation-invariant criterion: i.e., $\textbf{B}(\texttt{rot}(x_i, \theta))$ is not identical for all $\theta\in\{0, 30, \cdots, 330\}$ (Fig.~\ref{fig:gridsearch_cossim}).
Fig.~\ref{fig:gridsearch_lossratio} shows the ratio of restoration loss  $L_{\text{res}}(\textbf{B}, X_{\text{holdout}}) / L_{\text{res}}(\textbf{A}, X_{\text{holdout}})$ where hatched elements indicate $L_{\text{res}}(\textbf{B}, X_{\text{holdout}}) > 1.2 L_{\text{res}}(\textbf{A}, X_{\text{holdout}})$.
In Fig.~\ref{fig:gridsearch_lossratio}, navy elements are more likely to show at lower $\lambda_{\text{res}}$ ($\lambda_{\text{res}}=1, 10, 20$) and larger $\lambda_{\text{inv}}$ ($(\lambda_{\text{res}}, \lambda_{\text{inv}})  = (50,25.6), (80,51.2)$), suggesting that a larger $\lambda_{\text{res}}$ decreases the restoration error ratio, allowing for larger $\lambda_{\text{inv}}$ that satisfy the criterion.

To enable quantitative investigation of rotation-invariance, 
we show in Fig.~\ref{fig:gridsearch_cossim} the standard deviation of the cosine similarities for RI autoencoder outputs as a function of $\lambda$ values.
This measure of the similarity of two vectors $X$ and $X^{\prime}$ is defined in terms of
the cosine of the angle between them:
\begin{equation}\label{eq:cossim}
    \text{Cosine similarity} = \frac{<X, X^{\prime}>}{\|X\| \cdot \| X^{\prime}\|} .
\end{equation}

Equation~(\ref{eq:cossim}) gives 1 when elements of two vectors are exactly matched. 
As we assume $\texttt{rot}(x_i, \theta)$ for $\theta\in$ \{0, 30, $\cdots$, 330\}.), we computed the similarity metric among the output for $\theta = 0$ and the other angles, and then evaluated the standard deviation.
A standard deviation closer to 0 in Fig.~\ref{fig:gridsearch_cossim} means that cosine similarities in restored images with and without rotation obtain an identical representation, which verifies the autoencoder maps images with different transformations into a single canonical orientation.
Studying the result of standard deviations, we see that increasing $\lambda_{\text{res}}$ needs increasing $\lambda_{\text{inv}}$ to achieve the rotation-invariant criterion, while increasing $\lambda_{\text{res}}$ leads to better-quality restorations.
Fig.~\ref{fig:gridsearch_loss_cloud} shows the evolution of training loss over iterations with different parameter values.
The lowest convergence of the blue line in Fig.~\ref{fig:gridsearch_loss_cloud} proves that the optimal combination ($\lambda_{\text{inv}}$ = 32, $\lambda_{\text{res}}$ = 80) achieves the lowest rotation-invariant loss of the combinations shown.
(We do not consider $\lambda$ parameter values higher than those shown,
as that causes the minimization to be unstable.)

We note that the selection of learning rate is also key to discovering the optimal configuration. 
Learning rates lower than the 10$^{-2}$ considered here allow the network to take larger $\lambda$ values, which results in the latent representation not being able to achieve rotation-invariance.

\section{Dimensionality reduction technique}\label{sec:tSNE}
t-SNE is a probabilistic nonlinear dimensionality reduction technique that maps each data point in a high-dimensional space to a lower-dimensional point in such a way that, with high probability, similar data are placed near to each other and dissimilar data far apart. 
Suppose that we have $N$ latent representations $Z = \{z_1, \cdots, z_i, z_j, \cdots, z_N\}$ produced by an autoencoder.
Let $P$ be a joint probability distribution in the high-dimensional input space, 
and $Q$ a joint probability distribution in the low-dimensional projection space.
The t-SNE optimization minimizes the Kullback-Leibler (KL) divergence between $p_{j|i}$ and $q_{j|i}$. 
The conditional probability $p_{j|i}$ for the M-dimensional latent representation produced by our autoencoders is:
\begin{equation}
    p_{j|i}  = \frac{\exp(-\| z_i - z_j\|^2/2\sigma_i^2)}{\displaystyle \sum_{k\neq i}\exp(-\| z_i - z_k\|^2/2\sigma_i^2) },
\end{equation}
where $\sigma_i$ denotes the variance of a Gaussian distribution for data point $z_i$,
while for a two-dimensional map:
\begin{equation}
    q_{j|i}  = \frac{ \left(1 +\| z^{\prime}_i - z^{\prime}_j\|^2 \right)^{-1}}{\displaystyle \sum_{k\neq l} \left(1 +\| z^{\prime}_l - z^{\prime}_k\|^2 \right)^{-1} }.
\end{equation}
The cost function to be minimized by gradient descent is then
\begin{equation}\label{tsne-formula}
    \text{KL}(P||Q) = \sum_i \sum_j p_{j|i} \log{}\frac{p_{j|i}}{q_{j|i}} .
\end{equation}
Note that this joint distribution assumes a Student t-distribution with one degree of freedom in the low-dimensional map.


\section{The NRI Autoencoder}\label{sec:CNRI}
The 
\textit{non-rotation-invariant} (NRI) autoencoder~\cite{kurihana2019cloud}
introduces a more sophisticated loss function to address the reconstruction image issue shown in Fig.~\ref{fig:collapse}.
This loss function combines four metrics:
\begin{equation}\label{eq:cnri}
    L = L_1 + L_2 +  L_{\text{high-pass}} + L_\text{MS-SSIM}.
\end{equation}

The first two terms are the L1 and L2 loss, respectively, corresponding to $p = 1$ and $p = 2$ in Equation~(\ref{standardloss}).

The third term, a high pass filter to reduce noise in the reconstructed image,
is the high frequency error norm after passing data through a Sobel filter~\cite{sobel19683x3} to detect cloud edges in input images:
\begin{equation}
    \begin{split}
    L_{\text{high-pass}}(\theta) 
    &=  \sum_{x \in S} | g_X(x) - g_X(D_{\theta}(E_{\theta}(x)))| \\
    & \quad + | g_Y(x) - g_Y( D_{\theta}(E_{\theta}(x)) ) |,
    \end{split}
\end{equation}
where $g_X, g_Y$ denotes the gradients on the X and Y axes of the input patch, respectively. 

The fourth term is the multi-scale structure similarity index (MS-SSIM)~\cite{wang2003multiscale}, a multi-band version of the SSIM~\cite{wang2004image} index often used in computer vision to assess image similarity. 
This constraint rewards the production of similar latent representations for similar textures in cloud images:
\begin{equation}
    L_\text{MS-SSIM}(\theta) = \sum_{x \in S} \frac{(2\mu_{x} \mu_{\hat{x}_{\theta}+C_1)} (2\sigma_{x \hat{x}_{\theta}}+C_2)}{(\mu^2_{x} + \mu^2_{\hat{x}_{\theta}} + C_1)(\sigma^2_{x} + \sigma^2_{\hat{x}_{\theta}}+C_2)  },
\end{equation}
where $\mu$ is the mean of the patch, $\sigma$ is its variance; $C_1$, $C_2$ are variables to stabilize the division; and
$\hat{x}_\theta$ represents $D_{\theta}(E_{\theta}(x))$.
We use $C_1 = (0.01\cdot L)^2$ and $C_1 = (0.03\cdot L)^2$, where $L$ denotes the dynamic range for the sample data, i.e., the difference between the maximum and minimum pixel values.
\begin{figure}
	\centering
	\includegraphics[width=\columnwidth,trim=4mm 3mm 3mm 3mm,clip]{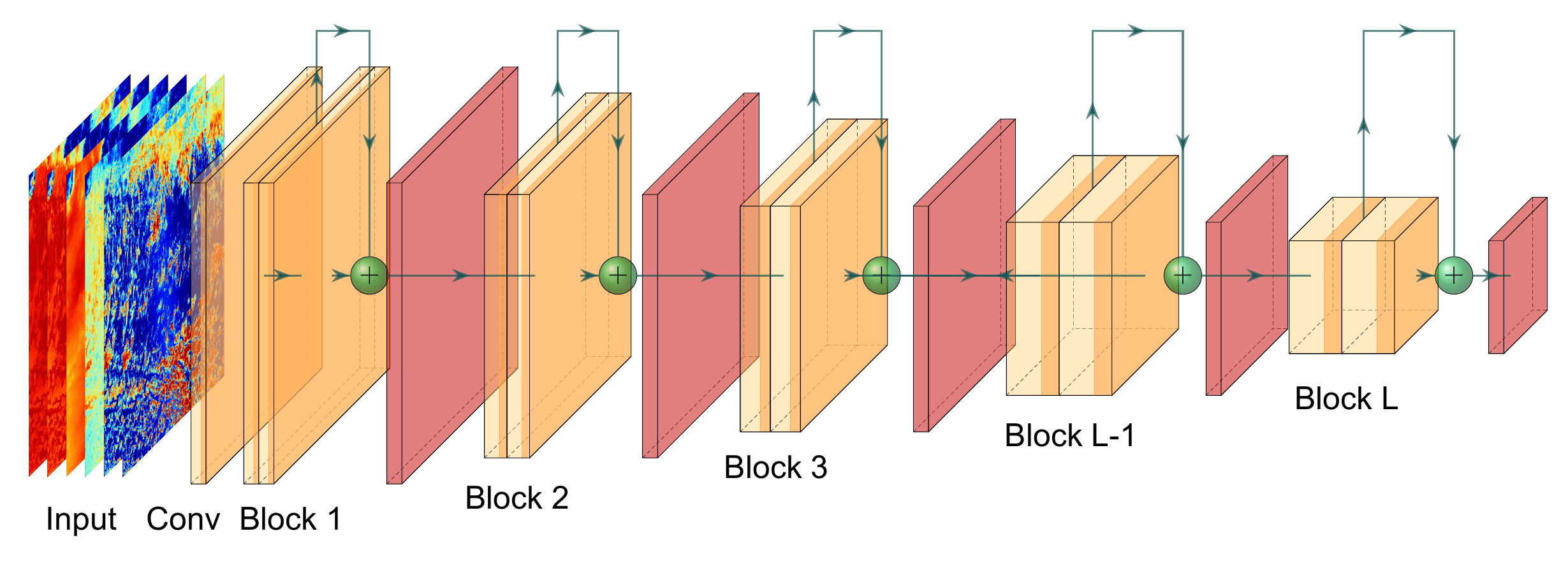}
	\caption{The NRI encoder architecture is identical to that of Fig.~\ref{fig:network:ricc},
	except that it has two rather than three convolutional layers per block, and skip connections (the arrows that skip over the convolutional layers).}\label{fig:cnri}
\end{figure}
The NRI autoencoder also has a slightly different network architecture: see Fig.~\ref{fig:cnri}.


\section{The Rotation-Aware Autoencoder}\label{sec:RA}
The NRI autoencoder 
produces different representations for images that differ only in their orientation.
This behavior is problematic for cloud clustering, because 
cloud formation is driven not by wind direction but by mechanisms such as adiabatic or non-adiabatic cooling, convection, advection, and terrestrial effects.
Furthermore, satellite image clouds from different angles.
Thus, any particular type of cloud can occur in different orientations in images.

To address this problem, the \textit{rotation aware} (RA) autoencoder 
optimizes simultaneously for both
an image reconstruction task (to learn spatial patterns) and 
a bottleneck representation task 
(to produce identical latent representations for an image and various rotations of that image).
Its loss function thus combines two terms:

\begin{eqnarray}\label{eq:RA}
    L_\text{RA}(\lambda) &=& L_{\textrm{\textrm{agn}}} + \lambda \cdot  L_{{\textrm{inv}^{\prime}}},
\end{eqnarray}
where $\lambda$ is a parameter used to balance the two terms. 

\textit{Rotation-Agnostic Loss:} 
The first loss term works to ensure that there
is at least one rotation angle for which the input image and its rotated
reconstruction are similar. 
To this end, it minimizes the difference between an input image, $x$,
and the restored image $D_{\theta}(E_{\theta}(x))$ when rotated by $R \in {\cal R}$.
It is defined as:
\begin{eqnarray}\label{eq:agnostic}
    L_{\textrm{\textrm{agn}}}(\theta) & = &
    \sum_{x \in S}{\min_{R \in {\cal R}} || x - R(D_{\theta}(E_{\theta}(x))) ||_{2}^{2}} .
\end{eqnarray}

\textit{Bottleneck Loss:} 
The second loss term works to ensure that any rotated version of the image input maps to a similar latent representation at the bottleneck layer.
It is defined as: 
\begin{eqnarray}\label{eq:RA:bottleneck}
   L_{\textrm{inv}^{\prime}}(\theta) &=&  \sum_{x \in S}{\max_{R \in {\cal{R}}}|| z_{\theta}(x) - z_{\theta}(R(x))||_{2}^{2}} ,
\end{eqnarray}
where $x$ and $R(x)$ are as for the rotation-agnostic loss, and
$z_{\theta}(\cdot)$ is an intermediate dimensionally reduced representation, meaning $z_{\theta}(x)$ is a bottleneck representation of the image $x$ and
$z_{\theta}(R(x))$ is that of the image rotated by $R \in {\cal R}$.

The key differences between $L_{\textrm{inv}}$ in Equation~(\ref{tinv}) and $L_{\textrm{inv}^{\prime}}$ in Equation~(\ref{eq:RA:bottleneck}) are that the latter computes a sum, rather than a max, over rotations ${\cal{R}}$, and computes distance in the original image space rather than bottleneck space.
Suppose that $z \approx E(D(z))$ for any latent representation $z$, 
and the autoencoder reconstructs high-fidelity images such that $D_{\theta}(E_{\theta}(x)) \approx x$.
Thus, we can rewrite the application of $E$ inside the norm in Equation~(\ref{tinv}) as follows, 
showing the similarity between $L_{\textrm{inv}}$ and $L_{\textrm{inv}^{\prime}}$.
\begin{equation}\label{eq:isometry}
\begin{split}
    x - D(E(R(x)) &\approx  E(x) - E(D(E(R(x)))) \\
    &\approx E(x) - E(R(x)) \\
    &= z(x) - z(R(x)) .
\end{split}
\end{equation}

In implementing the bottleneck loss, $L_{\textrm{inv}^{\prime}}$,
we need not apply the maximum operator 
to all variations of $R$ in [$0^{\circ}$, 360$^{\circ}$).
Instead, as we describe below,
we implement a data pipeline that for each image in a training set
first creates a fixed number of copies and then applies a random rotation to each copy,
to create a set $I$ of rotated versions.
Then, for each such set $I$, we
compute the sum of squares difference between the latent representation for each image $i$ in $I$ and those for all other images in $I$:
\begin{equation}\label{bottleneck_imp}
    L_{{\textrm{inv}}^{\prime}}^{i}(I) = \sum_{j \neq i} ||z_i - z_j ||^2_{2}.
\end{equation}

The bottleneck loss for each such set of rotated images in a minibatch is then:
\begin{equation}
    L_{\textrm{inv}^{\prime}}(I) = \sum_{i\in I} L^{i}_{\textrm{inv}^{\prime}}(I) \hspace{2mm}.
\end{equation}

Recall that the RA loss function (Equation (\ref{eq:RA})) contains a bottleneck loss term that is simplified by Equation~\ref{bottleneck_imp}.
To learn a similar representation of an image across the different rotations in similar types of images, 
we need to introduce a data augumentation scheme, allowing images in minibatch to consist of several transformations.

\begin{figure}[htbp!]
    \begin{center}
        \includegraphics[width=\textwidth,trim=20mm 22mm 23mm 48mm,clip]{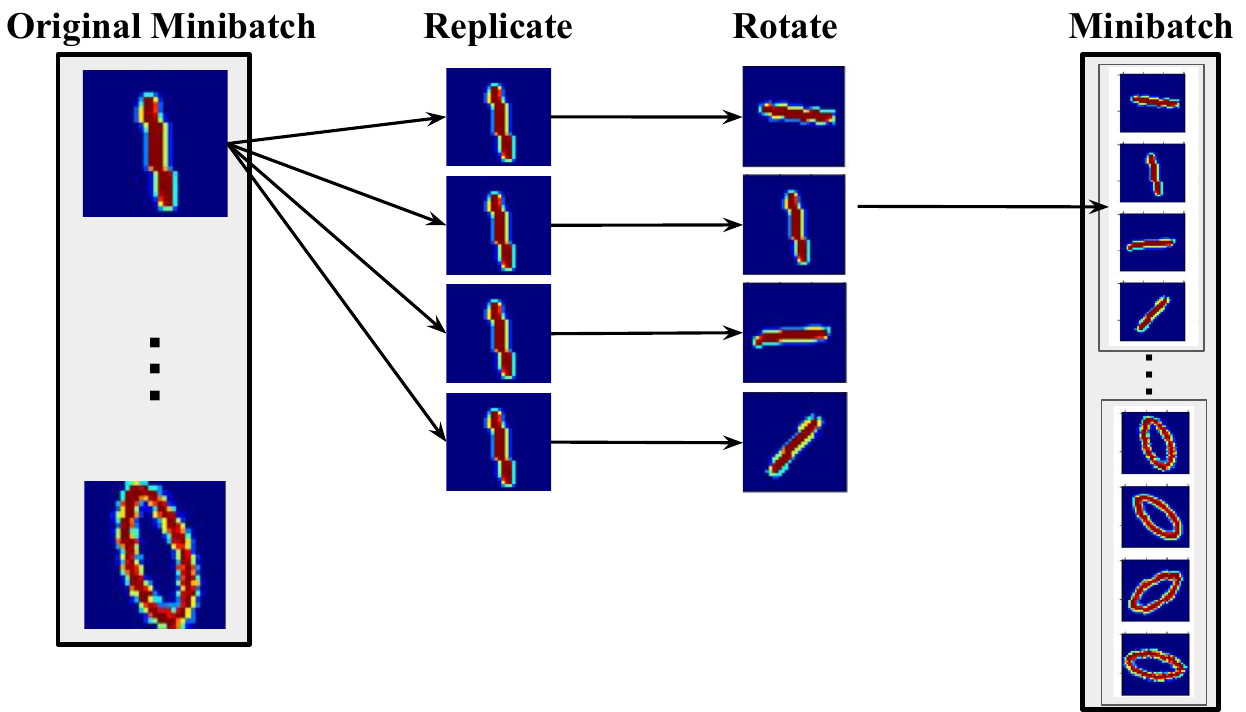}
    \end{center}
    \vspace{-2ex}
    \caption{Cartoon of the minibatch creation concept in our data pipeline. 1) We select eight images randomly from MNIST. 2) We create four identical copies of each image. 3) We apply a random rotation in the range [0$^{\circ}$, 360$^{\circ}$)
    to each of the 8$\times$4 = 32 images to obtain a minibatch of 32 images.}
    \label{minibatch.fig}
\end{figure}

We adjust our training minibatch to make our autoencoder learn the rotation-invariant features efficiently.
For instance, in learning MNIST set, we input all 10 digit classes. 
To implement Equation~(\ref{bottleneck_imp}) with diversity of transformation of training images in each minibatch, as illustrated in Fig.~\ref{minibatch.fig},
we randomly select eight digits from MNIST; 
replicate each image four times for a minibatch size of 32; 
and then rotate each of the 32 images by an angle selected at random from [0$^{\circ}$,  360$^{\circ}$) to produce a training minibatch.

\section{Physical reasonableness in \texttt{Phys}}\label{sec:appned_c1physdata}

\begin{figure}[htbp!]
    \centering
     \includegraphics[width=\linewidth,trim=0mm 1mm 0mm 1mm,clip]{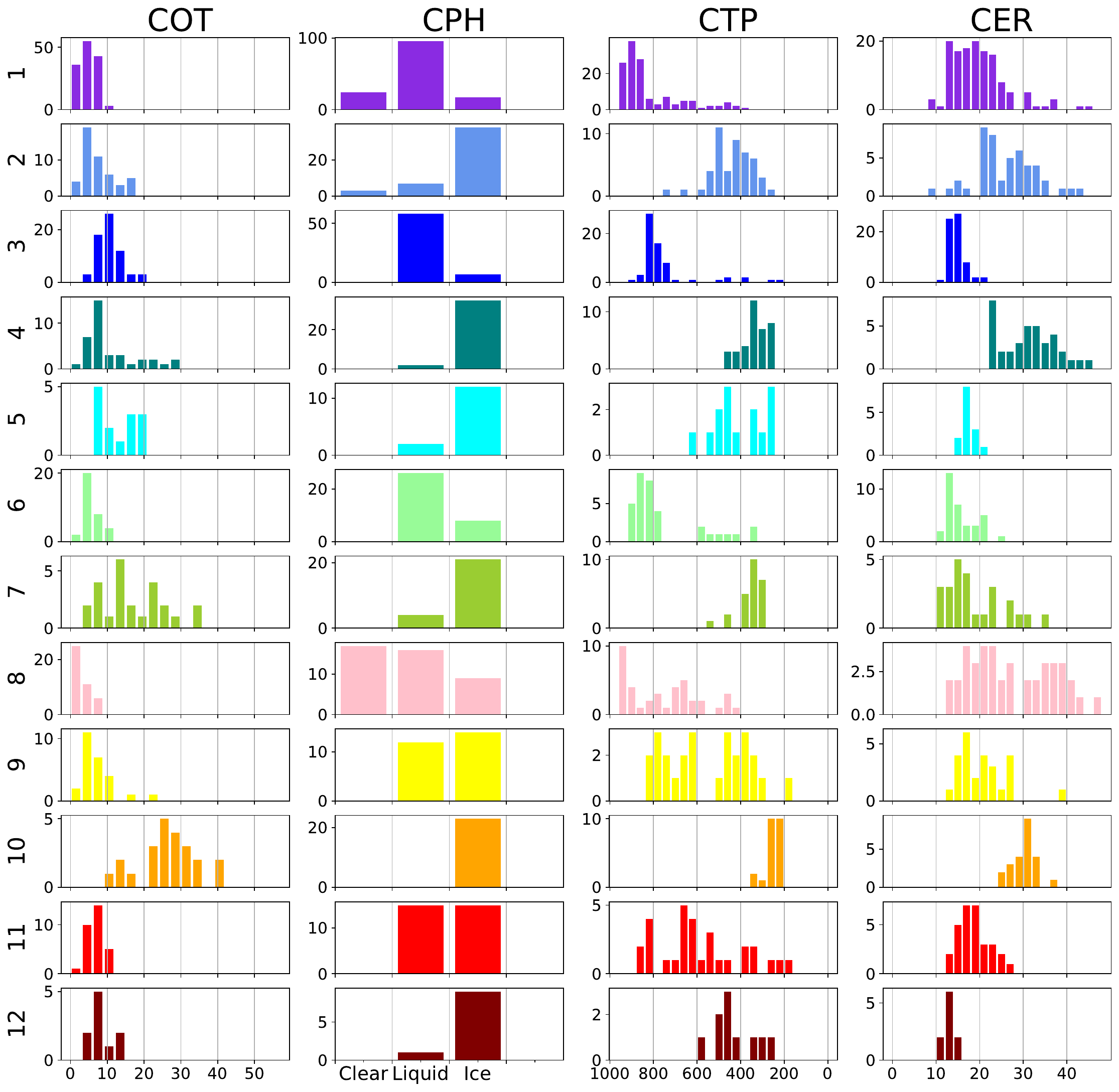}
    \caption{\emph{Test \ref{test11}--extension, cloud physics parameters for RICC applied to \texttt{Phys}}. Histograms of patch-mean values for four derived cloud physical parameters, cloud optical thickness (COT; no unit), cloud phase infrared (CPH; clear sky, liquid, ice), cloud top pressure (CTP; hPa), and cloud effective radius (CER; $\mu$m) show that these values are not randomly distributed within the 12 clusters produced by RICC.
    }
    \label{fig:physicalreasonablePhysdata}
\end{figure}

To verify the physical reasonableness of the clusters produced by RICC on the \texttt{Phys} dataset used in Sections~\ref{sec:test:spatial} and~\ref{sec:test:separable},
we examine histograms of physics parameter values in different clusters (they are not randomly distributed: 
see Fig.~\ref{fig:physicalreasonablePhysdata}), and compute the median inter-cluster correlation coefficient values (all are below our threshold of 0.6: 0.47 for cloud optical thickness, 0.45 for cloud phase infrared, 0.20 for cloud top pressure, and 0.39 for cloud effective radius).

\section{\Add{Visualizing Autoencoder Activity}}\label{sec:ricode}

\begin{figure}[htbp!]
    \centering
    \includegraphics[width=1\textwidth,trim=0mm 0mm 0mm 0mm,clip]{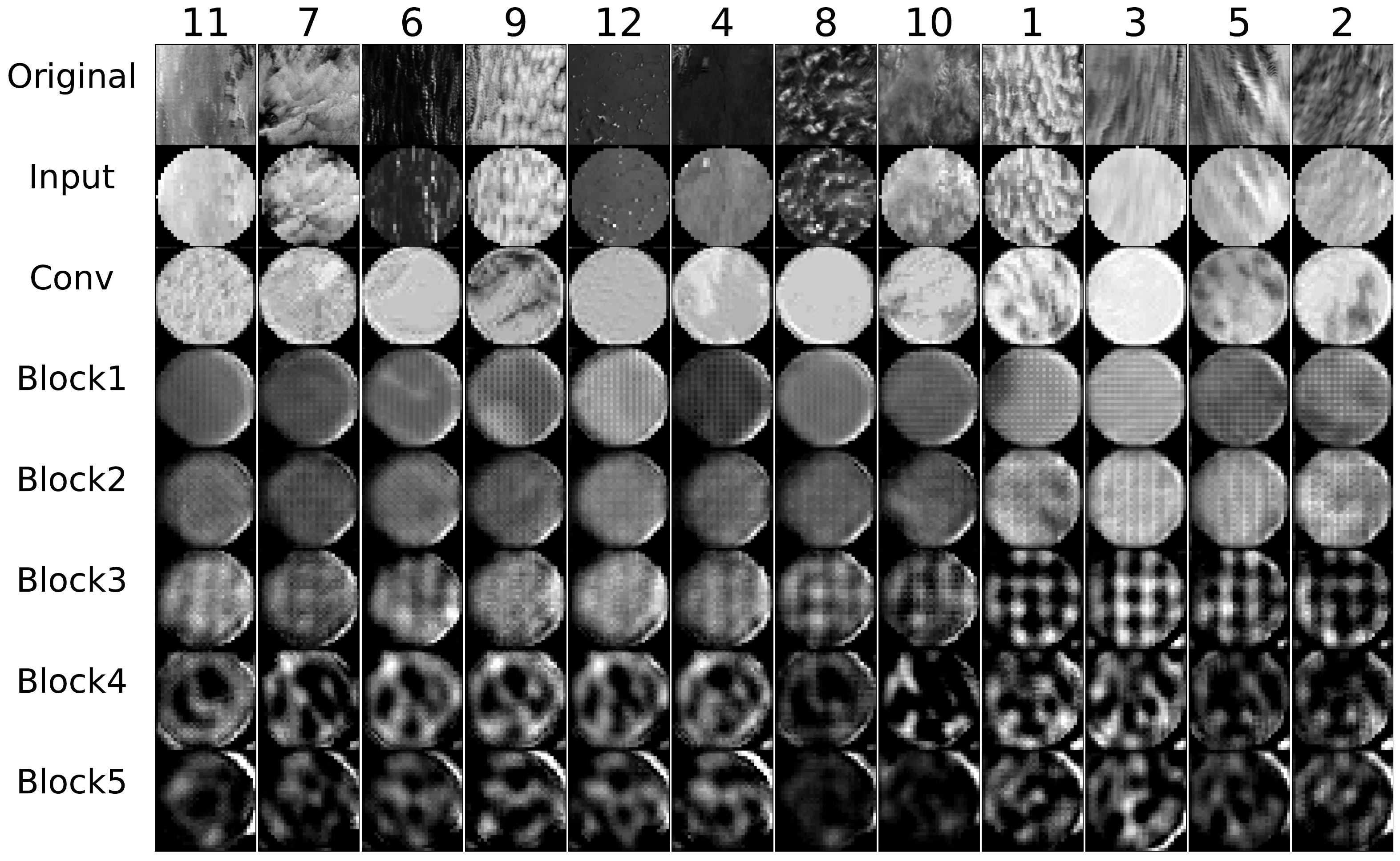}
    \caption{\Add{We visualize activations by using guided backpropagation~\cite{springenberg2014striving} in 12 example feature maps produced by the RI autoencoder. Rows from top to bottom show 12 patches from the 12 clusters of Fig.~\ref{fig:dendrogram} (cluster numbers are shown at the top aligned with Fig.~\ref{fig:dendrogram}), their inputs with circular masking, and the corresponding evolution at layers described in Table~\ref{model-table}.} 
    \Addafter{Blocks 1--5 (\emph{block} is a stacked three convolutional layers described in Table~\ref{model-table})}
    \Add{are saliency maps of their feature map projections at the final convolutional layer. Whiter colors show the larger gradient of weights, emphasizing features activated by neurons. Black colors represent zero or smaller gradients, meaning less attention of features by neurons. }}
    \label{fig:activationmap}
\end{figure}

The behavior of an autoencoder can be studied by visualizing the patterns learned at different convolutional layers.
\Add{It is useful to show activation maps~\cite{springenberg2014striving} for the perceptual interpretation of RI autoencoders by examining example images.
These images provide insights into what our autoencoder is learning.
We perform guided backpropagation where we set all negative gradients to 0 when propagating the gradient computation.
The constraining weights to be non-negative shows what cloud features neurons detect in different layers.
The sensitivity of detection by neurons in our RI autoencoder has large contrasts of the strength of gradients across different layers and different patches; 
white colors depict larger gradient values, and black colors are closer to zero or negative values.}
\Addafter{We observe in Fig.~\ref{fig:activationmap} that structures seen in the first layer have similarities to those seen in the inputs,}
\Add{and the checkerboard of Block 3} 
\Addafter{seems to capture cellular structures}
\Add{between with and without cloud pixels.} 
\Addafter{For other filters are, however, it is difficult}
\Add{to interpret their association between inputs and their activation maps.}

\AddAdd{Hence, we calculate correlation coefficients of activation maps with our physical properties shown in Fig.~\ref{fig:physicalreasonable}.
We observe that cloud effective radius (CER), a parameter that demonstrates the most distinct distributions based on Table~\ref{tab:stats:physicalreasonable}, shows moderate correlation (0.51) in Block 2, 
We also observe weak correlations at the first convolutional layer in cloud phased infrared (CPH) and cloud top pressure (CTP); 
the correlation coefficient of CPH is 0.37 and that of CTP is 0.32.
CTP also shows another weak correlation in Block 3 (0.36).
The remained parameter, cloud optical thickness, shows the peak correlation coefficient at the first convolutional layer but the value can be negligible (0.11). 
Since CER is associated with cloud optical properties~\cite{steven17}, and CPH and CTP are both associated with cloud top properties~\cite{baum12},
we conclude that the shallower layers can roughly capture intrinsic features of cloud properties in our selected six radiances
(Note again that we did not use MOD06 cloud product for our input images to our autoencoders), 
and the results correspond to how much resulting histograms are distinct in Fig.~\ref{fig:physicalreasonable}. 
}

\section{\Add{Pseudeocode for RICC Framework}}\label{sec:ripseudocode}

\Add{We show in Algorithm~\ref{alg1} the overall RICC training, clustering, and evaluation process described in Sections~\ref{sec:ricc} and~\ref{sec:evals}.
}
\begin{algorithm}
\caption{Pseudocode for the overall RICC training and clustering algorithm described in Section~\ref{sec:ricc} (lines 1--13) and the evaluation process described in Section~\ref{sec:evals} (line 14).}
\label{alg1}
\algnewcommand\algorithmicinput{\textbf{Input:}}
\algnewcommand\algorithmicoutput{\textbf{Output:}}
\algnewcommand\Input{\item[\algorithmicinput]}%
\algnewcommand\Output{\item[\algorithmicoutput]}%
\begin{algorithmic}[1]
    \Input $S =\{S_1, \cdots, S_n\}$ : set of MOD02 images  
    \Output $C = \{c_1, \cdots, c_{12}\}$: set of 12 clusters.
    \For{$S_1 \cdots S_n$:}
        \State Alight with MOD35 cloud mask product 
        \State Subdivide $S_i$ to $X =\{x_1, \cdots, x_N\}$: set of patches
        \State Apply quality control if cloud pixels per patch $> 30 \% $ 
        \State Circular masking and rescale in $[0,1]$
    \EndFor
    \State Train and test RI autoencoders
    \While{Passed}
        \State Train a RI autoencoder for a fixed $\lambda_{\textrm{res}}$
        \State Perform hyperparameter search using Fig.~\ref{gridsearch_protocol}
        \State Determine an optimal pair of $(\lambda_{\textrm{inv}},\lambda_{\textrm{res}})$
    \EndWhile
    \State Perform clustering, giving a set of cloud clusters $C = \{c_1, \cdots, c_{12}\}$
    \State Evaluate $C$ with MOD06 product to investigate; 
    \begin{itemize}
        \item Criterion 1: Physical Reasonableness using Test~\ref{test11}
        \item Criterion 3: Separable Clusters using Test~\ref{test31}
        \item Criterion 4: Rotation-Invariance using Test~\ref{test42}
    \end{itemize}
    \State Return clusters $C$ satisfying criteria 1--4.
\end{algorithmic}
\end{algorithm}

\end{appendices}
\end{document}